\documentclass[prd,aps,twocolumn,showpacs,superscriptaddress,floatfix]{revtex4-1}
\usepackage{bm}
\usepackage{amsmath}
\usepackage{txfonts}
\usepackage{graphicx}
\usepackage{dcolumn}
\usepackage{bm}
\usepackage{color}
\usepackage{multirow}
\usepackage{subcaption}
\usepackage{color}
\def\comment#1{}

\def\slashchar#1{\setbox0=\hbox{$#1$}           
	\dimen0=\wd0                                 
	\setbox1=\hbox{/} \dimen1=\wd1               
	\ifdim\dimen0>\dimen1                        
	\rlap{\hbox to \dimen0{\hfil/\hfil}}      
	#1                                        
	\else                                        
	\rlap{\hbox to \dimen1{\hfil$#1$\hfil}}   
	/                                         
	\fi}                                         %

\DeclareMathOperator\erfc{erfc}
\DeclareMathOperator\Ai{Ai}

\begin{document}


\title{The Stochastic Replica Approach to Machine Learning: Stability and Parameter Optimization}

\author{Patrick Chao $^*$}
\author{Tahereh Mazaheri $^*$}
\author{Bo Sun $^*$}
\author{Nicholas B. Weingartner}
\email{The first authors are alphabetically listed.}
\author{Zohar Nussinov}
\email{Correspondence to: zohar@wuphys.wustl.edu.}
\affiliation{Department of Physics, Washington University in St. Louis,
Campus Box 1105, 1 Brookings Drive, St. Louis, Missouri 63130, USA}%

\date{\today}

\begin{abstract}
We introduce a statistical physics inspired supervised machine learning algorithm for classification and regression problems. The method is based on the invariances or stability of predicted results when known data are represented as expansions in terms of various {\it stochastic functions}. The algorithm predicts the classification/regression values of new data by combining (via voting) the outputs of these numerous linear expansions in randomly chosen functions. The few parameters (typically only one parameter is used in all studied examples) that this model has may be automatically optimized. The algorithm has been tested on 10 diverse training data sets of various types and feature space dimensions. It has been shown to consistently exhibit high accuracy and readily allow for optimization of parameters, while simultaneously avoiding pitfalls of existing algorithms such as those associated with class imbalance. The ensemble of stochastic functions that we use suggests a way of deriving algorithm independent bounds on the accuracy. We very briefly speculate on whether spatial coordinates in physical theories may be viewed as emergent ``features'' that enable a robust machine learning type description of data with generic low order smooth functions.
\end{abstract}


\maketitle{}

\section{Introduction and Problem Description} \label{sec:introduction}

Humankind is unequivocally living in the age of ``Big Data". The rapid increase in connectivity between people, businesses and consumers, the media, and more has led to an explosion of publicly and privately available data. New information is constantly generated by social media, polling, market surveys, digital cameras, government surveillance, smart phones, scientific experimentation, and a multitude more of the technological sources and innovations of the past few decades. By the year 2020, the rate of production of digital data is projected to be 44 times as high as the rate in the year 2009, and the overall amount of available data is projected to be as high as 44 zettabytes (1 zettabyte =$10^{12}$ gigabytes). This wealth of information available in the digital ecosystem, combined with ever-increasing information storage capacity, has incredibly far reaching implications in diverse applications \cite{BigData,EconData,Science,Storage,Survey,WageData}. In order to realize the potential of the available data, methods for gaining meaningful insights must be developed. As the sheer quantity of available data exceeds human computational capability, efficient computer algorithms must be created and implemented. This is where the field of machine learning comes into play \cite{ML}. 

Broadly, machine learning is the process of allowing computer programs to parse available data and learn (infer) general rules. The notion of fundamental importance in the previous statement lies in the term ``general". The goal of machine learning is to find a model using input data which can be generalized and applied to new data in such a way that model performance increases with increasing amounts of input data. The basic scheme consists of analyzing a set of input data (``training data'') containing many entities (instances) to which we want to assign some value (label). Each instance is described by a set of quantities (features) which, theoretically, allow it to be mapped to a specific label. The problem, then, is to find a mapping algorithm (model) with parameters which the computer can fit to the given input data, and subsequently apply to future data (``testing data''). Towards that end, there are two main types of machine learning algorithms: supervised and unsupervised. Unsupervised learning involves training data with unknown labels or associations. Unsupervised learning algorithms seek to label instances based on their connections or commonalities with other instances, via methods such as clustering \cite{CD,CD1,CD2,CD5,CD6}; these include clustering methods \cite{CD3,CD4} that employ object called ``replicas'' somewhat similar in spirit to those that we will introduce in the current work for supervised learning. Supervised machine learning corresponds to learning on training data that has known outcomes, i.e., data for which the ``right answer" is known. The algorithm aims to fit the model by using the relationship between the features and known labels, to effectively generalize to new data with unknown labels. Since the advent of supervised machine learning a number of algorithms have been developed. These are of varying complexity and performance, with some of the most popular being ``Support Vector Machine'' (SVM) methods \cite{SVM,SVM1}. One may wonder why, in light of the plethora of currently available powerful methods, should we be concerned with the development of novel algorithms? Crudely, in addition to the benefits of having a robust ``toolbox" of multiple algorithms, it turns out that existing algorithms are not without their faults. 

In this paper, we will specifically focus on supervised machine learning corresponding to data with either discrete (classification) or continuous (regression) labels. We will introduce our new algorithm that learns by fitting an ensemble of stochastic series expansions to the training data, and then `votes' on the output of the label. We will demonstrate, through detailed case studies, that this algorithm, which we term the ``Stochastic Replica Voting Machines'' (SRVM) method, rivals the best performing contemporary models, and additionally surpasses them in various performance metrics. We will demonstrate that the algorithm applies equally well to both classification and regression.

The remainder of this article is organized as follows: In Section \ref{sec:m description}, we describe the statistical physics based inspiration for the current algorithm.
In Section \ref{nut}, we provide the essential detailed setup of the algorithm. We then proceed to test our algorithm against various benchmarks (Section 
\ref{sec: result}). Apart from underscoring the high accuracy of the SRVM method, we report on traits such as the dependence of the results on the specific parameter that underlie our algorithm (Section \ref{dependence+}), its stability (Section \ref{sec:stable.}), and the dependence on our results on different methods of pre-processing (scaling) the data (Section \ref{pre-p}). Of particular interest are the overlaps (Section \ref{overlap+}) between the different stochastic functions or ``replica'' solvers that underlie our method. These overlaps correlate with the accuracy of our predictions thus enabling us to pinpoint optimal values of the parameters defining our algorithm.
In Section \ref{sec:imbalance}, we illustrate that, by its very nature of including numerous independent stochastic solvers, our approach suffers from far less bias than the common SVM machine learning method. We describe (Section \ref{layer:sec}) a trivial generalization of our method to include multiple ``layers'' wherein the voting between the different solvers allows for various differently chosen weights. In particular, as we explain, one may find the optimal parameter values of our method (potentially including generalized weights) by applying regression machine learning to machine to recursively learn and predict the parameters values that will yield the best accuracy (or replica overlap). In Section \ref{sec:srvr}, we 
further present tests on the residuals of our method to demonstrate its strength also for regression studies. In Section \ref{al_in}, we suggest that SRVM may be employed to obtain algorotuhm independent bounds on the accuracy attainable by any algorithm. We conclude (Section \ref{sec:conclusion}) with a summary of our main results and a speculation concerning coordinates in physical systems as emergent features for which the representation of the data is most robust.

\section{Basic tenets}  \label{sec:m description}
\begin{figure*}
	\centering
	\includegraphics[width=1.8 \columnwidth, height= .3 \textheight,keepaspectratio]{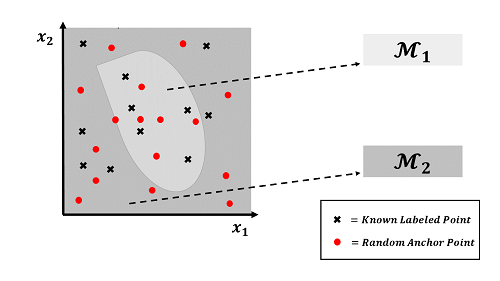}
	\caption{(Color Online.)`Phase space' representation of the feature space and the mapping of feature vectors to their respective classification label. Most variants of our algorithm rely on the use of `Anchor points' (see Section \ref{nut}).}
	\label{All.}
\end{figure*}
Recent decades have seen a flurry of advances in computer science that have been triggered and/or aided by various findings in the natural sciences. Indeed, artificial neural networks aim to emulate quintessential aspects of the biological networks of the brain. Neural networks witnessed tremendous success in advancing artificial learning \cite{ANN,DL}. The study of spin-glasses by physicists and materials scientists has led to the development of Hopfield networks. The incorporation of thermodynamic and statistical mechanics principles to these systems led to some of the most sophisticated machine learning models to date \cite{Hopfield, SpinGlass,SG1,SG2,SG3,Boltzmann}. It is evident that natural scientific principles (such as those from biology or physics) may serve as excellent bases for constructing learning algorithms. Recent results demonstrate that a certain theoretical basis may be required in order to enable learning algorithms to apply to scientific data \cite{BlackBox}. With these notions in mind, we formulated a novel algorithm for supervised learning that is motivated by statistical physics.

In the classical statistical mechanics of N-particle systems, e.g., \cite{huang} each particle carries its own phase space degrees of freedom: its position and momentum coordinates (thus in three-dimensional space, the state of each particle is defined by six degrees of freedom). At any given instant, the `list' of all particle coordinates and momenta for all the particles in the system completely specifies its instantaneous state. Thus, for a system of $N$ particles, this `microstate' can be represented as a single point in a 6N-dimensional phase space. The system itself, comprised of an extremely large number of particles, is macroscopic and can be described using only a few bulk degrees of freedom (i.e., temperature, pressure, magnetization, etc.). These bulk degrees of freedom characterize the observable state of the system in what is termed the `macrostate'. The dynamical evolution of the particles in the system causes the microstate to constantly change, transitioning to new points in the phase space (new `list' of 6N coordinates). If the system is in equilibrium, there is no change in macroscopic degrees of freedom with time, and this means that the microstates correspond in some way to the given macrostate. Additionally, the properties of the macrostate can be found by taking an \textit{ensemble average} over the microstates corresponding to the macrostate. In general, changing external constraints changes the microstates that are available to the systems particles, and the macrostate can also change. This implies that various sets of microstates correspond to specific macrostates, and this is indeed the case. More specifically, each microstate corresponds to only one specific macrostate. In the phase space picture, then, certain regions of phase space (corresponding to sets of microstates) will map directly to a single macrostate, and there will be boundaries in the phase space separating the different regions. 

The above description of statistical mechanical phase space is reminiscent of classification problems. As discussed in the Introduction, classification-based learning problems consist of instances (the particles) which are described by a set of features (positions and momenta). These values of the features for a given instance are cast into a feature vector which gives the `location' of the instance in high-dimensional feature space (phase space). Each instance has an associated classification label corresponding to the set of features, such that certain regions of feature space map to specific labels. The goal of the learning algorithm is to find the boundary between the classification labels in feature space, so that new instances (which correspond to some point in feature space) can be appropriately mapped to the proper label. A schematic is provided in Fig. \ref{All.}. 

In order to achieve this goal, we need an appropriate mapping function $f(\vec{x})$, where $\vec{x}$ is a vector representing a particular point in the space of all $d$ attributes (``features'') of the data. In the statistical mechanical framework, mapping to a specific macrostate is done via minimization of an appropriate free energy. Once the free energy is properly extremized, calculating its value for a given point in phase space will allow for the elucidation of the corresponding macrostate or phase. Twentieth century physicist Lev Landau studied free energies that could be expanded in a set of polynomial kernel functions of features (the so-called ``order parameters'') and their gradients. The kernel expansion with coefficients whose values were fixed through optimization could then be applied to determine which macrostate a region of phase space belonged to (see, e.g., \cite{huang,lev}). Thus, borrowing this idea, since we are interested in identifying classification boundaries, we will assert that the label, $y_i$ of a given instance can be expanded with unknown coefficients, in a set of kernel functions which take as their argument the feature vector. For a binary classification problem, the sign of a voting function weighted by different ``replica'' functions $f$
determines the classification of the vector $\vec{x}$. 

\begin{figure*}
	\centering
	\begin{subfigure}[t]{.45\textwidth}
		\centering
		\includegraphics[width=\linewidth]{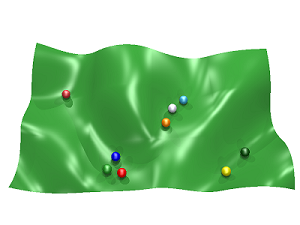}
	\end{subfigure}
	\begin{subfigure}[t]{.45\textwidth}
		\centering
		\includegraphics[width=\linewidth]{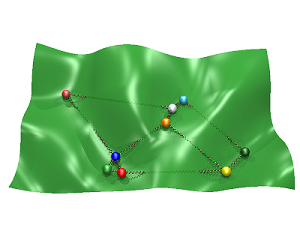}
	\end{subfigure}

		\caption{\label{fig:MRAlandscape} (Color Online.) Reproduced from Ref. \cite{mychapter}.  In panel (a), independent machine learning solvers 
		navigate the ``feature landscape'' and minimize the energy of Eq. (\ref{Fmin}). In panel (b), these independent solvers ``interact'' with each (as symbolically represented by springs) in order to 
		find better minima (and potentially larger better predictions). This qualitatively emulates a minimization of Eq. (\ref{FET}). In our work, the outcome of these interactions will be set to be a simple arithmetic average of the predictions from each of the individual replicas (Eq. (\ref{average})). 
		More complex interactions are possible (as will be briefly touched on in Section \ref{layer:sec}).}	
\end{figure*}

The general idea underlying our use of ``replicas'' is sketched in Fig. (\ref{fig:MRAlandscape}). In essence, the system may be examined  independently by 
random machine learning solvers (the ``replicas''). These replicas may collectively interact with one another in order to produce a collective prediction that is more stable and less biased than that potentially found by a single solver. This idea was used in \cite{CD3,CD4} for general unsupervised machine learning (clustering),
unsupervised image segmentation \cite{vision,vision1,vision2}, determining structure in various phases of complex many body systems \cite{phase1,phase2},
and for examining instances of the Traveling Salesman Problem \cite{TSM}. In using multiple replicas, we aim to capture an anthropological principle known as 
{\it wisdom of the crowds} \cite{wisdom}: the predictions made by a large crowd may far more accurate than the guess made by a single person (a single solver or ``replica'').

\section{The SRVM algorithm in a nutshell- mathematical details}
\label{nut}

We will now couch the above intuition in a rather concrete and exceedingly simple mathematical framework. The resulting recipe will lead to an algorithm that may be straightforwardly implemented. Similar to other supervised machine learning approaches, the algorithm that we construct will be trained using instances of known labels. Typically, such training data sparsely cover the space defined by the features of the data. To work around this, our algorithm will employ an ``ensemble averaging'' technique that randomly samples the feature space. We will generate a stochastic set of $v$ feature vectors (associated with points in feature space) that we will term `anchor points'. We will then use the proximity of these anchor points to training points to assign a classification label. Essentially, we will employ the known labels corresponding to training points in feature space (instances) with various kernel functions to attempt to classify the space around the known points so as to create general mapping functions. Specifically, we consider a specific input ``training'' data of size $N$ points each comprised of $d$ features (expressed as a $d-$ dimensional vector) for these points  $\{\vec{x}_{i}\}_{i=1}^{N}$ and the corresponding given correct classification $\{y_{i,c}\}_{i=1}^{N}$. With these preliminaries in place, we now define
\begin{eqnarray}
{y_{i,p}^\alpha \equiv f^{\alpha}(\vec{x}_i) \equiv \sum_{j=1}^{v} c_j^\alpha ~K(\vec{x}_i,\{\vec{\chi}_j^\alpha\})},
\label{Map}
\end{eqnarray}
and aim, as we will describe below, to set $y_{i,p}^\alpha$ equal to the known correct classification $y_{i,c}$.
Here, $\{\vec{\chi}_j^\alpha\}_{j=1}^{v}$ are fixed random vectors (which we will often term ``anchor vectors'') that are different for each ``replica'' $\alpha$,
and $K$ is a stochastically chosen function. It may, e.g., be any standard function, 

\begin{eqnarray}
\centering
{      K(\vec{x}_i,\vec{\chi}_j^\alpha)        =} \left\{
\begin{array}{l}
\exp\left(-\frac{(\vec{x}_i-\vec{\chi}_j^\alpha)^2}{2 {\sigma}_j^2}\right) \\             
\exp\left(-\frac{\lvert \vec{x}_i-\vec{\chi}_j^\alpha\rvert}{\xi}\right)  \\   
\erfc\left(\frac{\lvert\vec{x}_i-\vec{\chi}_j^\alpha\rvert}{\gamma}\right) \\
 \Ai\left(a\lvert\vec{x}_i-\vec{\chi}_j^\alpha\rvert\right) \\
 \frac{1}{1+\exp\left({q}_j\lvert\vec{x}_i-\vec{\chi}_j^\alpha\rvert\right)} \\
 ... \\
 \end{array},
\right. 
\label{Klist}
\end{eqnarray}
where $\sigma$, $\xi$, $\gamma$, $A$, and $q$ are constants that serve as defining parameters for the (Gaussian, exponential, complementary error function, Airy functions (of the first kind), and Fermi type distribution (the latter is also known, when $q=1$, as the Logistic function)) functions that appear above.  The kernels used in Eq. (\ref{Map}) need not all be of the same type. Any linear combination of different kernel types (such as the different explicit functions listed in Eq. (\ref{Klist}) or Eqs. (\ref{multi}, \ref{multig},\ref{Pade}) that we will introduce shortly) might also be chosen. Indeed, in order to avoid spurious behavior of functions $f^{\alpha}(\vec{x})$ of Eq. (\ref{Map}) stemming from the trivial asymptotics of the various individual kernel types, we may select the kernels that appear in Eq. (\ref{Map}) to be of multiple types.
The equations that we will result for the coefficients $c_j^\alpha$ will be linear in  all of these cases. On all the examples that we studied, the single kernel type expansion fared well (we found modest improvements on including multiple function types (Section \ref{layer:sec}). However, it is conceivable that on other data sets a heterogeneous set of kernel types may fare substantially better. As we will further explain, in Eq. (\ref{Map}), $\{f^{\alpha}\}_{\alpha=1}^{\mathcal{R}}$ is a set of viable functions of the variables $\vec{x}$
(different specific functions (either of various types (Eq. (\ref{Klist})) or, more commonly in our simplest analysis, functions of a certain general type having yet different fixed vectors $\{\vec{\chi}_{j}^{\alpha}\}$) associated with ``different replicas'' $\alpha$). 
We may trivially re-express the above as
$\vec{y}^{~\alpha}=\underline{K}^\alpha\vec{c}^{~\alpha}$ 
where $K_{ij}^\alpha=K(\vec{x}_i,\vec{\chi}_j^{~\alpha})$. Thus, inverting Eq. (\ref{Map}), we have
\begin{eqnarray}
{\vec{c}^{~\alpha}=({\underline{K}^{\alpha}})^{-1}\vec{y}_c},
\label{inverse-}
\end{eqnarray}
where $\vec{y}_c$ is the vector (with components $y_{i,c}$) of correct classification results
and $({\underline{K}^{\alpha}})^{-1}$ is the inverse of the Kernel matrix. With the aid of Eq. (\ref{inverse-}), we may solve for the coefficients $c_n^\alpha$. Typically, the systems that
we study are underdetermined. Therefore, the inverse matrix $({\underline{K}^{\alpha}})^{-1}$ is actually a pseudo-inverse; finding the coefficients $c_n^\alpha$
involves a least squares fit. 
The pseudo-inverse of Eq. (\ref{inverse-}) minimizes, for each replica $\alpha$, the ``learning energy''
\begin{eqnarray}
E^{\alpha}=\sum_{i=1}^{N} (y_{i,p}^{\alpha}-y_{i,c})^2.
\label{Fmin}
\end{eqnarray}
Here, $y_{i,p}^{\alpha}$ represent the predicted ($p$) results (as given by Eq. (\ref{Map})) while $y_{i,c}$, as noted above, are the replica independent correct ($c$) classification results that a good algorithm aims to uncover. Thus, the coefficients $c_j^\alpha$ that are calculated for a given replica $\alpha$ will appropriately map a given ``state'' $\vec{x}$ to the correct ``phase'' label given the phase space sampling information. We repeat the above calculation for multiple stochastic sets of replicas ($\mathcal{R}$ in total) in an attempt to ``ensemble average" based on knowledge of the actual phase space mapping to appropriately find the correct divisions. As the mapping functions $f$ are continuous while the classification labels are discrete, the output of the mapping function for each replica has to be thresholded. Once the system is ``trained'' with the training data (i.e., given the training data, the coefficients $c_j^\alpha$ are fixed by Eq. (\ref{inverse-})), we examine what occurs for new ``test'' input vectors $\vec{x}$. For the binary classification cases that will be largely studied throughout this paper, we will typically set, for each replica $\alpha$, 
\begin{eqnarray}
\centering
{y^{\alpha}(\vec{x})=}  sgn \Big(f^{\alpha}(\vec{x}) \Big) \equiv
\left\{
\begin{array}{ll}
-1 & f^{\alpha}(\vec{x})< 0 \\             
1 & f^{\alpha}(\vec{x})\geq 0 \\   
\end{array}.
\right. 
\label{Thresh}
\end{eqnarray}
This thresholding may be generalized for multi-class classification. In general Receiver Operating Characteristic (ROC) curves \cite{ROC} can be used to test for the best value of the threshold. Once the output of Eq. (\ref{Thresh}) is computed for all points in each of the replicas, the \textbf{overall classification} of an instance is found via \textbf{voting}. The ``overall'' label of a given instance is found by taking the average of the values predicted for that instance across all replicas, and then appropriately thresholding it (as in Eq. (\ref{Thresh})). 
\begin{eqnarray}
\mathcal{V}(\vec{x})= sgn \Big( \sum_{\alpha=1}^{\mathcal{R}} y^\alpha(\vec{x}) \Big),
\label{average}
\end{eqnarray}
where $y^\alpha(\vec{x})$ is the predicted label by the $\alpha$-th replica. The process of voting based on stochastic replicas allows for the correction of occasional mislabeling due to random fluctuations, and leads to a more reliable final result.

The specific, equal weight, voting scheme of Eq. (\ref{average}) is one of many possible voting choices that may be employed. As we will briefly touch on later, multiple voting methods could be used to increase the overall performance (Section \ref{layer:sec})); the multi-replica voting schemes qualitatively emulate ``interactions'' between the different replicas (diagrammatically represented by springs in Fig. \ref{fig:MRAlandscape}). Since the averaging implicit in voting leads to a continuous range of voting outcomes, the same thresholding methodology of Eq. (\ref{Thresh}) will be employed. In the tests that we performed, we contrasted our results with those found by SVM. 

For completeness, we remark on possible ``phase transitions'' that may appear in the data as a function of feature values (and that we largely did not test for in our analysis). In real systems (including physical ones as originally investigated by \cite{huang,lev}), different behavior might appear in distinct feature
regime values. In the contact of other data sets, this may, e.g., the performance of athletes before and after an injury. When such ``phase transitions'' are present then when the values of the features are varied across these boundaries, non-analyticities will appear; different functional forms will be 
needed to describe the system in its different phases. (These phase transitions in the data are different from the phase transitions associated with solvability 
and correct classification (see, e.g., \cite{dandan+,lenka} for transitions in unsupervised clustering/classification).) Towards this end, one may employ Eq. (\ref{Map}) in a subvolume of ``feature space'' (for all training data points that lie in this region) to see if the accuracies may vary and transitions are encountered (sharply distinct functional forms become optimal across phase boundaries) as evinced by striking changes in the overlap between different replicas. If the ultimate function that underlies the correct classification exhibits no (or only mild) such singularities then good classification may be obtained sans a detailed investigation as to how the classification results for a single point $\vec{x}$ change when training data in different subvolumes of feature space around $\vec{x}$ are used in our algorithm. 

To close our circle of ideas and description, we return to our main intuition. As noted in Section \ref{sec:m description} and underscored once again here, the guiding principle behind our method is, in a conceptual nutshell, that of  \newline

\fbox{\begin{minipage}{15em}
\center
{``Wisdom of the Crowds \it{for Fits}''.}
\end{minipage}}
\newline

By this statement, we mean that if different attempted fits (e.g., Eq. (\ref{Map}) with varying kernels) all yield the same prediction for a new data point $\vec{x}$ then regardless of the ``exact'' functional form (if such an exact function exists and may be solved for) that describes the data in physics or other problems, practically, the common classification predicted by all of these random fits (``replicas'') for the point $\vec{x}$ is likely to be the correct one. Indeed, the possibility of multiple fits that all yield a similar prediction appears across many fields of science. In all numerous problems, the precise underlying functional form explaining the data is unknown yet various fits all leads to similar predictions at temperatures, pressures, etc., where the experiments can be performed. 

The inter-replica voting that we use amongst the outcomes of the random real functions emulates interactions between the individual replica solvers. In a physics parlance, we not only minimize a cost function of Eq. (\ref{Fmin}) for individual solvers given training
data. We also take into account the collective (voting) outcome and correlations between the individual
solvers. Qualitatively, this emulates a minimization of a ``free energy'' type function of a free energy 
given energy and entropy,
\begin{eqnarray}
F(\vec{x}) =  E (\vec{x})- T S(\vec{x}),
\label{FET}
\end{eqnarray}
 where $E (\vec{x}) = \sum_{\alpha} E^{\alpha}(\vec{x})$
 and here $S(\vec{x})$ denotes the information about correlations 
between the replicas (i.e., in our case, the votes of Eq. (\ref{average})) as to the correct classification of feature space point $\vec{x}$.
In Eq. (\ref{FET}), the weight $T>0$ emulates the appearance of temperature as it appears in free energy minimization problems.
Eq. (\ref{FET}) is only provided for qualitative reference. The classification that we will use is that provided by Eqs. (\ref{Map} - \ref{average}). We will illustrate the utility of the ``wisdom of the crowds for fits'' maxim in our study of numerous examples that we embark on next. 

\section{Quantitative analysis of the SRVM Algorithm} \label{sec: result}

To assess the performance of the SRVM algorithm, we will apply it to several test data sets and examine various statistical performance metrics. 
In order to ascertain the ability of the SRVM algorithm to model the data, we split (as is customary) the data into two parts: a training set and a testing set. The training set was used to construct the model (i.e., the model was found by solving Eqs. (\ref{Map},\ref{Fmin})). Subsequently, the testing data set was used to evaluate the performance of the model. Some of the data sets employed in testing the SRVM algorithm that are discussed in this paper came with explicit testing data sets. For other data set benchmarks, no explicit test set are provided; in these cases, five-fold cross validation (CV) techniques are employed to fit and analyze the model. Five-fold CV involves randomly splitting the data set into five equal size subsets or folds, and using 4 of these folds together as a training set and the fifth fold as a testing set, while iteratively cycling through so that each fold serves as the testing set once. This allowed us to analyze the performance of the model for multiple folds, as well as report average performance metric values across all five folds. This five-fold CV was used throughout to ascertain the accuracy. Unless explicitly noted otherwise, all accuracies that we report were obtained by five-fold CV. 

\begin{figure}[h]
\centering
\includegraphics[width=0.95\columnwidth]{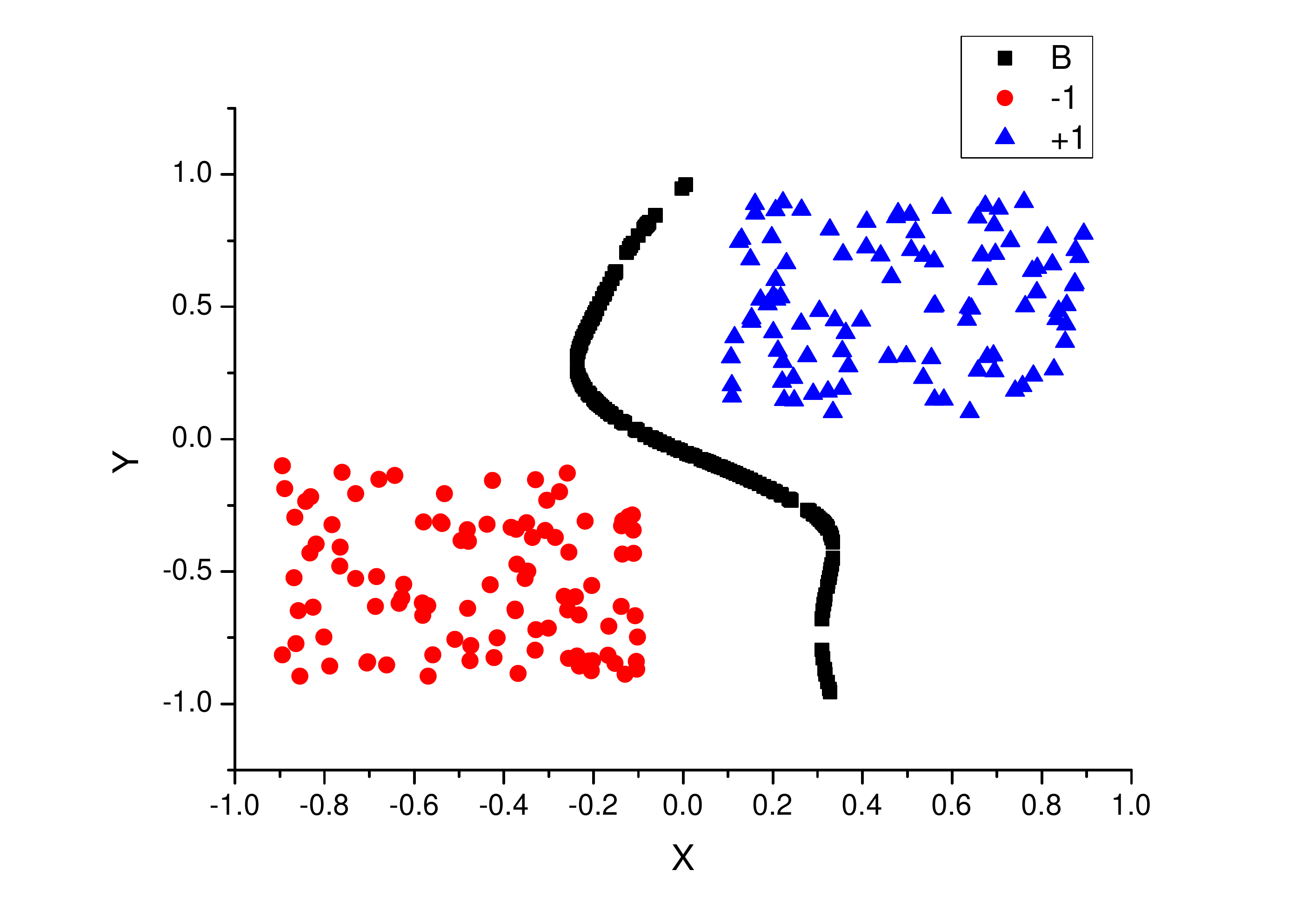}
\caption{(Color online.)
The boundary formed by Gaussian Kernel algorithm in the linearly separable case.}
\label{fig:figzerozero}
\end{figure}

\begin{figure}[h]
\centering
\includegraphics[width=0.95\columnwidth]{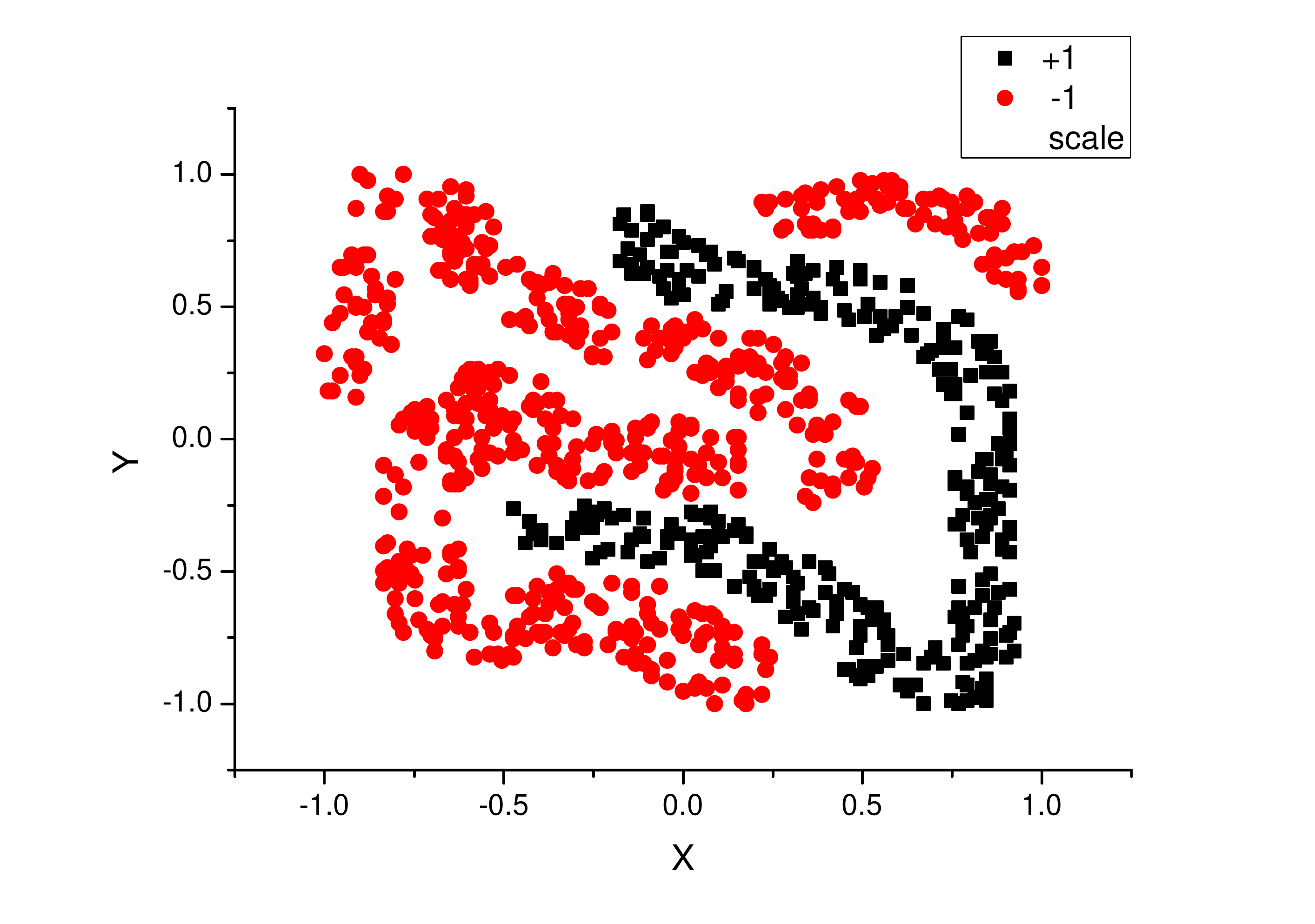}
\caption{(Color online.)
The raw data of the Four-class problem.}
\label{fig:figzero}
\end{figure}

\begin{figure}[h]
\centering
\includegraphics[width=0.95\columnwidth]{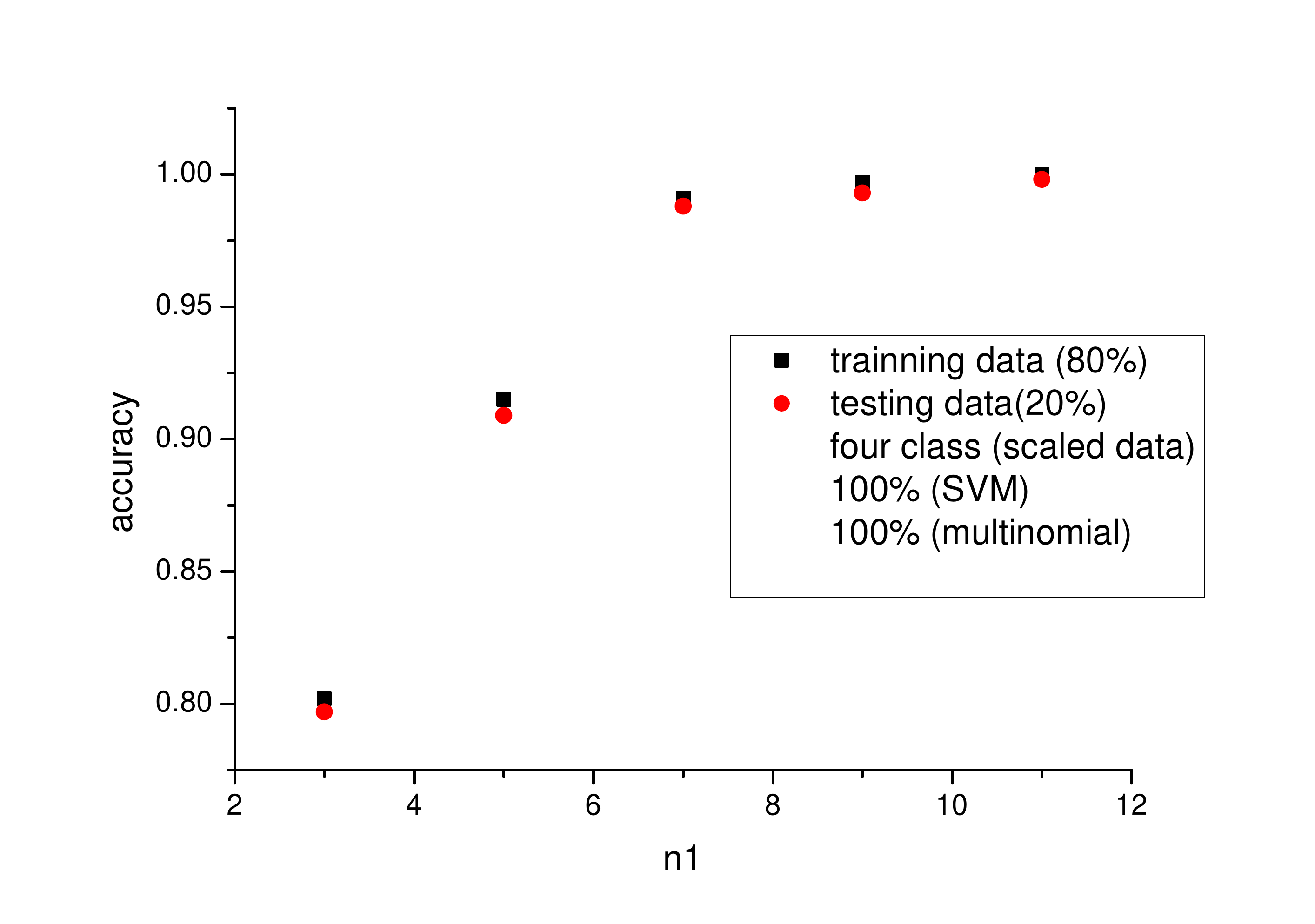}
\caption{(Color online.)
The accuracy of the multinomial variant of our SRVM algorithm for the Four-class problem.
Here, $n_{1}$ denotes the highest power (of any of the features $x_{k}$) in the multinomial expansion of Eq. (\ref{multi}).}
\label{fig:figzeroo}
\end{figure}

\begin{figure}[h]
\centering
\includegraphics[width=0.95\columnwidth]{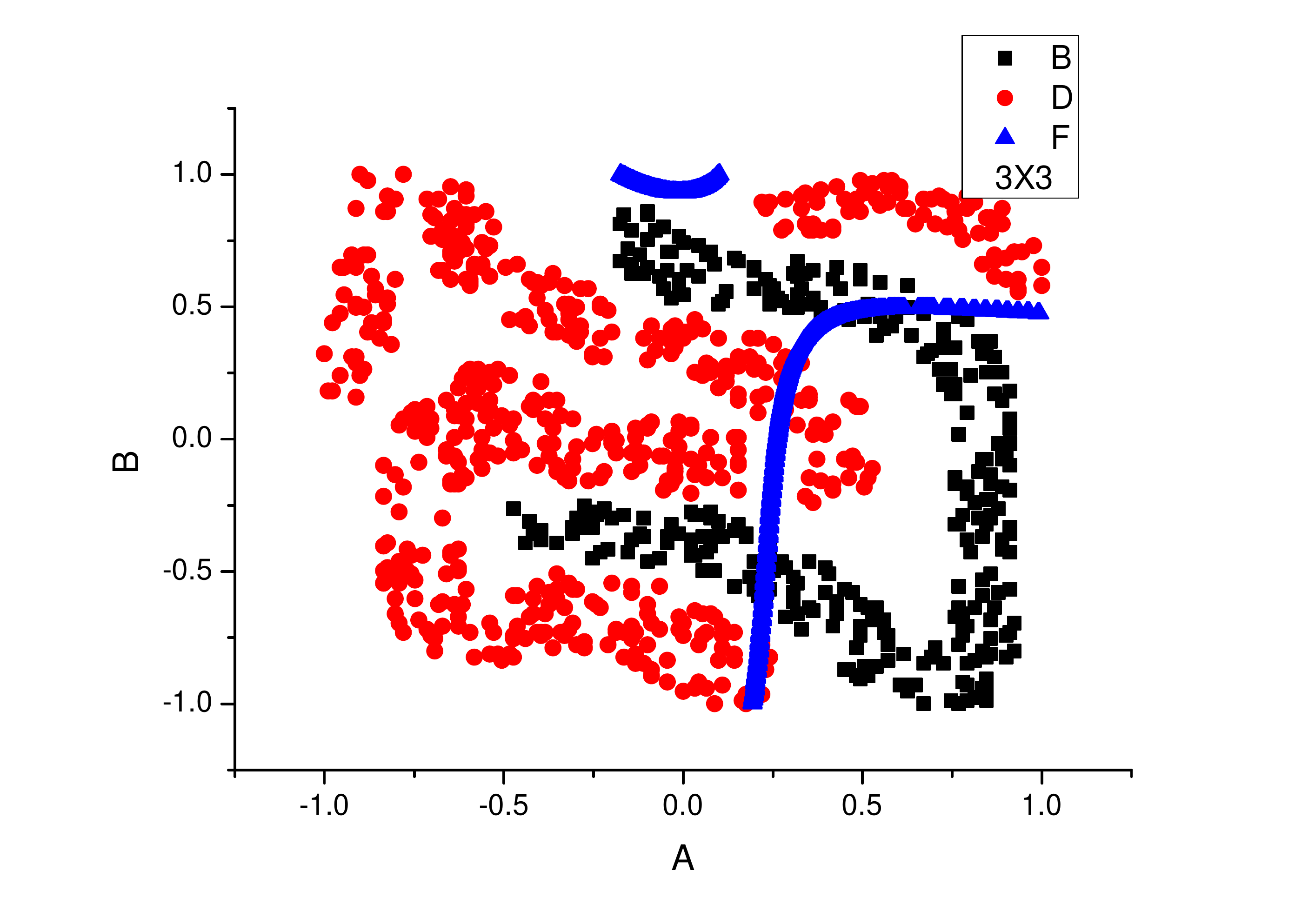}
\caption{(Color online.)
Boundary obtained subsequent to training using the multinomial kernel algorithm for the Four-class problem when $n_{1} (=n_{2}) = 3$.  Only the training points are shown. The signifiers B and D represent the +1 and -1 classes respectively. F denotes the boundary between these two classes as determined by the SRVM to this cubic order.}
\label{fig:figthree}
\end{figure}

\begin{figure}[h]
\centering
\includegraphics[width=0.95\columnwidth]{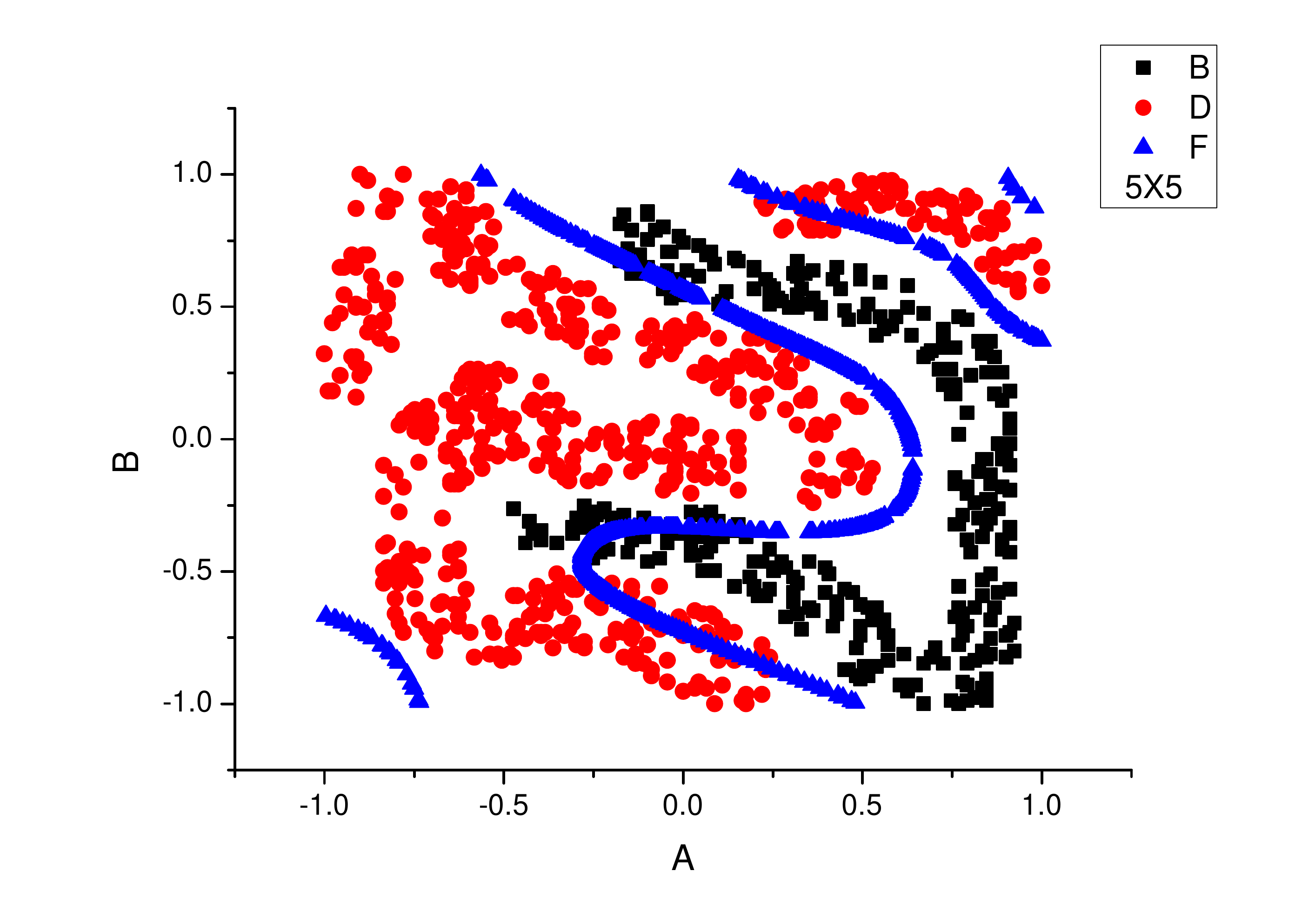}
\caption{(Color online.)
Boundary determined by the training data when using the polynomial kernel algorithm for the Four-class problem when $n_{1}=n_{2}= 5$ in Eq. (\ref{multi}). As before, B and D mark the +1 and -1 classes respectively. Only the training points are shown. The curve F denotes the boundary found by the SRVM algorithm between these two classes to this quintic order.}
\label{fig:figfive}
\end{figure}

\begin{figure}[h]
\centering
\includegraphics[width=0.95\columnwidth]{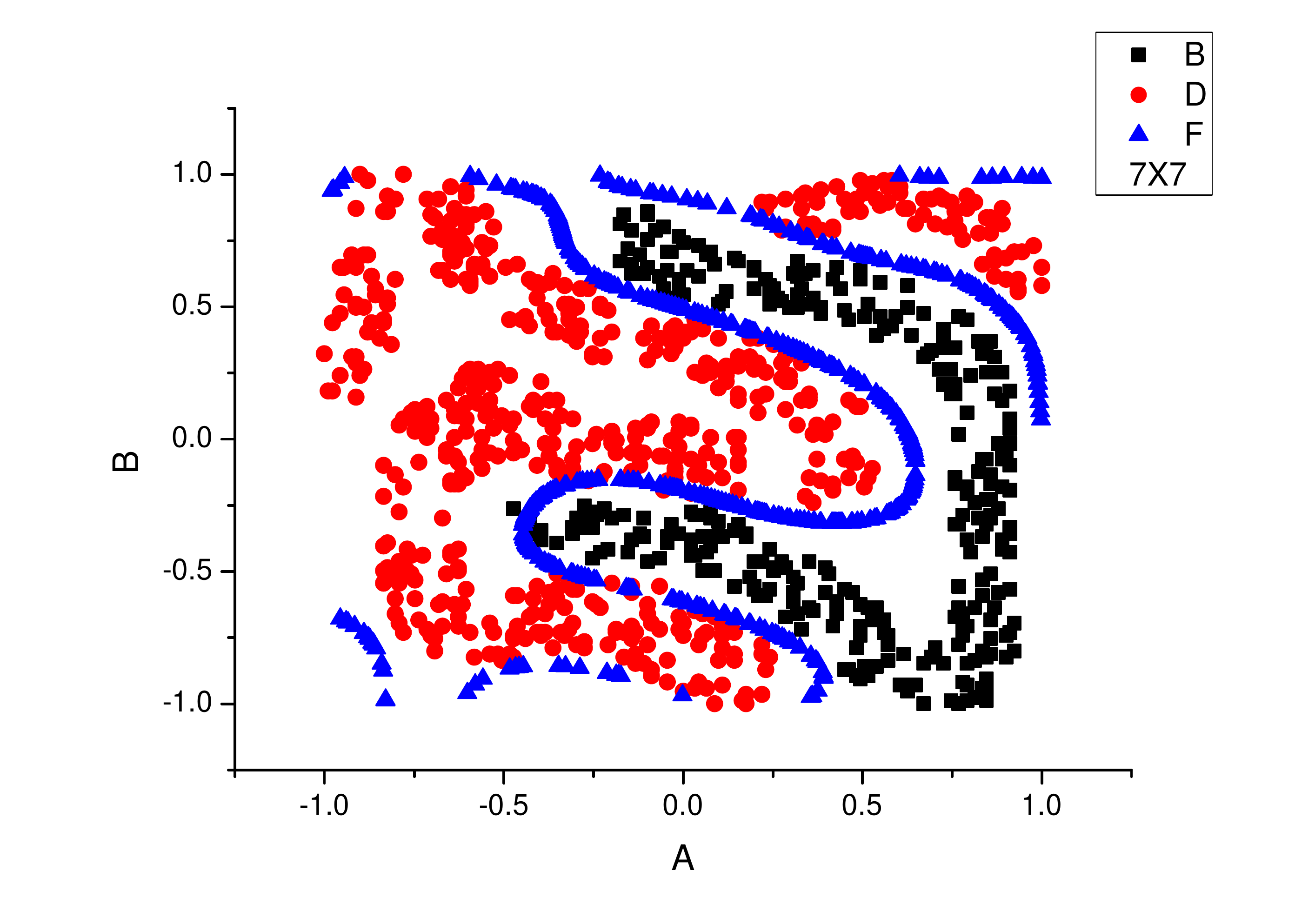}
\caption{(Color online.)
B and D represents the +1 and -1 classes respectively in the Four-class data set. The curve F denotes the boundary between these two classes to this Boundary predicted by the SRVM algorithm when using a multinomial of order $n_{1} = n_{2} = 7$. Only the training points are depicted.}
\label{fig:figseven}
\end{figure}

\begin{figure}[h]
\centering
\includegraphics[width=0.95\columnwidth]{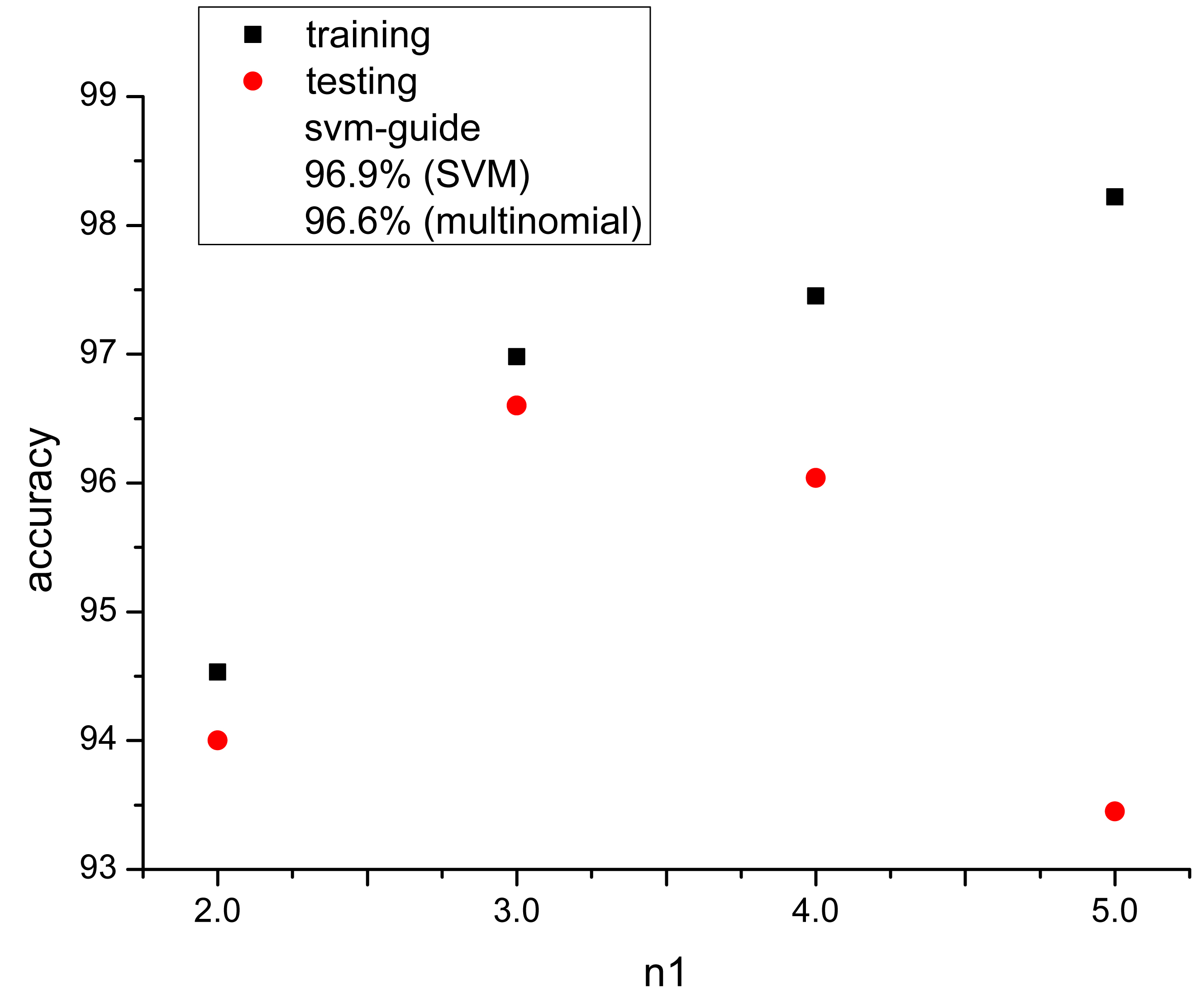}
\caption{(Color online.)
The accuracy and training set accuracy (the ability of the kernel to reproduce the training data) when using different order multinomial kernels (Eq. (\ref{multi})) in the SRVM algorithm when applied the svmguide1 data set. For comparison, we provide in the top panel, the optimal result found the SVM algorithm. 
}
\label{fig:figzerooo}
\end{figure}

Throughout, we used various kernels $K$ (Eq. (\ref{Klist})) when performing the expansions of Eq. (\ref{Map}). 
We further examined multinomial forms of particular maximum degrees $n_{k}$ in each single feature $x_{k}$,
\begin{eqnarray}
f^{\alpha}(\vec{x})= \sum_{m_{1}=0}^{n_{1}} \sum_{m_{2}=0}^{n_{2}}  \cdots \sum_{m_{d}=0}^{n_{d}} c^{\alpha}_{m_{1}m_{2} \cdots m_{d}} \prod_{k=1}^{d} x_{k}^{m_{k}}.
\label{multi}
\end{eqnarray}
Here, $\{ c_{n_{1}n_{2} \cdots n_{d}} \}$ are constants that may, similar to Eq. (\ref{Map}), be determined by Eq. (\ref{inverse-}) (the minimization of Eq. (\ref{Fmin})).
Such a form is natural if the predicted quantity is an analytic function of each of the features; analyticity is expected in physical systems in the absence of phase boundaries. Different replicas $\alpha$ may be associated with different orders $ \vec{n} \equiv (n_{1}, n_{2}, \cdots, n_{d})$.

For completeness, although we will not further explore it in the current work, we must underscore that, generally, there is, of course, nothing special about the simple decomposition of Eq. (\ref{multi}); one may replace the single features $x_{k}$ by any of their functions $g_{k}(x_{k})$ and consider the trivial generalization
\begin{eqnarray}
\label{multig}
f_{\vec{g}}^{\vec{n}} (\vec{x})= \sum_{m_{1}=0}^{n_{1}} \sum_{m_{2}=0}^{n_{2}}  \cdots \sum_{m_{d}=0}^{n_{p}} c^{\vec{n}}_{m_{1}m_{2} \cdots m_{d}} \prod_{k=1}^{d} (g_{k}(x_{k}))^{m_{k}}.
\end{eqnarray} 
Here, the subscript shorthand $\vec{g} \equiv (g_{1}, g_{2}, \cdots, g_{d})$. Different choices of $\vec{g}$ (for a given $\vec{n}$) lead to additional replicas. Supplanting the global forms of Eqs. (\ref{multi}, \ref{multig}), one may also readily construct other multinomial approximants to be additional replicas by taking $f^{\alpha}(\vec{x})$ to be tensor product splines of various generalized orders $\alpha$ \cite{tensor}. One may naturally also consider Laurent type multinomials (possibly also with different anchor points (i.e., shifted
coordinates (feature values) $x_{k} \to (x_{k}-\chi_{k})$ with constant $\{\chi_{k}\}_{k=1}^{d}$)) and, more generally, the Pad\'e type ratios 
\begin{eqnarray}
\label{Pade}
f_{\vec{g},\vec{g'}}^{\vec{n},\vec{n'}} (\vec{x}) =  \frac{\sum_{m_{1}=0}^{n_{1}} \sum_{m_{2}=0}^{n_{2}}  \cdots \sum_{m_{d}=0}^{n_{p}} c^{\vec{n}}_{m_{1}m_{2} \cdots m_{d}} \prod_{k=1}^{d} (g_{k}(x_{k}))^{m_{k}}} {\sum_{m'_{1}=0}^{n'_{1}} \sum_{m'_{2}=0}^{n'_{2}}  \cdots \sum_{m'_{d}=0}^{n'_{p}} c^{\vec{n}'}_{m'_{1}m'_{2} \cdots m'_{d}} \prod_{k'=1}^{d} (g'_{k'}(x_{k'}))^{m_{k'}}}.
\nonumber
\end{eqnarray} 
Given the above, disparate replicas $\alpha$ may be defined by the highest powers $(n_{1}, n_{2}, \cdots, n_{d};  n'_{1}, n'_{2}, \cdots, n'_{d})$ of the functions $\{g_{k}(x_{k})\}_{k=1}^{d}$ and $\{g'_{k'}(x_{k'})\}_{k'=1}^{d}$ appearing in the above ratio (as well as the choice of the functions $\{g_{k}(x_{k})\}_{k=1}^{d}$ and $\{g'_{k'}(x_{k'})\}_{k'=1}^{d}$).

We next explicitly turn to the ten examples that we tested. 

$\bullet$ The first test case is that of our own synthetic data that allow for a simple linear separation between two sets with 
non-intersecting convex hulls (the two sets appear in the upper right and lower left sides in Fig. \ref{fig:figzerozero}). The goal of the algorithm is to detect this structure and correctly classify different points as belonging to either of these two data sets. We used Eqs. (\ref{Map}, \ref{inverse-}) with a Gaussian kernel $K$ for $v=50$ fixed vectors $\{\vec{\chi}_j^\alpha\}$
that were randomly chosen for each of the $\mathcal{R}$ replicas; this led to an accuracy (as ascertained by the 5-fold CV) of 100\%. Fig. \ref{fig:figzerozero} illustrates the distribution of the two data sets and the boundary formed by the Gaussian Kernel SRVM algorithm. The boundary obtained by our method is a smooth surface- not a straight line as found by other class classification algorithms that we tested (e.g., SVM with a linear kernel, logistic regression, and other linear classifiers);
the linear kernel SVM algorithm similarly achieved an accuracy of $100\%$. 

In the remainder of this paper, we will focus on far more pertinent non-linearly separable problems and examine nine different benchmarks. 

$\bullet$ The next data set that we will test is that of the ``Four-class" \cite{ref:libsvm} benchmark- a binary classification problem having $d=2$ features for each of its 862 data points. Fig. \ref{fig:figzero} visually depicts the data on a $d=2$ dimensional map. Similar to our first example, the goal of the machine learning algorithm is to correctly identify the binary classification of input data (similarly set to be +1 (marked black in Fig. \ref{fig:figzero}) or -1 (red)). We obtained a perfect (i.e., $100\%$) accuracy when applying SVM with a radial kernel.  We studed this system with our SRVM method with the multinomial kernel of Eq. (\ref{multi}). Fig. \ref{fig:figzeroo} demonstrates how the prediction accuracy varies with the multinomial order $n_{1}(=n_{2} = \cdots = n_{d})$. In the tested range, is monotonic with increasing polynomial order. When the multinomial order $n_1 =11$, the accuracy is 100 \%.  Figures \ref{fig:figthree}, \ref{fig:figfive}, and \ref{fig:figseven} provide the boundaries found when $n_1$ equals 3, 5 and 7 respectively. Only the training points are shown in these figures. We see that when $n_1 =7$, a smooth boundary between the two classes results.  We similarly applied our algorithm with a Gaussian kernel $K$ to the Four-class problem. We first discuss the single replica results. The number of fixed vectors $v$ in Equation \ref{Map} plays an important role in predicting the results. We initially randomly produced $v=50$ fixed vectors (less than a tenth of the number of data set points). This led to an average accuracy of 
99.09\%. Reducing the number of fixed vectors to only $v=10$ resulted in an accuracy decrease to 81.18\%. In this and other instances, we saw that (not unexpectedly) when the number of fixed vectors became too small, the prediction accuracy diminished. In Section \ref{dependence+}, we will discuss this trend in greater depth.
As discussed in Section \ref{nut}, the SRVM combines the single replica results via voting (Eq. (\ref{average})). To avoid a gridlock when performing such a vote, we chose the number of replicas $\mathcal{R}$ to be an odd number (we picked $\mathcal{R} = 7$ here). Each replica $\alpha$ corresponds to a possible predictor $y^{\alpha}$ that is related to a different set of fixed vectors $\{\vec{\chi}_{j}^{\alpha}\}$. Averaging over replicas (Eq. (\ref{average})) produced an accuracy of $99.76\%$.

$\bullet$ Our subsequent test case is that of ``svmguide1'' benchmark \cite{ref:libsvm}. This well studied benchmark problem (originating from astroparticle physics) 
consists of training file and testing file (i.e., there is no need
to perform CV). The number of data points in training file and testing file are, respectively, 3089 and 4000; each data point has $d=4$ features. 
Optimizing and using the best parameters for a radial basis SVM kernel enabled a $96.9\%$ accuracy. We applied our SRVM algorithm with a polynomial kernel (see Fig. \ref{fig:figzerooo}) to this benchmark. Contrary to the Four-class problem, the accuracy initially grew with increasing polynomial order $n_{1}$; however, at larger $n_{1}$
the accuracy diminished. The peak prediction accuracy for the test data is $96.6\%$. In Section \ref{overlap+}, we will discuss how the best value of $n_{1}$ may be ascertained from replica overlap (without being given the results for the test data). We further applied the Gaussian kernel algorithm to the svmguide1 problem and tested three different value of number of fixed vectors ($v=50, 100, 200$). In single replica tests, the highest accuracy ($95.6\%$) was realized for $v=100$ fixed vectors. Setting $v=50,200$ gave rise to accuracies of $94.62\%$ and $94.98\%$ respectively. Using $\mathcal{R} = 7$ replicas in the Gaussian kernel algorithm, improved the accuracy to $95.8\%$.

$\bullet$ The ``Liver disorder'' data set \cite{ref:libsvm} is a benchmark problem that has 345 data points which has $d=6$ features for each input. It has no testing file so that we performed the CV tests as before. We first investigated the performance of SVM. Optimizing the SVM parameters in a radial basis enabled an average CV accuracy of $71.88\%$. Next, we applied the ($n_{1}= \cdots = n_{6}=3$) multinomial SRVM. This led to an average CV accuracy of $65.5\%$. Lastly, we applied the Gaussian kernel SRVM algorithm to the problem. We found the optimal number of fixed vectors is $v=80$. This led, for the single replica variant, to an accuracy of $66.29\%$. We then couple different replicas ($\mathcal{R} = 7, 15$, and $21$). The results illustrate that replica voting indeed improves the accuracy. Specifically, $\mathcal{R} = 7$ replicas led to an accuracy of $68.40\%$. In the case of $\mathcal{R} = 15 $ replicas, we achieved an accuracy 
of $66.97\%$. For $\mathcal{R} = 21$, the average CV accuracy became $68.40\%$.

$\bullet$ As another example, we also tested the Heart disease data set from the UCI machine learning repository database \cite{Heart}. This is a binary classification problem consisting of 270 data points with $d=13$ features. For calculations in this paper, the data will be scaled (to lie in the $[-1,1]$ interval). We will present various aspects
of our results for this prominent benchmark in later sections.

$\bullet$ The results from the Statlog Australian Credit Approval data set \cite{Australian} (hereby abbreviated to ``Australian'') will, similarly, also be presented.
This benchmark is comprised of 126 binary-classified instances with 309 features and, as we will demonstrate, possesses characteristics which make it an excellent representative data set. Similar to the Heart benchmark, the data presented for the Australian data set are also scaled such that each of the $d=309$ features spans the interval $[-1,1]$.

$\bullet$ An additional example on which we performed detailed analysis is that of LSVT voice rehabilitation data set \cite{LSVT}. This is a binary classification problem
in which each of the 126 instances has 309 features.  

$\bullet$ One more binary classification benchmark on which we tested our algorithm is that of ``Internet Advertisement Data Set'' \cite{ads}. 
This benchmark contains 3279 instances each of which has 1558 attributes. 

$\bullet$ Another data set that we examined was the ``IRIS'' flower data set \cite{IRIS}. This benchmark tabulates four features (the length and the width of the sepals and petals of the flowers) for three different types of irises. 

$\bullet$ The last benchmark on which we tried our SRVM algorithm was that of the ``Breast Cancer Wisconsin'' data set \cite{Wisconsin}. This is a binary classification problem. In this benchmark, given ten different (geometrical and texture) features of cell nuclei that are seen in a digitized image of a fine needle aspirate (FNA) of a breast mass), a tumor is to be classified as being benign or malignant. The original dataset contains a few points with missing features; these points were excluded from our study.

When analyzing data sets using classification or regression algorithms, it is important to begin by pre-processing the data to be studied. In many data sets, it is common to have various instances which are missing values corresponding to certain features. Numerous methods exist to deal with missing values through various types of imputation \cite{imputation1,imputation2}. Typically the act of imputing data for missing values is itself a learning step, which inherently adds complexity to the analysis process. In the data sets studied here, the number of instances with missing values was small enough that these instances were discarded. 

In addition to handling missing values, the pre-processing step also typically involves scaling of the data, so that the values corresponding to a given feature are of the same scale as all of the other features. This suppresses any effects of a feature with high variance and magnitude, dwarfing features with smaller variances and magnitudes. The three main feature scaling types are (i) scaling to the range [0,1], (ii) scaling to the range [-1,1], and  (iii) normalization of the range of values for a given feature such that they have a mean of zero and variance of one. In Section \ref{pre-p}, we will test whether there is any statistically significant difference in the performance of the algorithm with different feature scaling types.

Table \ref{tab:problemss} provides a synopsis of the accuracies obtained by SRVM method for the above common nine benchmarks that we examined. As seen therein, the accuracy of our algorithm was, on average, better than that of SVM by an insignificant margin.

\subsection{Accuracy Dependence on the Number of Replicas and Anchor Vectors}
\label{dependence+}

\begin{figure*}
	\centering
	\begin{subfigure}[t]{.45\textwidth}
		\centering
		\includegraphics[width=\linewidth]{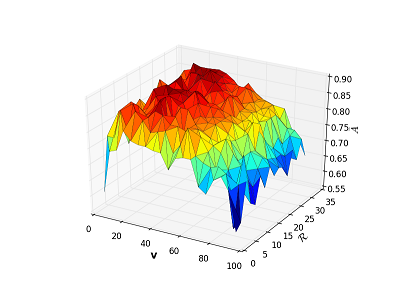}
		\caption{}\label{Acc:Reg_1}
	\end{subfigure}
	\begin{subfigure}[t]{.45\textwidth}
		\centering
		\includegraphics[width=\linewidth]{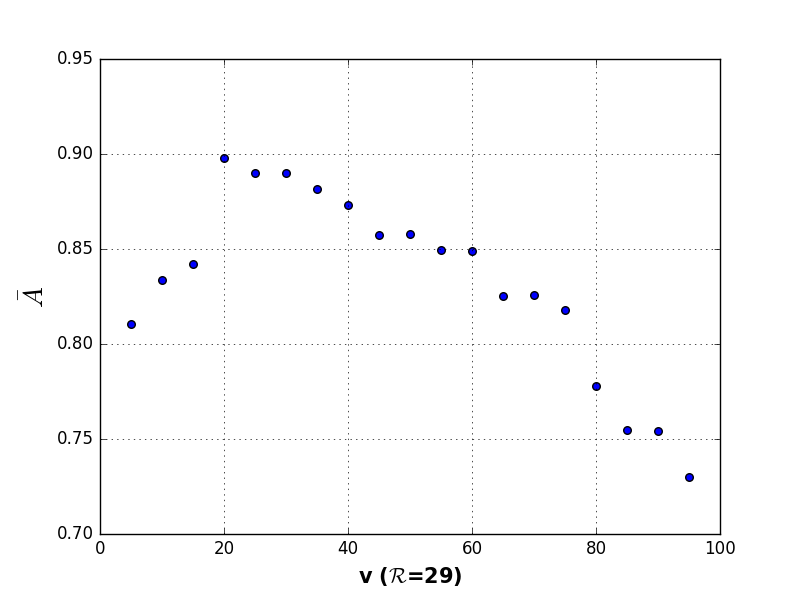}
		\caption{}\label{Acc:Reg_2}
	\end{subfigure}
	
	\medskip
	
	\begin{subfigure}[t]{.45\textwidth}
		\centering
		\vspace{0pt}
		\includegraphics[width=\linewidth]{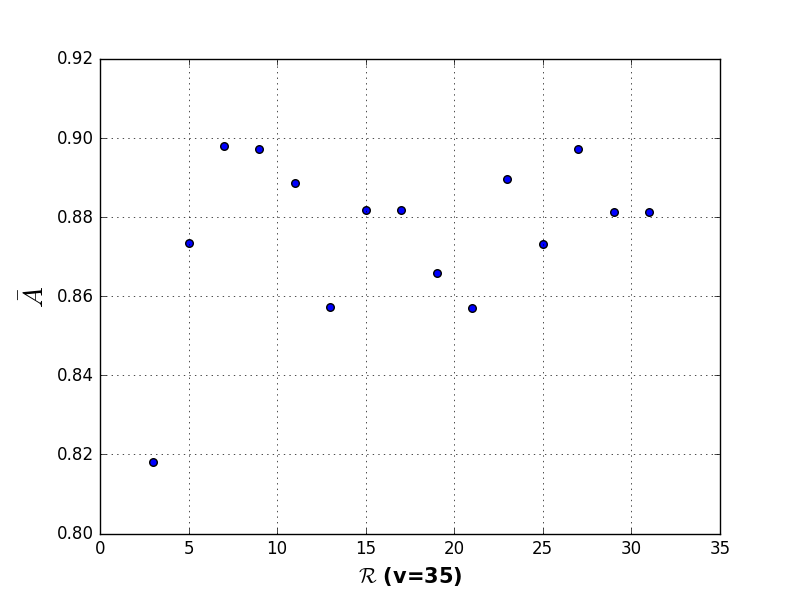}
		\caption{}\label{Acc:Reg_3}
	\end{subfigure}
	\begin{minipage}[t]{.45\textwidth}
		\caption{\label{fig:LSVTacc-anchor-repnum} (Color Online.) Tests of the accuracy of the SRVM algorithm (with a Gaussian kernel) to the LSVT data set. (a) 3D surface plot of the average accuracy (ascertained by 5 fold CV) as a function of the number of anchor points, $v$, and the number of replicas, $\mathcal{R}$. (b) The 2D projection of this 3D plot (a) into the accuracy-anchor point plane with constant replica number, $\mathcal{R}=29$ to show accuracy as a function of number of anchor points. The accuracy initially increases with more anchor points; beyond a threshold maximum value at $v = 20$, the accuracy drops (due to overfitting). (c) A projection of accuracy surface of (a) into the accuracy-replica plane with constant number of anchor points, $v=35$ in order to highlight the dependence of the accuracy on the number of replicas. The accuracy initially rises, very rapidly, with an increase of the number of replicas
		and then nearly saturates.}
	\end{minipage}
	
\end{figure*}

When evaluating the performance of a binary classification model, the first step is typically to measure the accuracy of the classifier when applied to the testing data of known labels. The accuracy is simply defined as the percentage of correctly labeled instances in the testing set. In analyzing the 
LSVT data set \cite{LSVT}, we primarily used the Gaussian kernel of Eq. (\ref{Klist}). A priori, the spread ($\sigma$) of this Gaussian may assume any value. We observed that setting $\sigma=\sqrt{N_{features}}$ yielded the best results. Consequently, this was the value used in our analysis. We employed the five-fold CV and examined the average accuracy, $\bar{A}$, across all five folds for various numbers of anchor points ($v$) and replicas ($\mathcal{R}$). The results of this analysis are presented in Fig. \ref{fig:LSVTacc-anchor-repnum}. In panel (\ref{Acc:Reg_1}), we show a 3D surface plot of the average accuracy as a function of the number of anchor points and number of replicas. In panels (\ref{Acc:Reg_2}) and (\ref{Acc:Reg_3}), we show projections of the 3D plot for constant $v$ and $\mathcal{R}$, respectively. It is evident from these plots that the accuracy quickly reaches an asymptotic value with increasing replica number. Once a maximum is reached, further changes in the number of replicas have little net impact on the accuracy. Additionally, it is evident that (regardless of the number of replicas $\mathcal{R}$ used) the accuracy increases rapidly with number $v$ of anchor points, levels off at a maximum, and then decays with further increasing $v$. The decay of average accuracy with increasing $v$ beyond a certain value is indicative of over-fitting. Analysis of the accuracy data presented in Fig. \ref{fig:LSVTacc-anchor-repnum} suggests that a maximum accuracy of $\bar{A}$=88.9$\%$ for the LSVT data set occurs at $v=35$ and $\mathcal{R}=29$.

\begin{figure*}
	\centering
	\begin{subfigure}[t]{.45\textwidth}
		\centering
		\includegraphics[width=\linewidth]{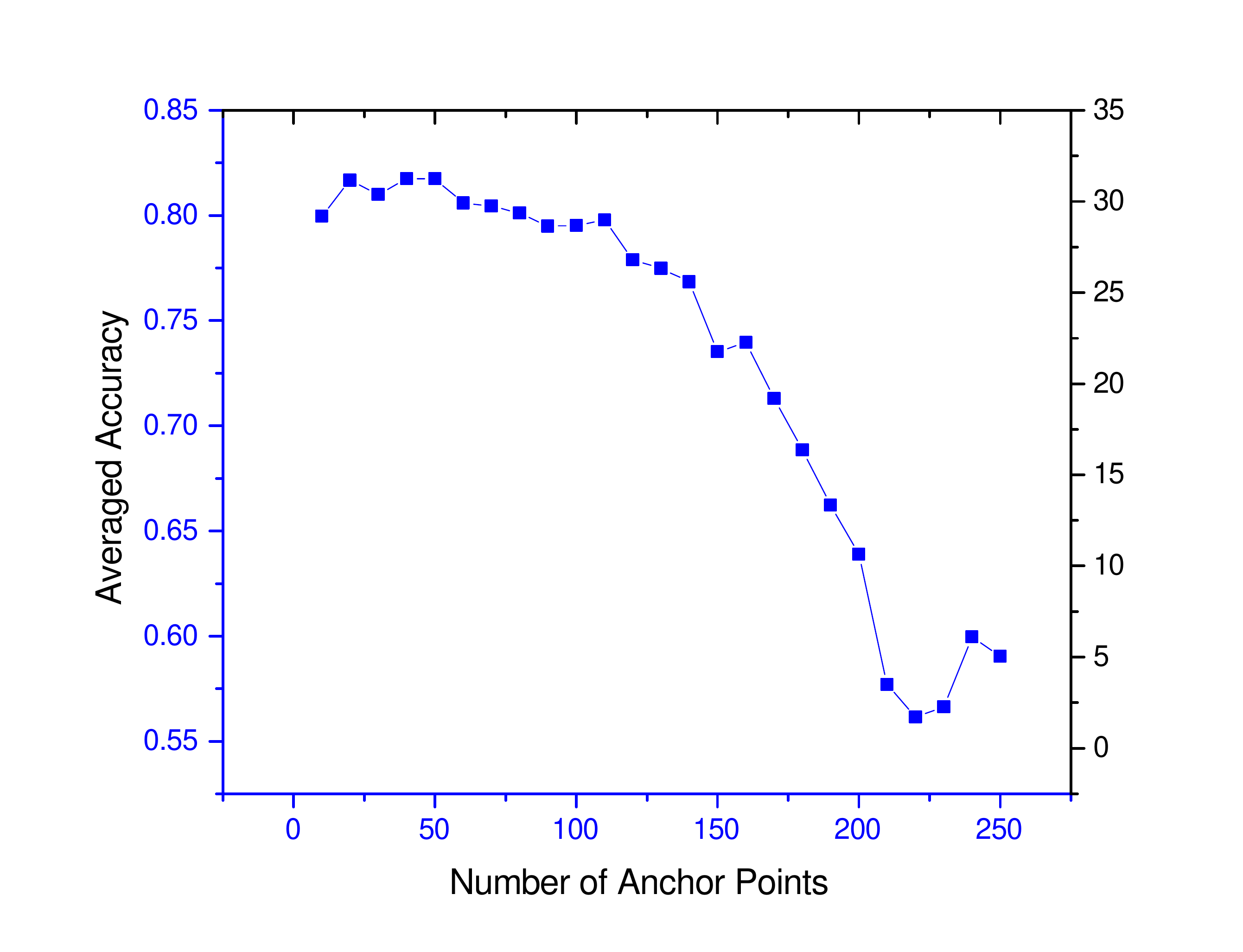}
	\end{subfigure}
	\begin{subfigure}[t]{.45\textwidth}
		\centering
		\includegraphics[width=\linewidth]{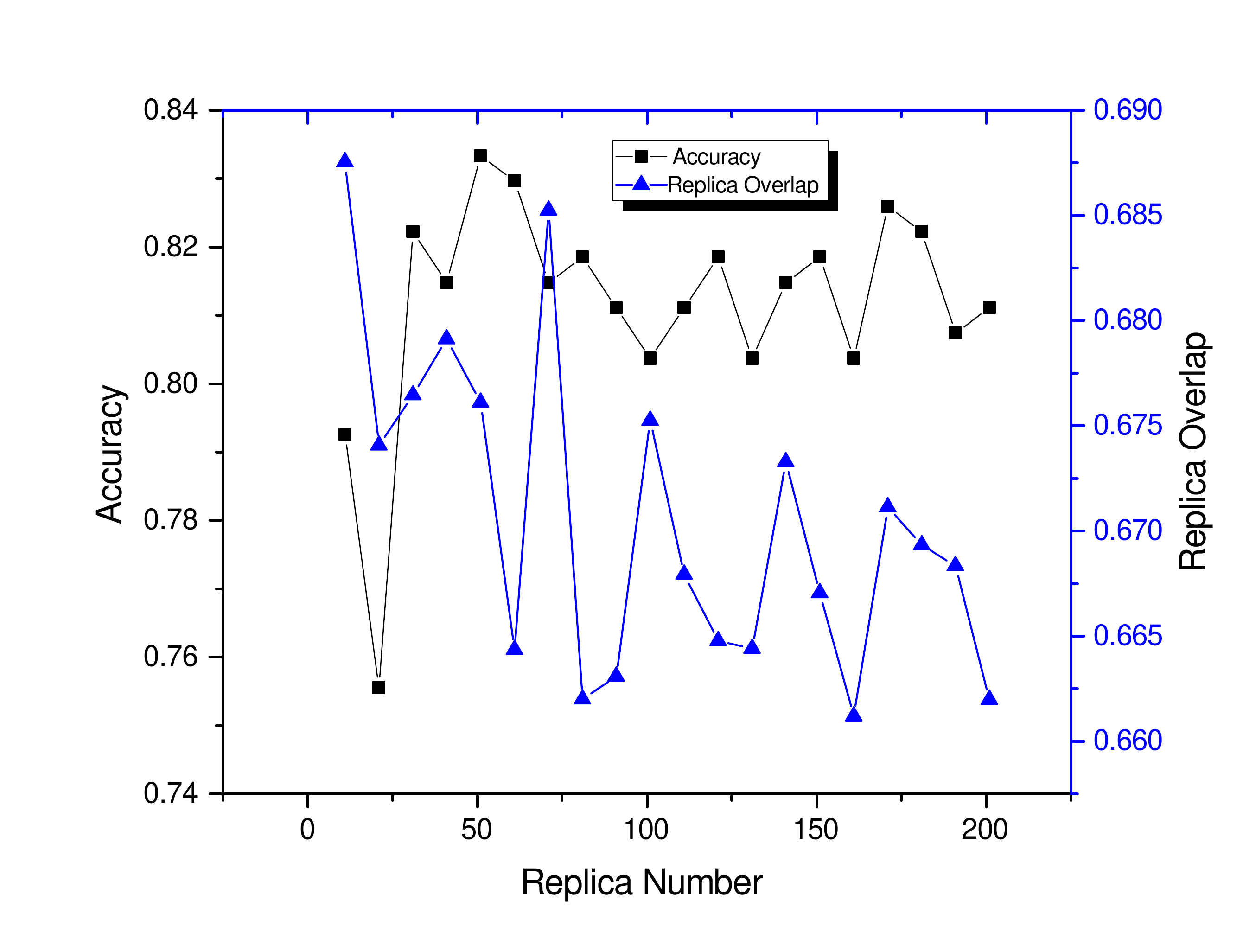}
	\end{subfigure}

		\caption{\label{fig:Heartacc-anchor-repnum} (Color Online.) Accuracy of the SRVM algorithm (with a Gaussian kernel) when applied to the Heart data set. (a) Graph of the average accuracy as a function of the number of anchor points $v$ and number of replicas $\mathcal{R} = 31$. (b) Plot of the average accuracy as a function of the number of replicas, $\mathcal{R}$ when the number of anchor points is held fixed at $v = 50$. In Section \ref{overlap+} we will define and analyze
		inter-replica overlaps (the one plotted here is the normalized variant of Eq. (\ref{O2})). As seen here, the 
		average replica overlaps correlate with the accuracy of the predictions.}	
\end{figure*}

\begin{table*}[t]
	\begin{tabular}{| c | c | c | c | c | c |}
		\hline
		Data Set & Classes & Number of Instances & Number of Features & SVM & SRVM\\ 
		\hline
		LSVT & 2 & 126 & 309 & 0.873 & 0.889\\
		Advertisement & 2 & 3279 & 1558 & 0.973 & 0.963\\
		Iris & 3 & 150 & 4 & 0.927 & 0.967\\
	        Australian & 2 & 690 & 14 & 0.853 & 0.863\\
		Heart & 2 & 270 & 13 & 0.825 & 0.824\\
		Four-class & 2 & 862 & 2 & 1.00 & 1.00\\
		svmguide1 & 2 & 3089 & 4 & 0.969 & 0.966\\
		Liver-disorders & 2 & 345 & 6 & 0.718 & 0.684\\
		Breast-cancer & 2 & 683 & 10 & 0.942 & 0.945\\
		\hline
	\end{tabular}
	\caption{Summary of the Optimized Accuracy for both (a) the standard SVM algorithm (after finding the best parameters for the different data sets) and (2) our SRVM algorithm for three different data sets of varying class and instance number. Generally, the accuracies for both methods are comparable. The virtue of the SRVM method (apart from being systematically able to detect optimal parameters by examining the inter-replica overlap) is that the SRVM suffers from far less data bias that SVM; we will expand on this in Section \ref{sec:imbalance} (and Table \ref{tab:problems+} therein). 
	The average accuracy of the SRVM algorithm (0.900) is slightly higher than that of SVM (0.898)
	yet this difference is not statistically meaningful.}
	\label{tab:problemss}
\end{table*}

\begin{figure*}
	\centering
	\begin{subfigure}[t]{.45\textwidth}
		\centering
		\includegraphics[width=\linewidth]{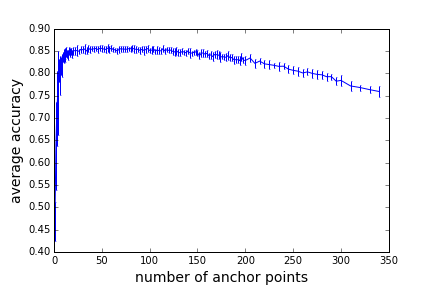}
	\end{subfigure}
	\begin{subfigure}[t]{.45\textwidth}
		\centering
		\includegraphics[width=\linewidth]{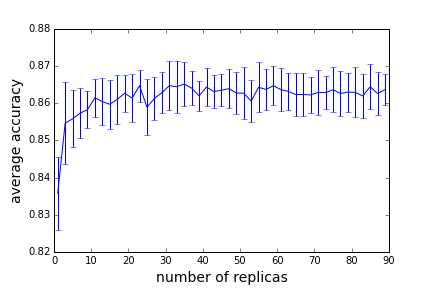}
	\end{subfigure}

		\caption{\label{fig:AUSTacc-anchor-repnum} (Color Online.) Accuracy tests for the Australian data set. We show a (a) Plot of average accuracy as a function of the number of anchor points $v$ for a fixed small number of replicas ($\mathcal{R} = 5$), and (b) plot the average accuracy as a function of the number of replicas, $\mathcal{R}$ for $v = 50$ anchor points.}

\end{figure*}

We now turn to a similar analysis for the ``Heart''  data set \cite{Heart}. For simplicity, we set the value of $\sigma$ to unity. In order to find the optimal number of anchor points for these data, we increased the number of anchor points $v$ from 10 to 250 in increments of 10 (see panel (a) of Figure
\ref{fig:Heartacc-anchor-repnum}). The resulting accuracy was averaged over 10 different sets of $\mathcal{R} = 31$ replicas analyzed with a 5 fold CV. The highest accuracy was achieved when $v=40$. A further minimum in the accuracy appears for $v \sim 220$ anchor points. For anchor points as low in number as $v=12$, our procedure yields an accuracy above 80\% (a value quite close to the highest obtained accuracy of 82\% that we obtained when using $v=50$ anchor points).
In panel (b) of  Figure \ref{fig:Heartacc-anchor-repnum}, we show the effect of increasing replica number on the average accuracy in Heart example. The range of the number of replicas is quite wide, $11 \le \mathcal{R} \le 201$. Both curves in this panel (corresponding to the average accuracy and the replica overlap) display an oscillatory behavior about the averaged result and the amplitude of oscillations decreases as the number of replicas $\mathcal{R}$ increases. 
Already for $\mathcal{R}=31$ replicas, we achieved an average accuracy of 82\%. Considering that the highest accuracy the we reached (as is seen in the graph) is 
83.3\% for $\mathcal{R} = 51$ replicas, in further analysis of the Heart data set, we used the more modest number of $\mathcal{R} =31$ replicas.

An important point that we will underscore and reiterate throughout this work (and discuss, more specifically, in Section \ref{overlap+}), 
is that we may determine the optimal number of replicas $\mathcal{R}$,
number of anchor points $v$, and any other undetermined quantity by noting when the average inter-replica is (near) maximal 
as a function of these parameters. 

We return to our analysis of the Australian data set. The dependence of accuracy on number of anchor points is tested on the Australian data set with Gaussian kernel models with replica number $\mathcal{R} = 5$. Each point of the plot is the average of 20 randomly generated models; see Fig. \ref{fig:AUSTacc-anchor-repnum}(a).

We observe that as the number $v$ of fixed vectors is increased, initially the fitted model becomes more sophisticated and the prediction accuracy rises rapidly. This shows that the model can be quite accurate even with a low number of fixed vectors. Beyond a certain point, increasing the number of fixed vectors $v$ starts leading to over-fitting and the prediction accuracy drops, however the drop is rather gradual, indicating that the model is robust against overfitting. 

The dependence of the accuracy on the replica number $\mathcal{R}$ was tested in the Australian data set by performing 50 five-fold CVs and taking the average accuracies across the SRVM results with $v = 50$ anchor points for the Gaussian kernel and investigating the results when the number of replicas $\mathcal{R}$ was varied from 1 to 89. The results are plotted in Fig. \ref{fig:AUSTacc-anchor-repnum}(b).

In addition to assessing the accuracy of the SRVM algorithm, it is important to compare its performance to established learning algorithms and to try and quantify any relative advantages and/or deficiencies. To that end, as we noted earlier, we took the Support Vector Machines (SVM) algorithm \cite{SVM,SVM1} as a baseline for comparison. For the LSVT data set, we used a `brute force' method of finding the optimal parameters for this contender to our method- the SVM model- by running it for all values in a grid in parameter space. Once the optimal parameters were found, it was observed that the maximum accuracy for SVM was 0.873. The difference in accuracy between our SRVM method (in which optimized parameters were found by replica overlap not by comparing to the solution) and the standard SVM algorithm (now optimized to achieve highest accuracy) is 0.016.  This difference is not statistically significant, so the relative advantage of either method might not be immediately clear. 

\begin{figure*}
	\centering
	\begin{subfigure}[t]{.45\textwidth}
		\centering
		\includegraphics[width=\linewidth]{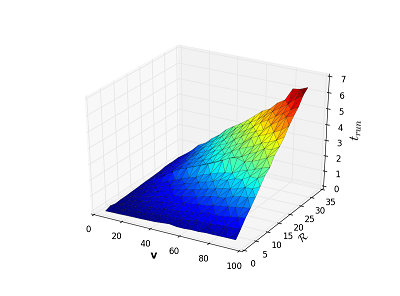}
		\caption{}\label{Run:Reg_1}
	\end{subfigure}
	\begin{subfigure}[t]{.45\textwidth}
		\centering
		\includegraphics[width=\linewidth]{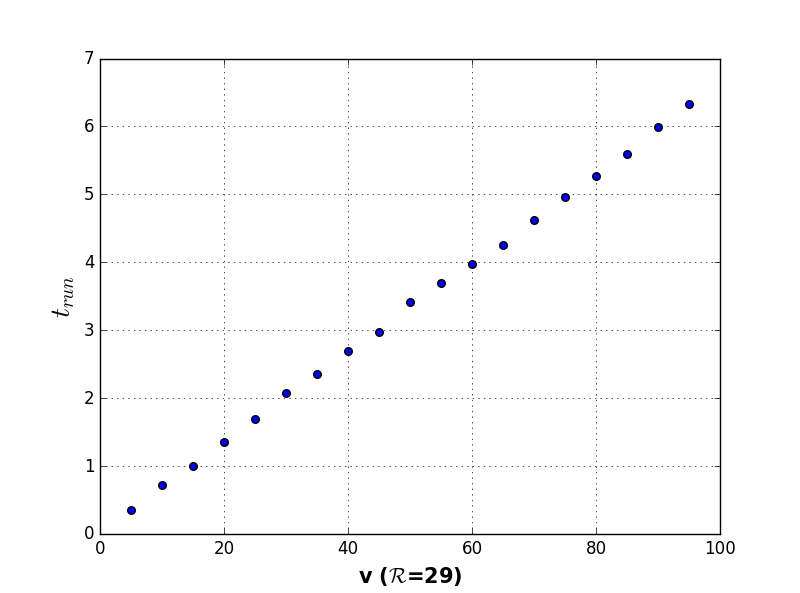}
		\caption{}\label{Run:Reg_2}
	\end{subfigure}
	
	\medskip
	
	\begin{subfigure}[t]{.45\textwidth}
		\centering
		\vspace{0pt}
		\includegraphics[width=\linewidth]{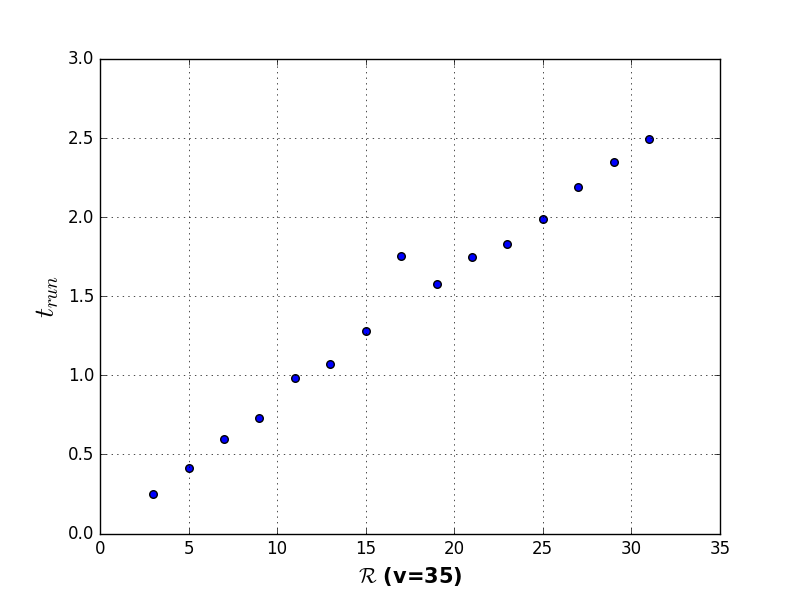}
		\caption{}\label{Run:Reg_3}
	\end{subfigure}
	\begin{minipage}[t]{.45\textwidth}
		\caption{\label{Run}(Color Online.) LSVT data set. (a) 3D surface plot of SRVM runtime as a function of the number of anchor points $v$ and number of replicas, $\mathcal{R}$. (b) Projection of runtime surface of (a) into the accuracy-anchor point plane with constant replica number, $\mathcal{R}$=29 to show runtime as a function of number of anchor points. (c) Projection of runtime surface of (a) into the accuracy-replica plane with constant number of anchor points, $v$=35 to show runtime as a function of the number of replicas. It is clear that in both cases, the runtime scales linearly with both parameters, allowing for prediction of runtime from from small parameter samples. If it is known that many coefficients (a single coefficient is associated with each anchor point) will be needed in Eq. (\ref{Map}) then one may estimate the requisite runtime of anchor points from the known runtime from smaller $v$.}
	\end{minipage}	
\end{figure*}

To dig further into the comparison, while simultaneously exploring the SRVM performance on a deeper level, we next examine the runtime of both the SVM and SRVM algorithms. We ran the SRVM algorithm on the LSVT data set for various values of $v$ and $\mathcal{R}$. The runtime is considered to be the time that it takes to calculate the average CV accuracy, and does not include finding the optimal number of parameters, or pre-processing steps. Figure (\ref{Run:Reg_1}) shows a 3D surface plot of the runtime versus the number of anchor points and replicas. In figures (\ref{Run:Reg_2}) and (\ref{Run:Reg_3}), the runtime is exhibited as a function of the number of anchor points $v$  and the number of replicas $\mathcal{R}$. The data make clear that the runtime increases linearly with increasing $v$ and $\mathcal{R}$. This observation suggests that it is possible to find the runtime at low numbers of both variables in order to assess how long a run will take with larger values. 

\begin{figure*}
	\centering
	\begin{subfigure}[t]{.45\textwidth}
		\centering
		\includegraphics[width=\linewidth]{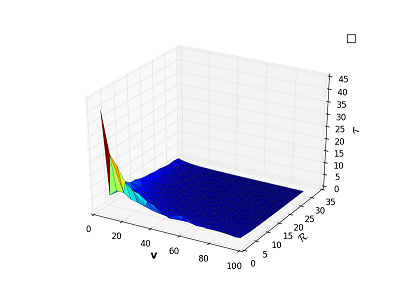}
		\caption{}\label{Cop:Reg_1}
	\end{subfigure}
	\begin{subfigure}[t]{.45\textwidth}
		\centering
		\includegraphics[width=\linewidth]{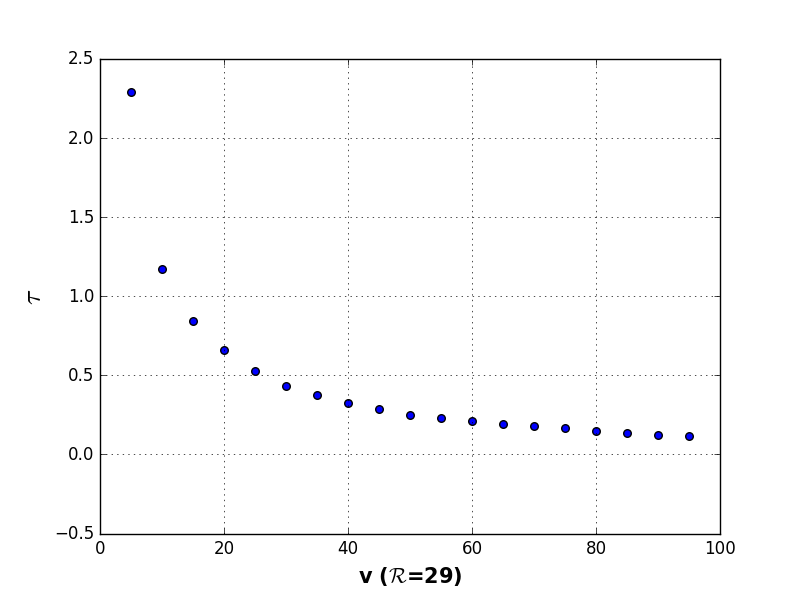}
		\caption{}\label{Cop:Reg_2}
	\end{subfigure}
	
	\medskip
	
	\begin{subfigure}[t]{.45\textwidth}
		\centering
		\vspace{0pt}
		\includegraphics[width=\linewidth]{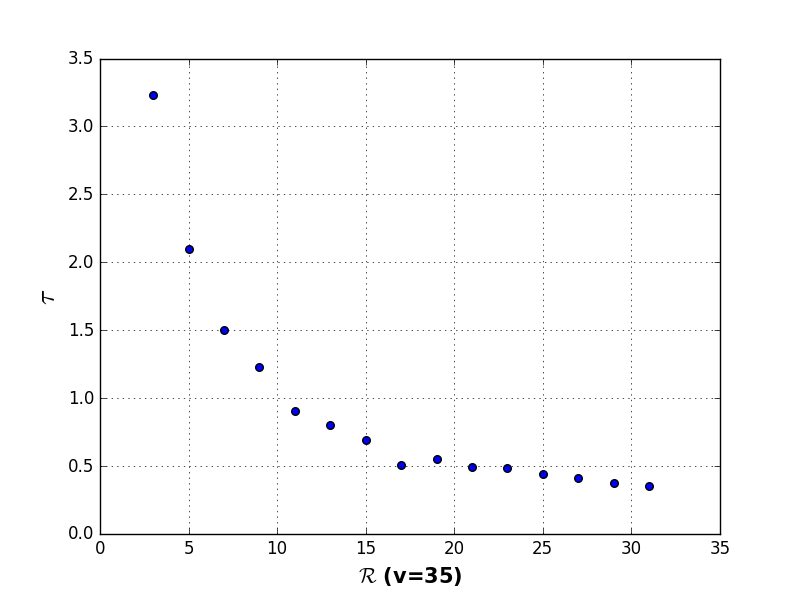}
		\caption{}\label{Cop:Reg_3}
	\end{subfigure}
	\begin{minipage}[t]{.45\textwidth}
		\caption{(Color Online.) Analysis of the LSVT data set. (a) 3D surface plot of SRVM Coefficient of Performance (COP) of Eq. (\ref{COP}) as a function of the number of anchor points $v$ and number of replicas $\mathcal{R}$. (b) Projection of COP surface of (a) into the accuracy-anchor point plane with constant replica number, $\mathcal{R}$=29 to show COP as a function of number of anchor points. (c) Projection of COP surface of (a) into the accuracy-replica plane with constant number of anchor points, $v$=35 to show COP as a function of the number of replicas. As they trivially must, the trends for the COP of Eq. (\ref{COP}) in all panels encapsulate the behavior of both the accuracy (i.e., they combine the results displayed in Fig. (\ref{fig:LSVTacc-anchor-repnum}) along with the (near linear) run time (Fig. (\ref{Run})) dependence on $v$ 
		and $\mathcal{R}$).}
	\end{minipage}
	\label{Cop.}
\end{figure*}

The general optimization of model performance involves maximizing accuracy while simultaneously minimizing the necessary runtime. Therefore, it would be beneficial to have a measure of the compounding of these two goals. To assess the intersection of accuracy and run time, we can define a metric which we call the coefficient of performance, $\tau$, which we define as
\begin{eqnarray}
{\tau\equiv\frac{\bar{A}}{t_{run}}}.
\label{COP}
\end{eqnarray}
This metric allows for an efficient via for simultaneously looking at optimal accuracy and run time. Using the results of the runtime and accuracy measures discussed above for the LSVT data set, we calculated the values of $\tau$ as a function of $v$ and $\mathcal{R}$. Detailed results highlighting various aspects are shown in 
panels (\ref{Cop:Reg_1},\ref{Cop:Reg_2},\ref{Cop:Reg_3}).

It is clear that the COP decays as a function of both the number of anchor points $v$ and the number of replicas $\mathcal{R}$. This is consistent with the linearity of the runtime and the asymptotic behavior of the accuracy. Locating the `knee'  in the COP data allows for extracting reasonable values of the parameters for the trade-off between accuracy and runtime.

\begin{figure*}
	\centering
	\begin{subfigure}[t]{.45\textwidth}
		\centering
		\includegraphics[width=\linewidth]{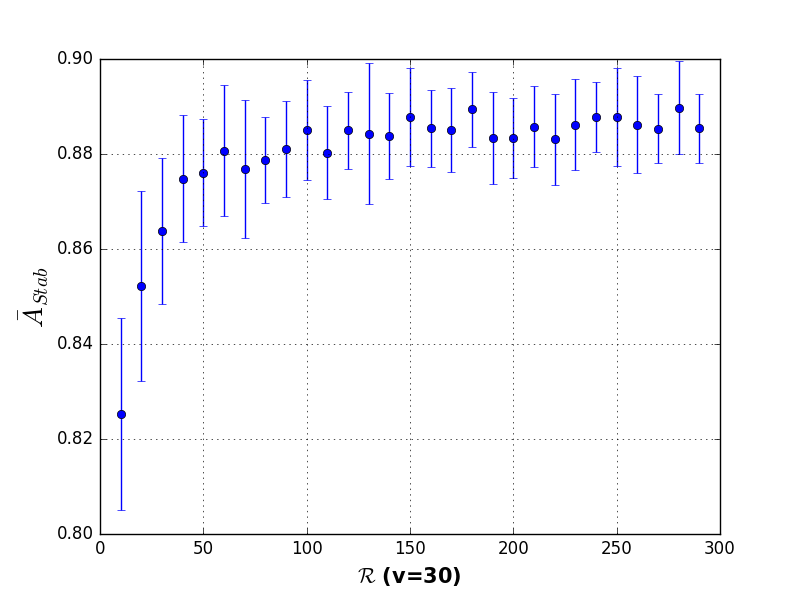}
	\end{subfigure}
	\begin{subfigure}[t]{.45\textwidth}
		\centering
		\includegraphics[width=\linewidth]{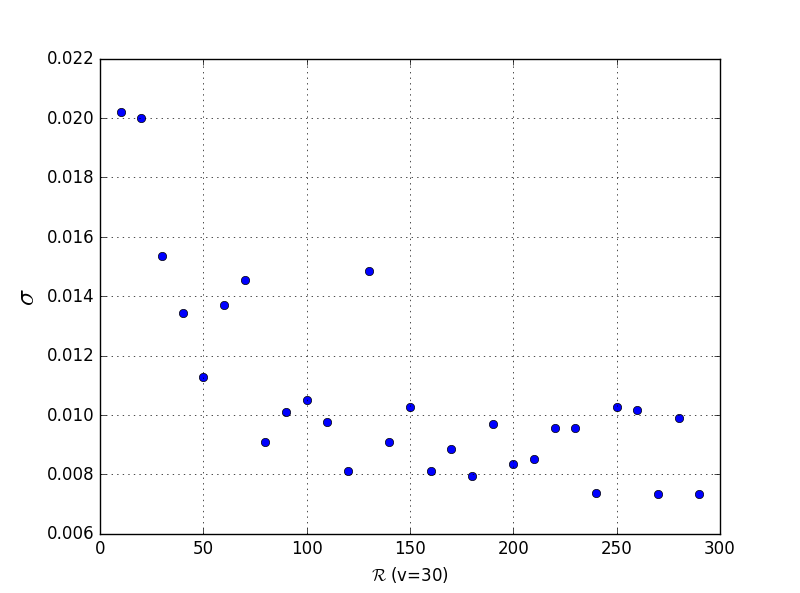}
	\end{subfigure}

		\caption{\label{Stab.} (Color Online.) LSVT data set. A defining feature of the SRVM is that each replica is stochastically generated (leading, a priori, to different results.) To that add end, (a) we display the average accuracy with $v=30$ anchor points for variable numbers of replicas, simulated 20 times each with different random replica generation seeds in each simulation. Associated standard deviations are shown as error bars. (b) Plot of these standard deviations in the accuracy that are associated with runs for various replica numbers. The monotonic decrease in the standard deviation with increasing number of replicas demonstrates that prediction results become more stable with increasing replica number $\mathcal{R}$; when additional replicas (for an increasing yet still small $\mathcal{R}$) vote, the final outcome becomes progressively more stable to statistical fluctuations from the stochastic generation of the anchor point vectors. The plot further makes clear that the standard deviation quickly reaches a leveling-off point at which further replica increase does not have a statistically significant impact on stability.}
\end{figure*}

In addition to calculating COPs for the SRVM algorithm, we also calculated them for SVM. Broadly, the SVM COP is considerably better than that of SRVM, and this is entirely due to the fact that SVM runs much quicker. This is likely due to the fact that the SVM algorithm has been highly streamlined and optimized in various software packages over the decades, whereas our algorithm is new.  It could also be due to the fact that, within all implementations of our algorithm, we computed the pseudo inverse exactly (Eq. (\ref{inverse-})) instead of approximating it with methods such as gradient descent. Furthermore, we have not employed other methods such as regularization to increase the accuracy with all of the other parameters fixed. Additional  improvements of the algorithm structure will likely decrease the runtime.

\subsection{Stability}
\label{sec:stable.}

A central characteristic of the SRVM algorithm is the use of voting between replicas to increase the accuracy. Intuitively, one would expect that the number of replicas and the accuracy should be positively correlated: the use of more replicas leads to improved accuracy. 
Another quintessential feature of the SRVM is that the $v$ anchor vectors associated with each individual replica are generated stochastically. This allows for a robust classification of new instances. This also means that each run of the algorithm will be different, with different outcomes possible. Therefore, it is important to examine the stability of the output. It is expected that for a low number of replicas ($\mathcal{R}$), the overall vote can change rather dramatically with different runs, so the accuracy can fluctuate. It is further expected that as the number of replicas increases, the fluctuations will be suppressed by the presence of more information in the overall vote. To test this, we ran the SRVM algorithm on the LSVT data set with $v$=30 anchor vectors per replica 20 times each, for varying number of replicas. In panel (a) of Fig. (\ref{Stab.}), we display the average accuracy across all 20 runs with a fixed number of replicas as this number (${\mathcal{R}}$) increases. The error bars in the figure reflect the standard deviation in accuracy. In panel (b) of Fig. (\ref{Stab.}), we plot the standard deviation in accuracy versus number of replicas. From the panels of Fig.  (\ref{Stab.}), it is clear that the standard deviation decreases rapidly with increasing number of replicas and eventually levels off to a roughly constant value. This is consistent with the earlier observation that the runtime is linear and the accuracy approaches a leveling off before decreasing. Further, the result implies that beyond a certain number of replicas, the overall accuracy is largely stable to fluctuations associated with stochastic generation of anchor vectors, thus alleviating a potential weakness of the method.

Tie stability was further tested using the Australian data set by performing 50 five-fold CVs and computing the average accuracies across models with replica numbers ranging from 1 to 89. The results are provided in Fig. \ref{fig:AUSTacc-anchor-repnum}. As this figure makes evident, for this data set, the average accuracy rises relatively quickly at the beginning from just one replica and maintains a general monotonic trend as the replica number increases. We begin to observe diminishing returns somewhere after 15 replicas. This is to be expected, as the amount of available information in the data set is objectively limited so there is a cap on achievable accuracy. Note that since we do a simple majority vote, the replica numbers are all odd to ensure that no ties appear during voting. Another example where we tested the accuracy as a function of the number of replica is shown in Fig. \ref{fig:Heartacc-anchor-repnum}(b) for the Heart benchmark.  As seen therein, the accuracy and replica overlap achieve their maximal values when we used ${\mathcal{R}} =31$ replicas. One may expect that as the number $v$ of fixed vectors increases, initially the fitted model becomes more sophisticated and the prediction accuracy rises. Above a certain value, increasing the number of fixed vectors starts leading to over-fitting and the prediction accuracy drops. Using more fixed vectors also results in a slower algorithm. Therefore it would be very useful if we had a way of estimating how many fixed vectors are appropriate for a certain problem.  In Fig. \ref{fig:AUSTacc-anchor-repnum}(a), we notice that the curves for both the average accuracy and the average replica overlap rise rapidly from their values for a single fixed vector ($v=1$) to a nearly flat maximum that appears when the number of around fixed vectors $20 \lesssim v   \lesssim100$; when $v \gtrsim 100$, the accuracy begins to taper off due to the alluded to overfitting. The two curves indeed follow each other closely, supporting  the notion of using replica overlap to estimate the dependence  of the expected accuracy on $v$. We found similar behaviors for other parameters other than the anchor vector number $v$. There are some other outstanding features of the figure: the rapid rise of the two curves at low fixed vector number shows that the model can be quite accurate even with low fixed vector number, and the slow tapering off of the two curves indicates that the model is robust against overfitting.

\subsection{Impact of pre-processing}
\label{pre-p}

In the beginning of this section, we alluded to the possibility that the specific pre-processing method employed may have an impact on the performance of the SRVM algorithm. In this subsection, we will examine the impact of pre-processing the data using feature scaling on our final results. To that end, 
we preprocessed the data for the LSVT data set in three different ways: \newline

$\bullet$ (1) Linearly transforming the data such that domain of each feature over the entire data set ranges from $0$ to $1$. \newline

$\bullet$ (2) Linearly transforming the data such that each feature assumes values in $[-1,+1]$ (i.e., scaling the data to have a difference of two between the maximal and minimal value of each feature),  and \newline

$\bullet$ (3) Normalizing the scaled data with mean $0$ and standard deviation equal to unity.  \newline

We examined the average accuracies (and their variances) associated with these three different pre-processing methods using statistical tests. 
 The results (see table \ref{tab:problems}) demonstrate that one must absolutely reject the null hypothesis 
$H_{0}$ that all of the means are equal. The disparate pre-processing methods definitely lead to different results. 
The specific testing of the means were performed both (i) assuming normal distribution of the averages (the f-statistic) and (ii) without this assumption (the h-statistic using Kruskall Wallis test \cite{H}). Both tests revealed that the average accuracy was not statistically uniform across all methods of feature scaling. To quantitatively investigate which methods were intrinsically different from one another, individual t-tests of the means were undertaken. We performed three different t-tests \cite{student-t-test}. These tests demonstrated that there is
no difference between pre-processings to the types (1) and (2). However, these two cases however are different from the normalization (pre-processing type (3)). The normalization pre-processing tends to be the most accurate of the three for lower numbers of anchor points ($v$) and 
replicas $(\mathcal{R})$. In general, the data sets that were scaled normally (pre-processing (3)) had higher accuracy with lower numbers of anchor points. We further tested that the variances were equivalent using a Levene test \cite{Levene};  the results indicated no statistically significant difference between the variances across all pre-processing methods employed. Taken together, these outcomes indicate that one must consider the specific type of pre-processing undertaken when assessing the performance of the SRVM algorithm. 
\begin{table}[t]
	\begin{tabular}{| c | c | c | c |}
		\hline
		Null Hypothesis & Test Statistic & p-Value & Conclusion  \\ 
		\hline
		$\mu_1$=$\mu_2$=$\mu_3$ & f=4.4542 & p=0.0159 & Reject H$_0$ \\
		$\tilde{x}_1$=$\tilde{x}_2$=$\tilde{x}_3$ & h=8.9006 & p=0.0116 & Reject H$_0$ \\
		$\sigma^2_1$=$\sigma^2_2$=$\sigma^2_3$ & w=2.1240 & p=0.1289 & Fail to Reject H$_0$   \\
		$\mu_1$=$\mu_2$ & t=2.7440 & p=0.0092 & Reject H$_0$ \\
		$\mu_1$=$\mu_3$ & t=2.7440 & p=0.0092 & Reject H$_0$ \\
		$\mu_2$=$\mu_3$ & t=0 & p=1.0 & Fail to Reject H$_0$ \\
		\hline
	\end{tabular}
	\caption{LSVT data set. Results of statistical hypothesis tests undertaken to assess whether different pre-processing techniques impact algorithm accuracy for the same sets of parameters ($v$ and $\mathcal{R}$). Here, $H_{0}$ denotes the hypothesis that all of the three pre-processing methods yield identical results.}
	\label{tab:problems}
\end{table}

\subsection{Optimization via replica overlap metrics}
\label{overlap+}
When choosing the optimal values of the parameters for a learning algorithm, it is helpful to have a reference function which does not require the calculation of the accuracy, which still relays information about model performance. This gives a more `fair' way of choosing the best values of the parameters without a brute force method. In the SRVM algorithm, because we have many replicas which are voting together by a simple majority vote, it seems reasonable that some measure of the overlap between the replicas would be a measure of model performance. Indeed, when all of the replicas are largely in agreement, it should imply the model is performing optimally and vice versa. However, it is possible that all of the replicas could be in agreement, with all of them being incorrect. Therefore, it is important to test whether proposed replica overlaps are agreement with the accuracy. To test this, we propose two different replica overlap functions, and test them on the LSVT data set. The first overlap function is defined as

\begin{figure}
	\centering
	\includegraphics[width=1.8 \columnwidth, height= .3 \textheight,keepaspectratio]{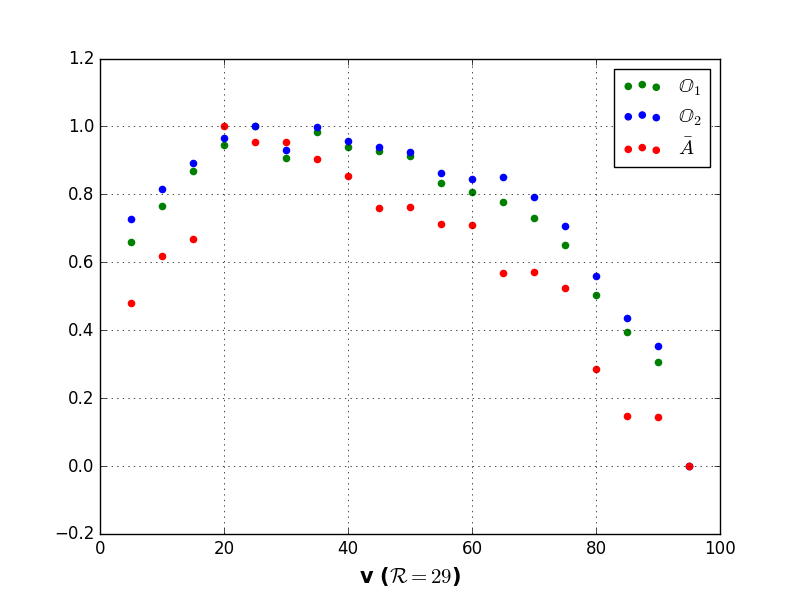}
	\caption{(Color Online.) LSVT data set. Plot of the two overlaps $O_{1}$ and $O_{2}$ (each of which is now scaled by their respective maximum value) of accuracy and the accuracy as a function number of anchor points $v$ for $\mathcal{R}$=29 replicas. It is observed that both overlap (associated with almost identical numerical values) scale with the accuracy
	of the predictions. As in the other examples that we studied, this correlation (and others like it) illustrates that instead of having to rely on exact calculations of the accuracy one may use the overlaps to ascertain the optimal values of the parameters defining the SRVM model (in this case, the optimal number of anchor points $v$ and the number
	of replicas $\mathcal{R}$).}
	\label{AO.}	
\end{figure}
\begin{figure*}
	\centering
	\begin{subfigure}[t]{.45\textwidth}
		\centering
		\includegraphics[width=\linewidth]{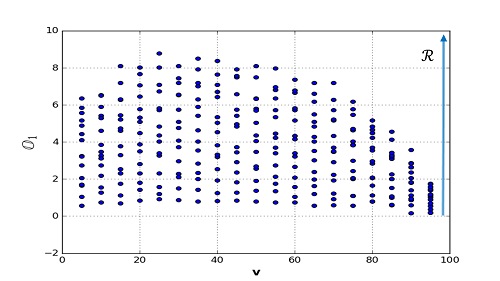}
	\end{subfigure}
	\begin{subfigure}[t]{.45\textwidth}
		\centering
		\includegraphics[width=\linewidth]{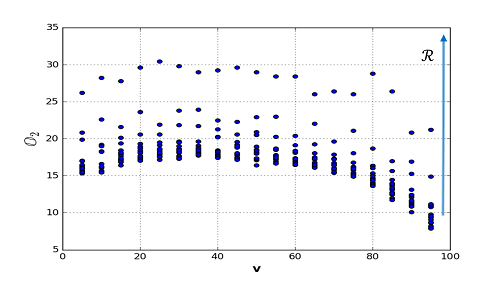}
	\end{subfigure}
		\caption{ \label{1z} (Color Online.) 
		 LSVT data set. Plot of the overlap functions defined in Eqs. (\ref{O1},\ref{O2}) as a function of the number of anchor points $v$ for various numbers of fixed replica numbers. The replica number increases (in increments of two from 1 to 31) along the vertical axis.}
\end{figure*}
\begin{figure}[h]
\centering
\includegraphics[width=0.95\columnwidth]{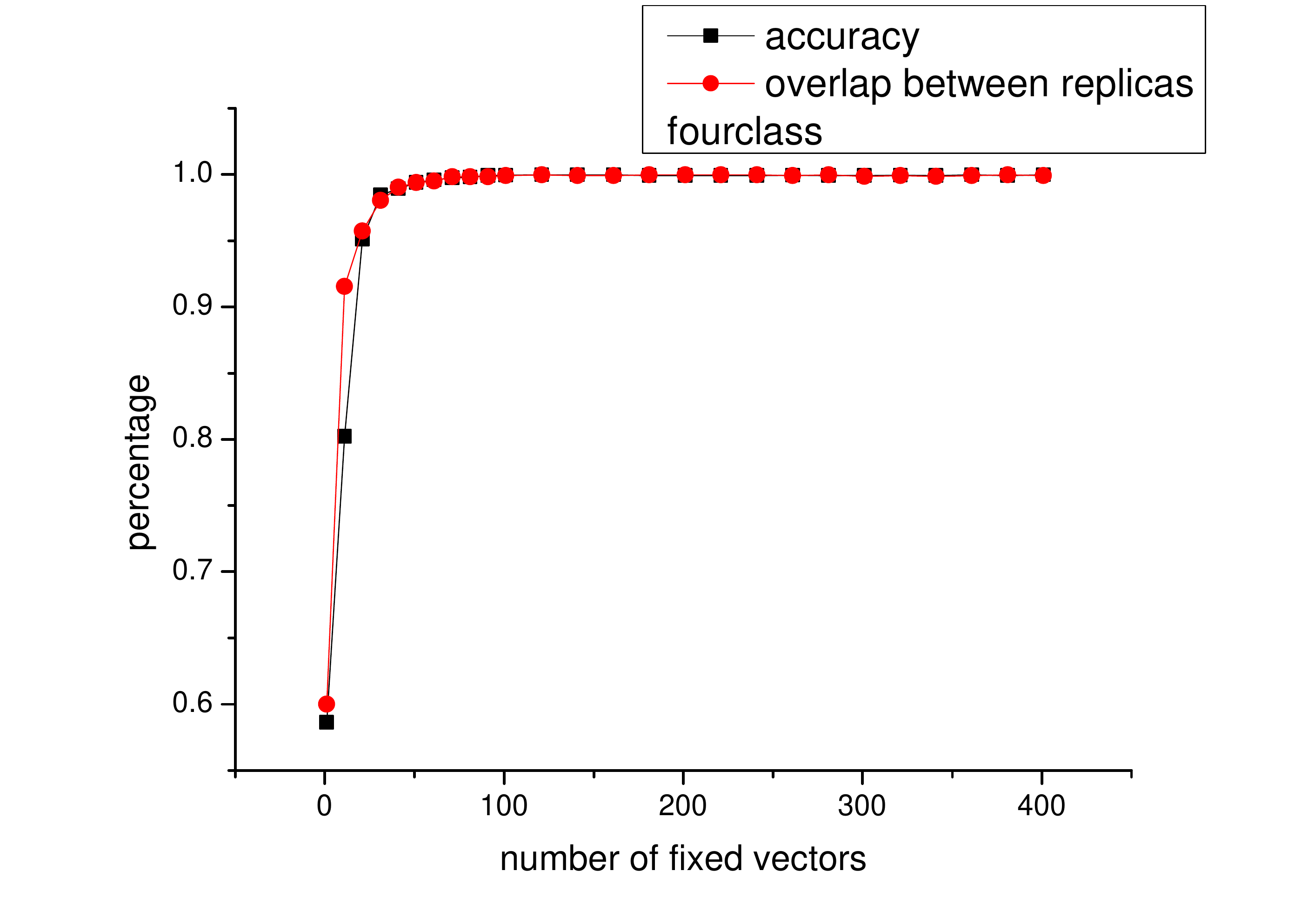}
\caption{A comparison between (1) the average replica overlap (as computed via the averaged variant of Eq. (\ref{O2})) and (2) average accuracy of a model with 
$\mathcal{R} = 5$ replicas for the Four-class benchmark when using the SRVM algorithm with a Gaussian kernel. The horizontal axis corresponds to the number of anchor points $v$.}
\label{Bo_Sun3}
\end{figure}
\begin{figure}[h]
\centering
\includegraphics[width=0.95\columnwidth]{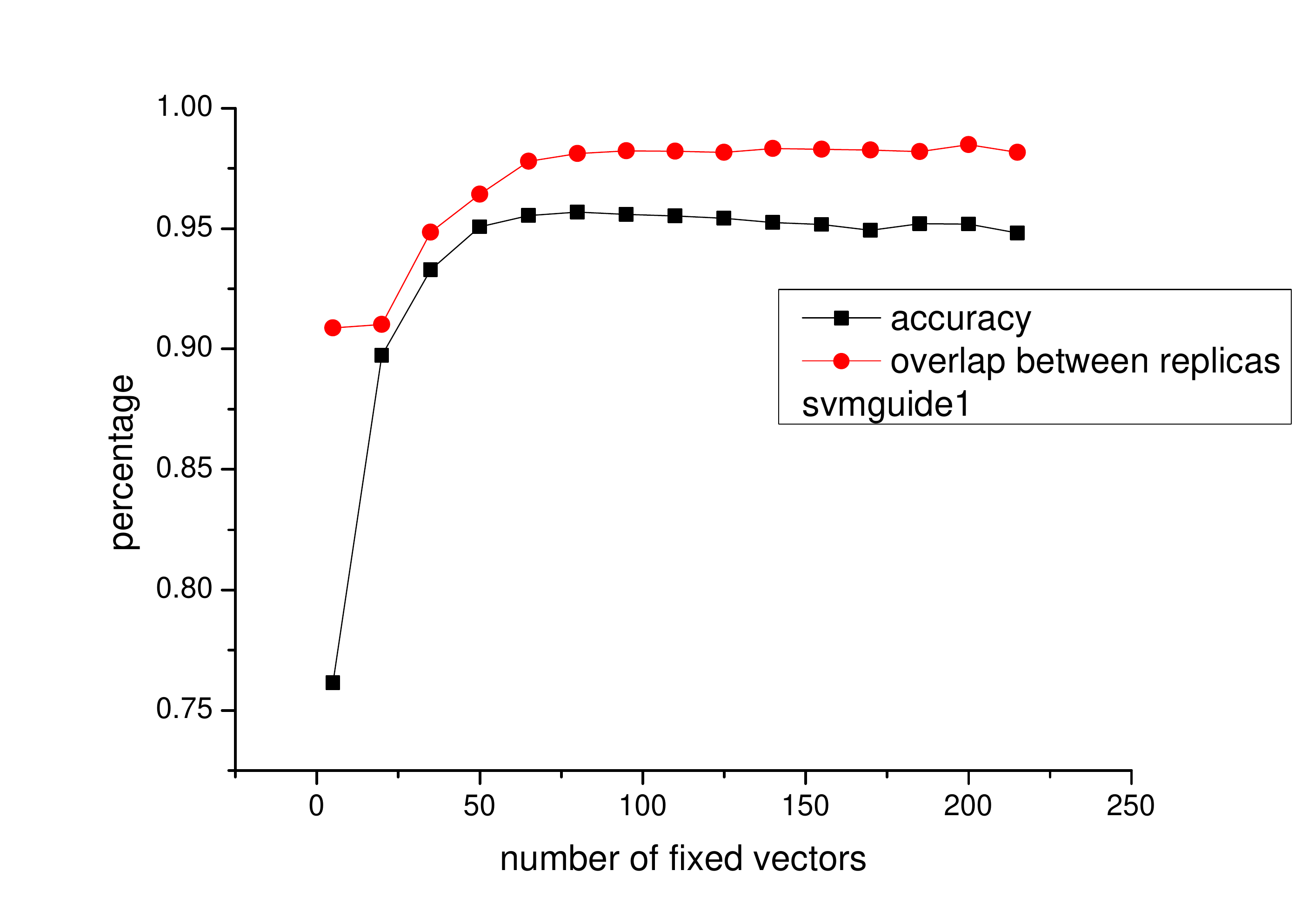}
\caption{Plotted on the same axes are (1) the average overlap between different pairs of replicas (calculated with the replica averaged 
variant of Eq. (\ref{O2})) and (2) the average accuracy as a function
of the number of anchor points $v$. This analysis was performed 
for the ``svmguide1'' benchmark classification problem with
the Gaussian based SRVM algorithm with $\mathcal{R} = 5$ replicas.}
\label{Bo_Sun1}
\end{figure}
\begin{figure}[h]
\centering
\includegraphics[width=0.95\columnwidth]{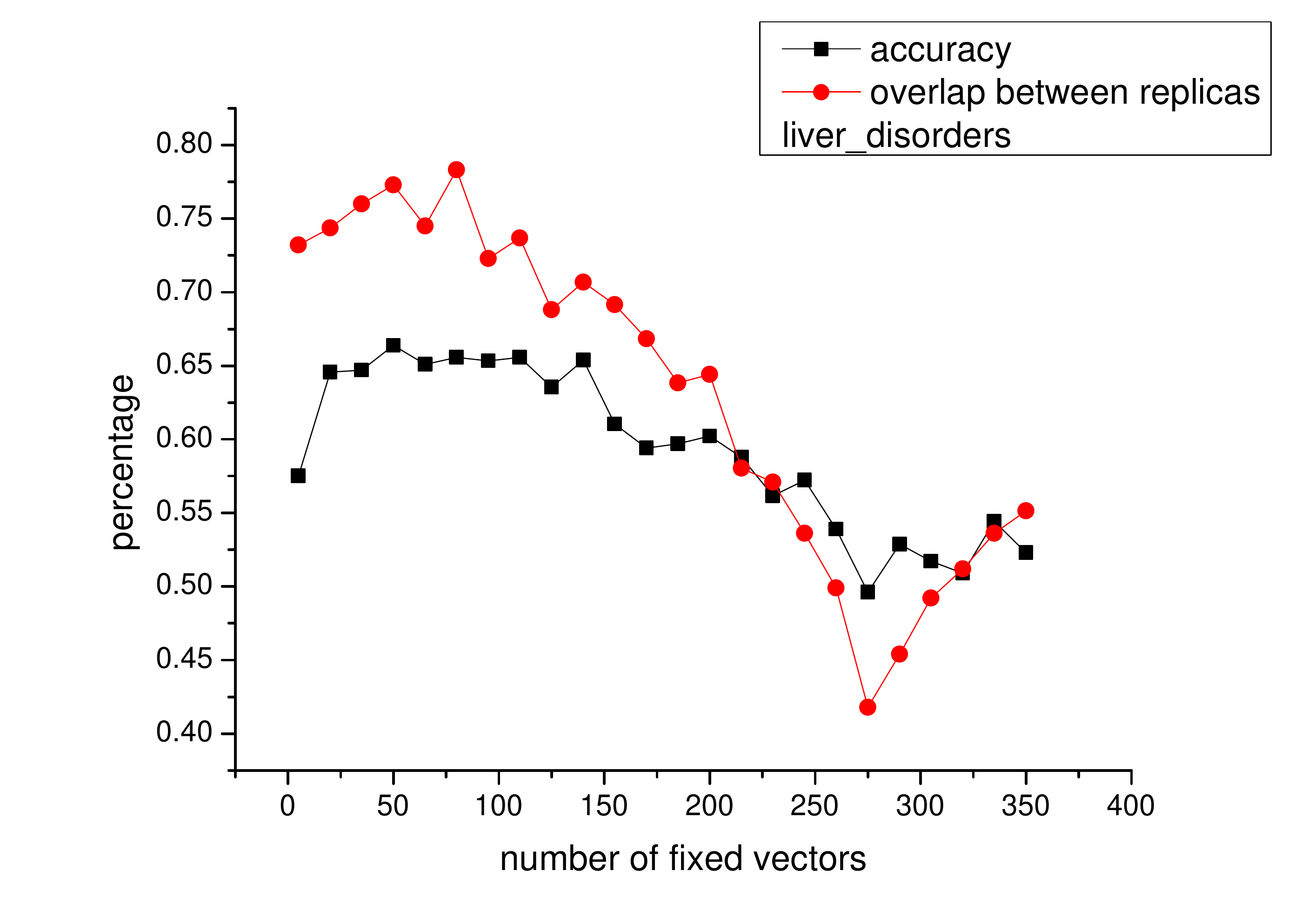}
\caption{The results of the SRVM algorithm for a Gaussian kernel with $\mathcal{R} = 5$ replicas for the ``liver disorder'' data set.
Similar to Figs. (\ref{Bo_Sun3},\ref{Bo_Sun1}), we plot the inter-replica overlap and accuracy as a function of the number of anchor points $v$
on the same set of axes.} 
\label{Bo_Sun2}
\end{figure}

\begin{eqnarray}
{\mathcal{O}_1=\sum_{\alpha>\beta} \vec{y}^{~\alpha}\cdot\vec{y}^{~\beta}}
\label{O1}
\end{eqnarray}
and measures the total overlap of the predicted labels of all replicas.
For each of the replicas $ 1 \le \alpha \le {\mathcal{R}}$, the vectors $\vec{y}^{~\alpha}$ have $g$ components (where $g$ is the number of distinct predicted (or, in some 
rare cases, fitted) data points $\vec{x}$ in each replica $\alpha$). 
This metric may be trivially averaged by dividing by the total number of 
distinct replica pairs, i.e., by multiplying the righthand side of Eq. (\ref{O1}) by $\frac{2}{g\mathcal{R}(\mathcal{R}-1)}$. 
A similar, but computationally more efficient, overlap function is defined as
\begin{eqnarray}
\mathcal{O}_2=\sum_{\alpha} \vec{y}^{~\alpha}\cdot\vec{\mathcal{V}}
\label{O2}
\end{eqnarray}
which measures the overlap of each replica with the overall vote as determined by Eq. (\ref{average}). The calculation of Eq. (\ref{O2}) requires storing less information than computing the overlap of Eq. (\ref{O1}). Similar to each vector in the set
$\{\vec{y}^{~\alpha}\}_{\alpha =1}^{\mathcal{R}}$, the vector $\vec{{\mathcal{V}}}$ has $g$ components (one component (the voted prediction) for each of the $g$ distinct examined data points $\vec{x}$).
(Similar to Eq. (\ref{O1}), an average is trivially calculated by dividing the righthand side of Eq. (\ref{O2})
by $(g \mathcal{R}$).) In Fig. \ref{1z}, we plot the results of the overlap measures of the LSVT data set. In these panels, the number of anchor points, $v$, is plotted on the x-axis with data sets for increasing values of replica numbers, $\mathcal{R}$, with increasing $\mathcal{R}$ being proportional to increasing y-axis values, shown. The data in Fig. \ref{1z} seems to display the same overall characteristics as that of the accuracy shown earlier, but it is important we compare them directly. In Fig. \ref{AO.}, we plot the average accuracy, and two overlap functions for $\mathcal{R}$=29 replicas and various numbers of anchor points, all scaled by their maximum values so as to be able to fall on the same plot. It is abundantly clear from the data, that the two definitions of the replica overlap and the accuracy scale in a one-to-one fashion, making either overlap function an excellent candidate to find the optimal parameter values without necessitating the calculation of model accuracy.
Similar trends are seen in Figs. (\ref{fig:Heartacc-anchor-repnum}(b), \ref{1z},\ref{Bo_Sun3}, \ref{Bo_Sun1}, \ref{Bo_Sun2}). 

In numerous situations, having a measure of the likelihood that a specific data point will be correctly labeled is important for identifying the location of classification boundaries. The existence of replicas in the SRVM algorithm precisely affords exactly such a measure. To this end, we may define a single data point overlap across replicas which can approximate the probabilities of classification. This allows us to determine whether a given point $\vec{x}_i$ is near a classification boundary, as when result in different labels, the point is likely near a boundary. We call this single instance overlap the `Agreement', and define it as

\begin{eqnarray}
\mathcal{A}_i \equiv \frac{1}{\mathcal{R}}|\sum_{\alpha = 1}^{\mathcal{R}} y_i^\alpha|.
\label{agree+}
\end{eqnarray}
This single instance metric complements the global overlaps of Eqs. (\ref{O1}, \ref{O2}). 
In general, the overlap $\mathcal{A}_{i}$ increases in tandem with the probability $\mathcal{P}_{i}$ of correctly classifying a given point $\vec{x}_i$. 
\begin{figure*}
	\centering
	\begin{subfigure}[t]{.45\textwidth}
		\centering
		\includegraphics[width=\linewidth]{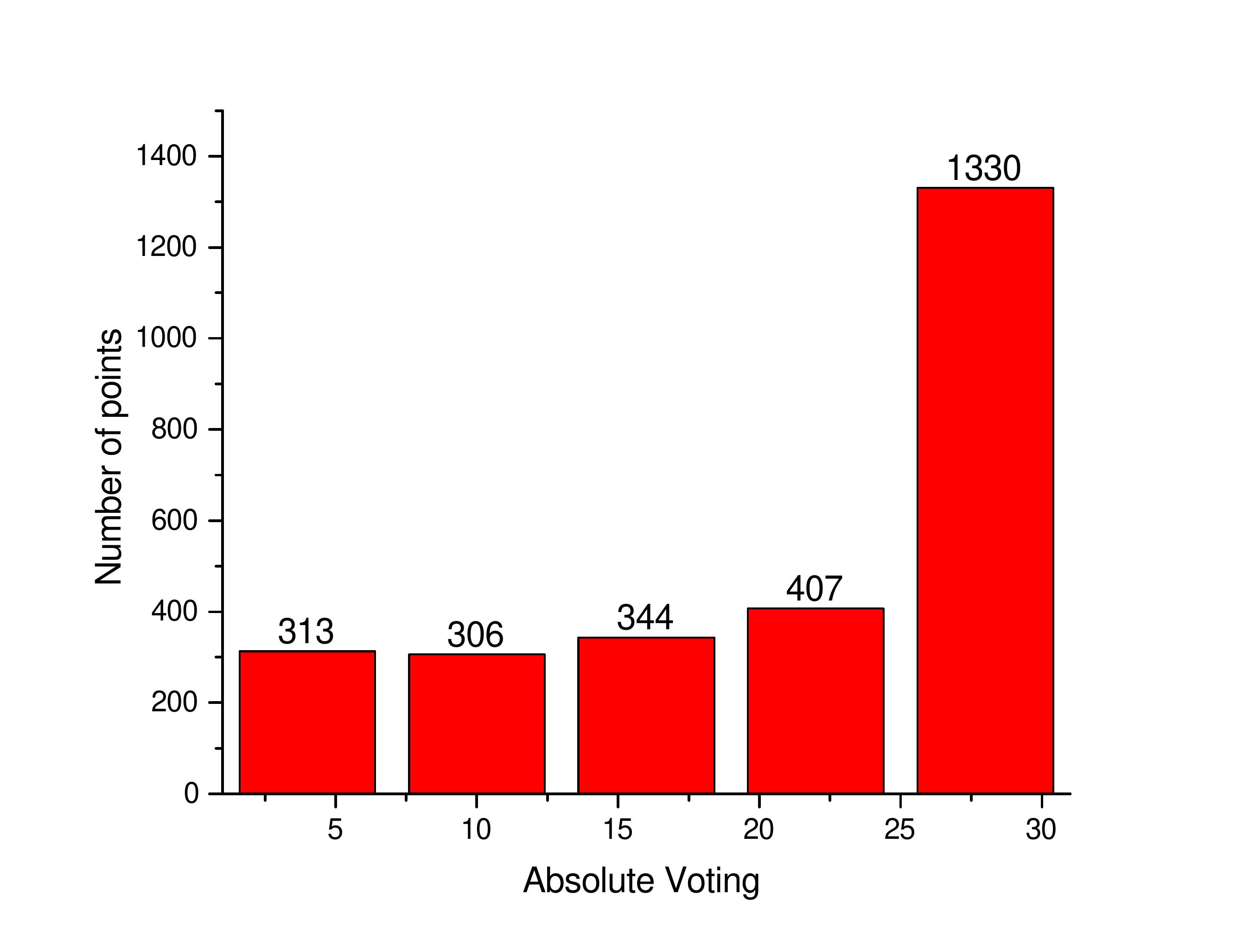}
	\end{subfigure}
	\begin{subfigure}[t]{.45\textwidth}
		\centering
		\includegraphics[width=\linewidth]{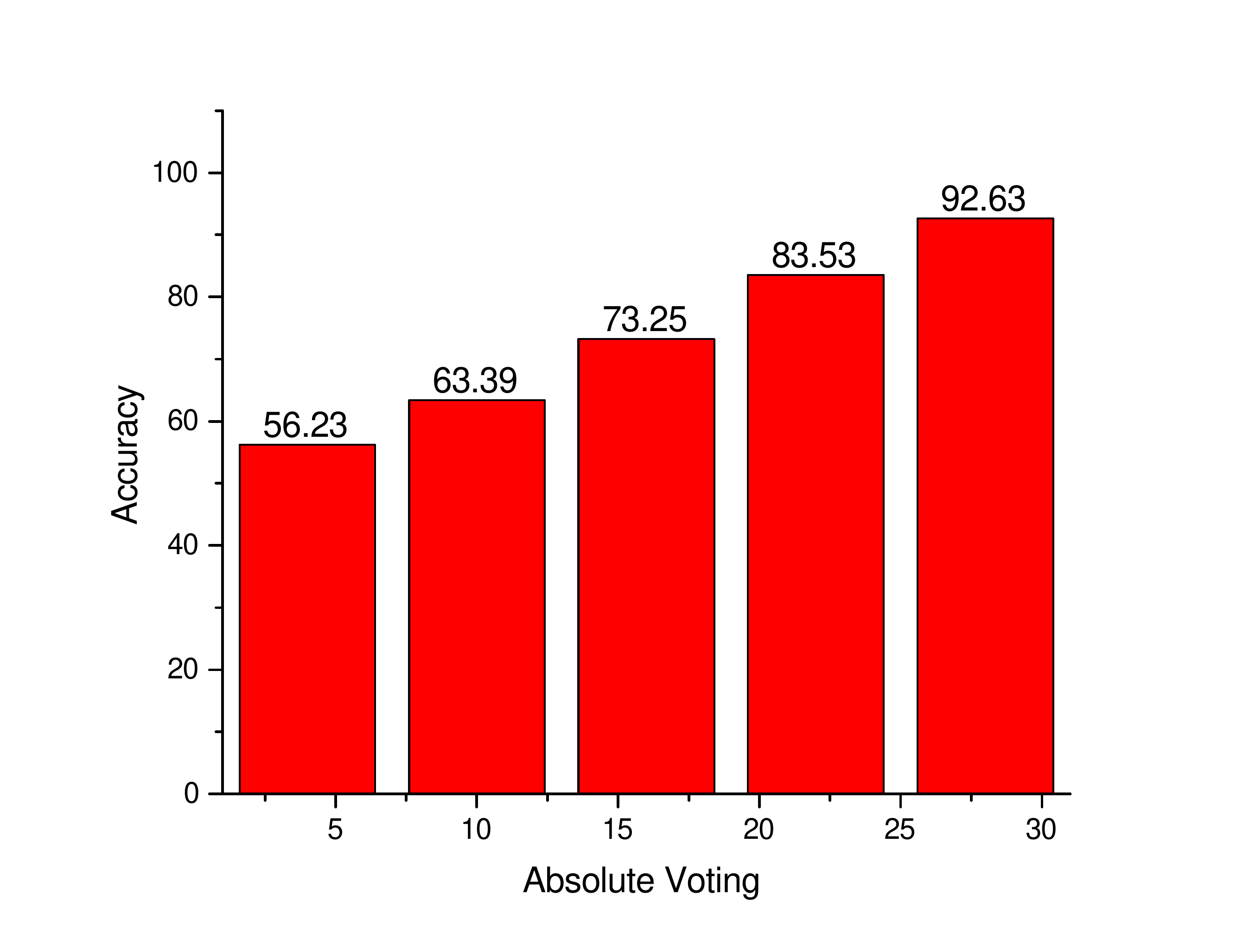}
	\end{subfigure}

		\caption{\label{fig:HeartAgreement} (Color Online.) An analysis of the Heart benchmark via ${\mathcal{R}} = 31$ replicas. The Heart benchmark has $270$ data points. The CV calculations were replicated 10 times so, overall, $270 \times 10 = 2700$ points were classified.		  (a) Bar chart showing the distribution of agreement values of the Heart benchmark data points across 10 five-fold cross-validations.
		  The rightmost bar denotes the number of points (1330 out of 2700) that were 
		   in nearly the same way by all 31 replicas. The leftmost bar corresponds to
		   the 313/2700 data points that were classified with a minimal Agreement (Eq. (\ref{agree+}))
		   amongst the 31 replicas. (b) A histogram of the average accuracy of the Heart benchmark data points in each bin of the Agreement values. Higher Agreement positively correlates with a higher average accuracy.}
\end{figure*}	

\begin{figure*}
	\centering
	\begin{subfigure}[t]{.45\textwidth}
		\centering
		\includegraphics[width=\linewidth]{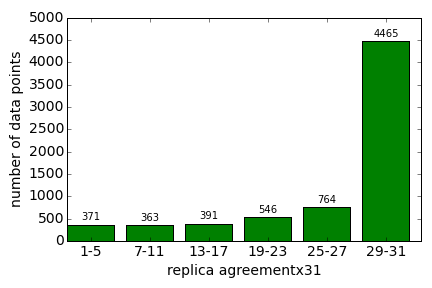}  
	\end{subfigure}
	\begin{subfigure}[t]{.45\textwidth}
		\centering
		\includegraphics[width=\linewidth]{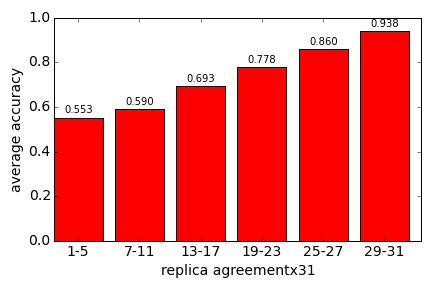}
	\end{subfigure}

		\caption{\label{fig:AustralianAgreement} (Color Online.) ``Australian'' data set. (a) Bar chart showing the distribution of agreement values of the data points across 10 five-fold cross-validations. (b) Bar chart showing average accuracy of the data points in each bin of agreement values. Higher agreement positively correlates with higher average accuracy.  On the horizontal axis, we plot the un-normalized sum of Eq. (\ref{agree+}), i.e., $|\sum_{\alpha = 1}^{\mathcal{R}} y_i^\alpha|$.}
\end{figure*}

\begin{figure}
\centering
\includegraphics[width=0.95\columnwidth]{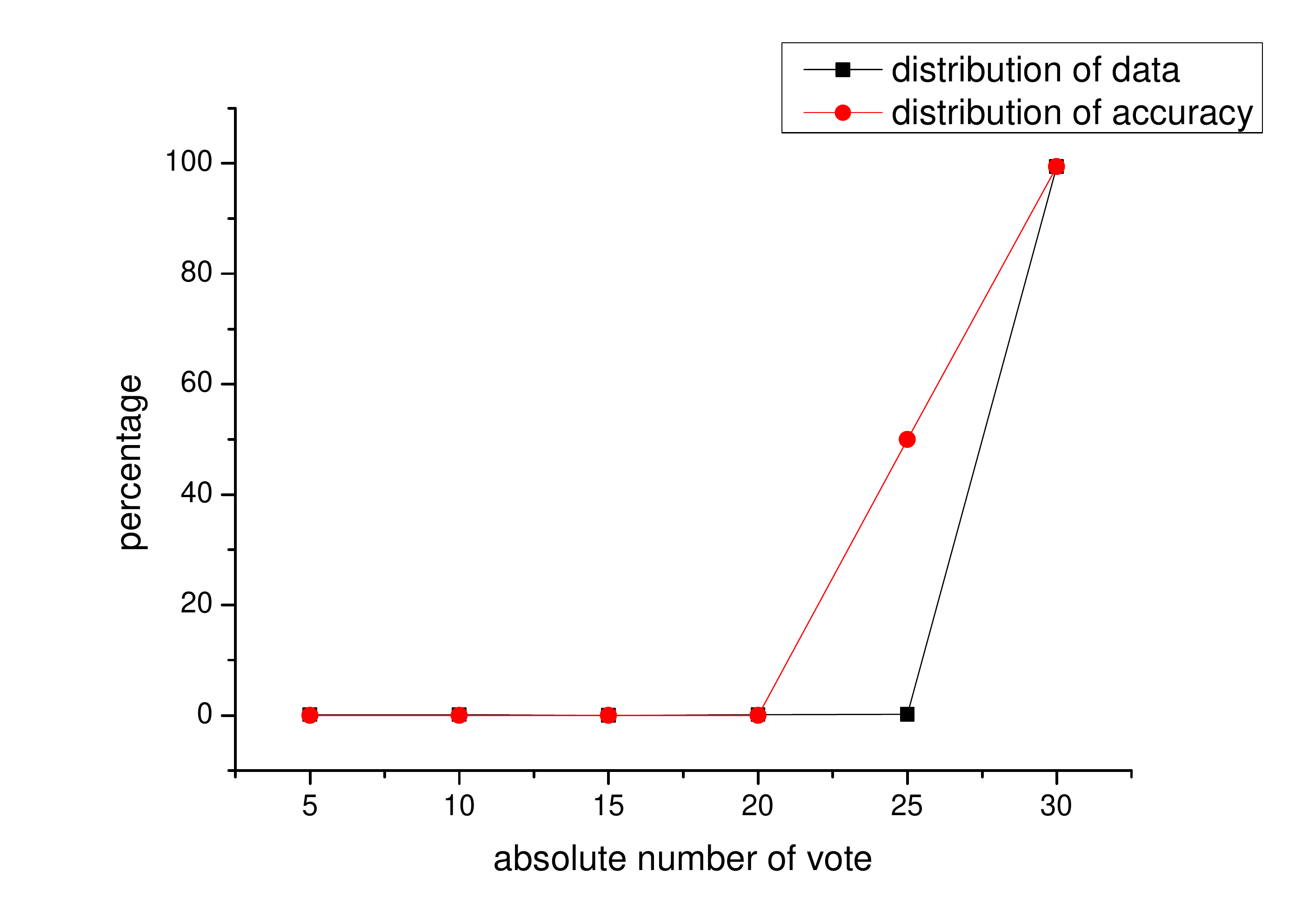}
\caption{(Color online.)
Distribution of data points and accuracy for the Four-class benchmark. On the horizontal axis, we plot the``absolute number of votes''- the un-normalized sum of Eq. (\ref{agree+}), i.e., $|\sum_{\alpha = 1}^{\mathcal{R}} y_i^\alpha|$. The ``distribution of data points" marks which fraction of the data points
have a given ``absolute number of votes'' (thus the sum of this distribution over all possible ``absolute number of votes'' is unity). The accuracy curve is, generally,
monotonic in the replica overlap as is manifest here by the ``distribution of data points'' fraction.}
\label{fig:fig221}
\end{figure}

\begin{figure}
\centering
\includegraphics[width=0.95\columnwidth]{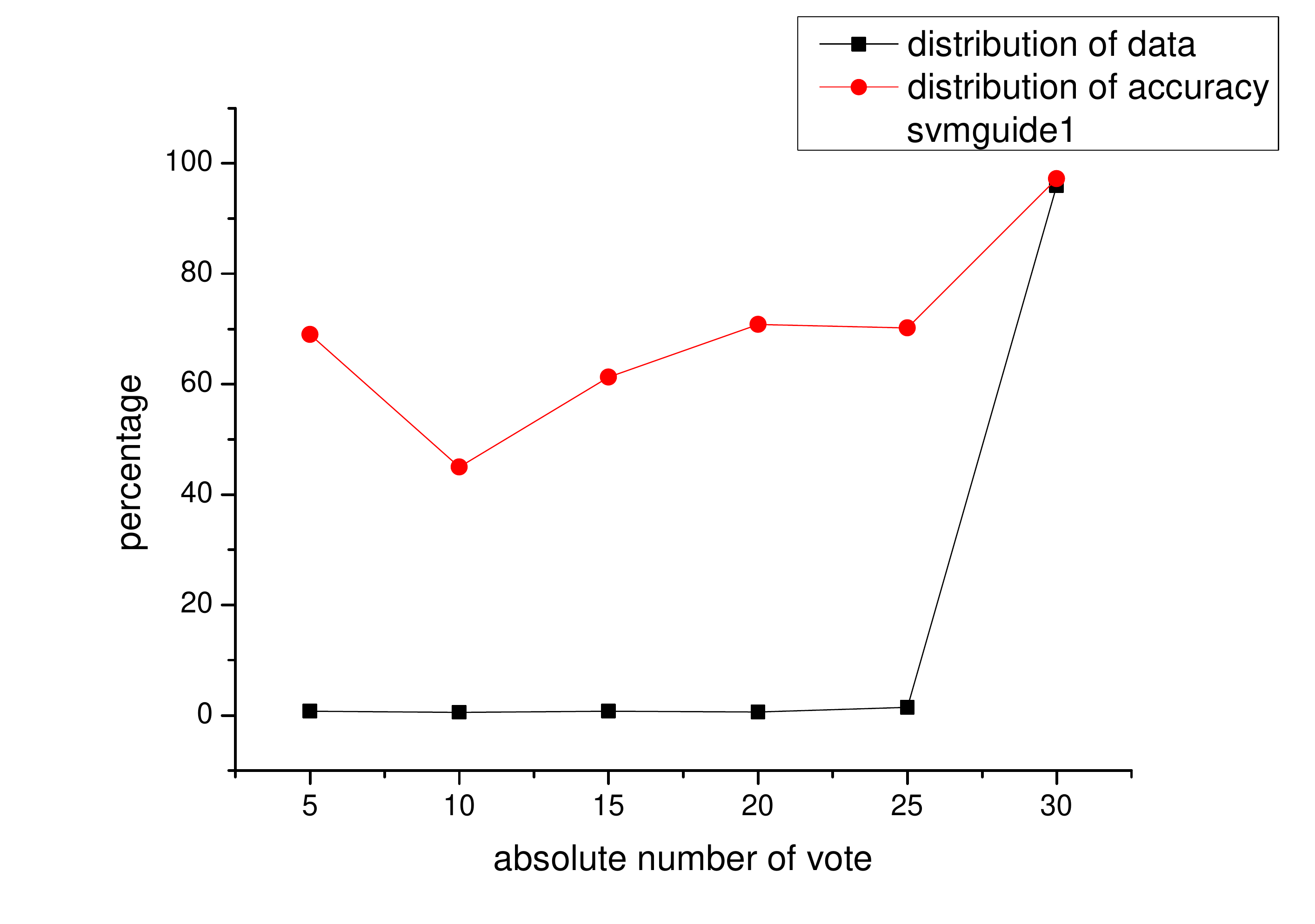}
\caption{(Color online.)
The distribution of data points and accuracy for the svmguide1 data set. See the caption of Fig. \ref{fig:fig221} for the definition of the axes and curves.}
\label{fig:fig221+}
\end{figure}

\begin{figure}
\centering
\includegraphics[width=0.95\columnwidth]{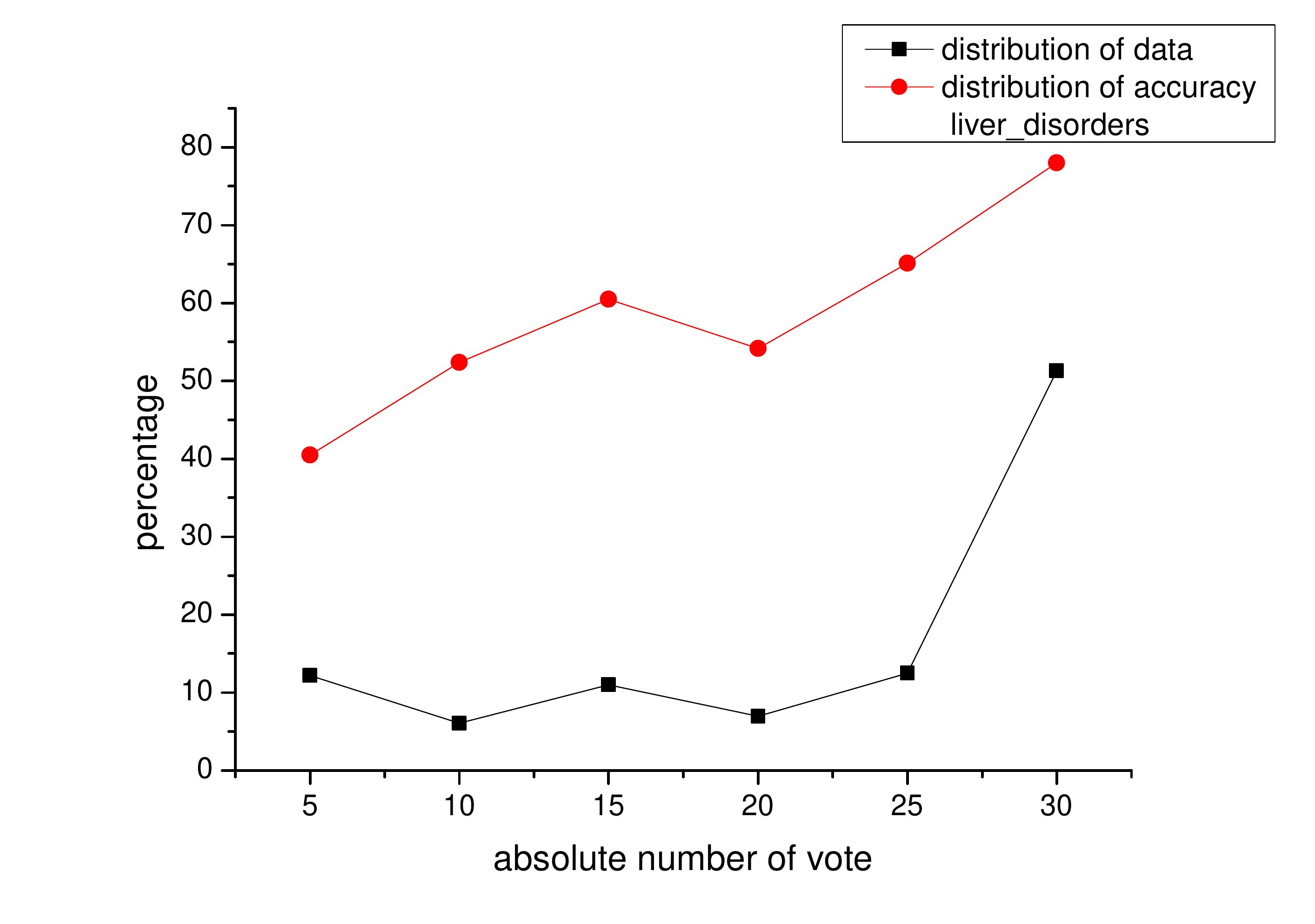}
\caption{(Color online.)
The correlation between the replica correlations (``distribution of data points'') and accuracy for the liver disorder benchmark. The definition of graph is similar to that in the caption of Fig. \ref{fig:fig221}.}
\label{fig:fig222}
\end{figure}

 The bar chart in figure \ref{fig:HeartAgreement} (a) shows how well the predictions of different replicas agree.
The majority of points fall in the last bin (meaning that for most points, different replicas predicted the same result). 
We also calculated the corresponding accuracy for each bin in Fig. \ref{fig:HeartAgreement}(b). 
Again, it is apparent from the bar chart that the points with the maximum replica voting agreement have the highest accuracy as it is expected.
As we mentioned earlier, the Heart data consists of  270 data points each of 13 features. Replicas are of 50 points in thirteen dimensions. The average accuracy after 10 runs of voting was 81.25\%; the corresponding accuracy of SVM is 82.4\%. Thus, in this example (as in others), the accuracies obtained by both SRVM and SVM are
nearly the same.

\begin{figure*}
	\centering
	\begin{subfigure}[t]{.45\textwidth}
		\centering
		\includegraphics[width=\linewidth]{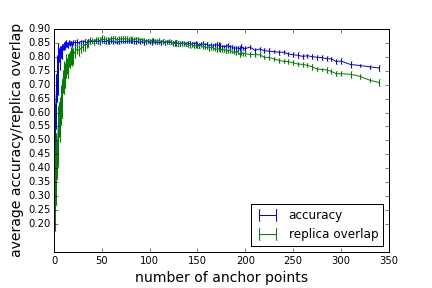}
	\end{subfigure}
	\begin{subfigure}[t]{.45\textwidth}
		\centering
		\includegraphics[width=\linewidth]{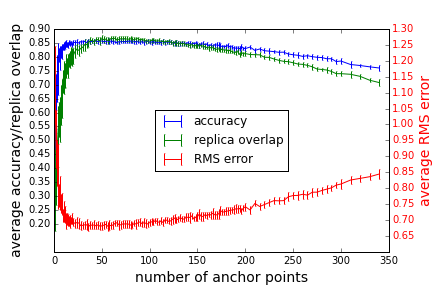}
	\end{subfigure}

		\caption{\label{fig:AvgAgreementRMSError}(Color Online.) Australian data set. Each plotted data point represents the average over 20 runs. (a) Average agreement and average accuracy with varying number of anchor points (similar to Fig. \ref{fig:AUSTacc-anchor-repnum}). Right: (b) Average agreement, average accuracy and average RMS error with varying number of anchor points.}

\end{figure*}

The correlation between availability and classification accuracy was also investigated using the Australian data set. The Australian data set was used to demonstrate the correlation between replica overlap and datapoint prediction accuracy. For this test, we used Gaussian kernel models containing 31 replica with 50 anchor points each. A total of 10 models were generated, and a 5-fold cross-validation on the data set was ran on each model. For each 5-fold cross-validation, every data point in the data set will be a test data point exactly once. Any test data point $x_i$ was classified by its replica agreement $\mathcal{A}_i$. 
In Fig. \ref{fig:AustralianAgreement}, we binned the test data points based on their agreement values (multiplied by replica number ($\mathcal{R} = 31$) for clarity of presentation) and calculated an aggregate accuracy for each bin. We see that indeed the data points with higher replica agreement in general are also being predicted with higher accuracy, showing a clear positive correlation between replica agreement and prediction accuracy. Another feature from Figs. (\ref{fig:AUSTacc-anchor-repnum}, \ref{fig:AvgAgreementRMSError})  is that the vast majority of data points have good replica agreement.  In Figs. (\ref{fig:fig221}, \ref{fig:fig221+}, \ref{fig:fig222}), we report on similar tendencies found for the Four-class, svmguide1, and liver disorder benchmarks. 

The Australian data set was also used to show that the average agreement across the data set is also a useful replica overlap function. Using the same procedure for Fig. \ref{fig:AUSTacc-anchor-repnum}, we used Gaussian kernel models with varying number of anchor points and 5 replicas each, and calculated the average agreement of all datapoints. The results are plotted with the average accuracy in Fig. \ref{fig:AvgAgreementRMSError}(a), and we see strong correlation between the average accuracy and the average agreement. If we introduce the average RMS error as the learning energy $E$ scaled by the number of data points: 
$RMS = \mathcal{F}/N$, we see in Fig. \ref{fig:AvgAgreementRMSError}(b) the average RMS error correlates negatively with 
both the average accuracy and the average agreement.

\begin{figure*}
	\centering
	\begin{subfigure}[t]{.45\textwidth}
		\centering
		\includegraphics[width=\linewidth]{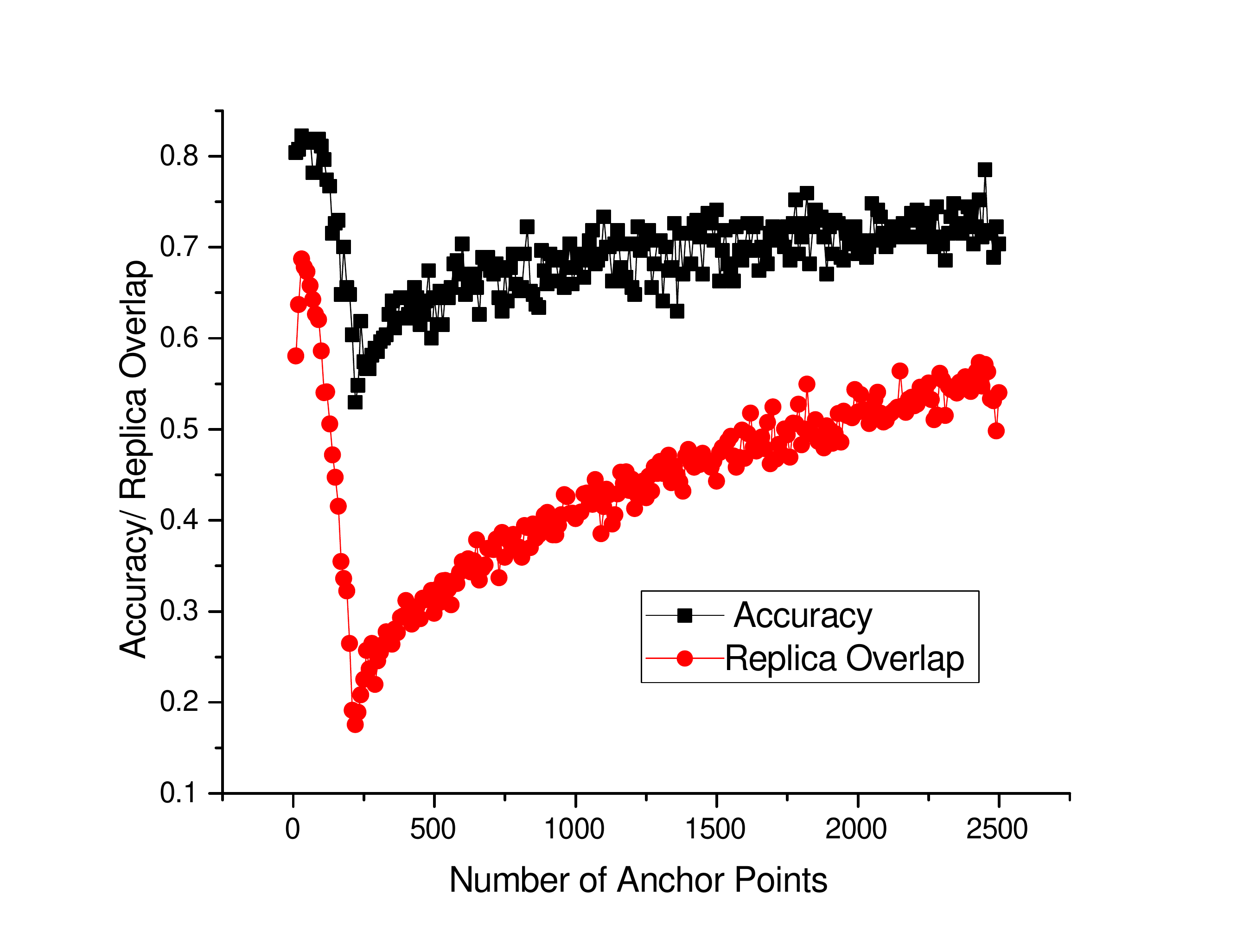}
	\end{subfigure}
	\begin{subfigure}[t]{.45\textwidth}
		\centering
		\includegraphics[width=\linewidth]{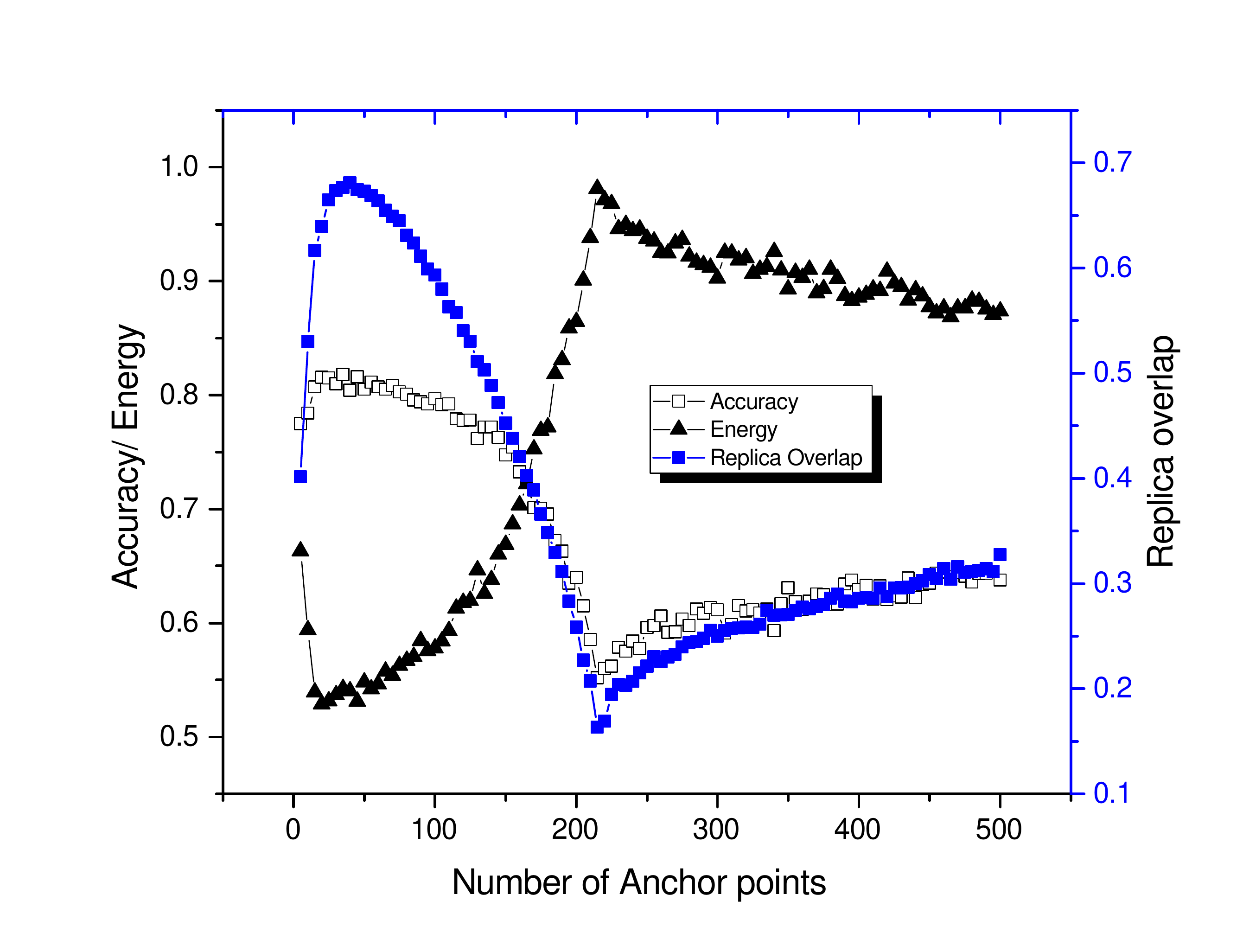}
	\end{subfigure}

		\caption{\label{fig:Heart-RMS} (Color Online.) Heart data set. Each data point is average of 20 runs. (a) Average agreement and average accuracy with varying number of anchor points. (b) Average agreement, average accuracy, and average RMS error with varying number of anchor points. \cite{artifact}}

\end{figure*}

In Fig. \ref{fig:Heart-RMS}, we show the correlation between averaged replica overlap and averaged accuracy for different number of anchor points. The range of anchor points runs from 10 to 500 while the number of replicas are being kept fixed at 31. Similar to the Australian and other data sets that we analyzed, both the replica overlap and the accuracy closely tracked each another (and further correlated with the value of the energy function of Eq. (\ref{Fmin})). Here, these quantities were non-monotonic as a function of the number of anchor points. The ``energy'' curve is this figure corresponds to the average of Eq. (\ref{Fmin}) over 31 different replica realizations. Five-fold CV was employed in our tests for the accuracy of the predictions. Perusing this Figure, we see that the averaged penalty function 
of Eq. (\ref{Fmin}) becomes minimal when the highest inter-replica overlap is achieved and when the predicted classifications are of the highest accuracy.

\begin{figure}[h]
\centering
\includegraphics[width=0.95\columnwidth]{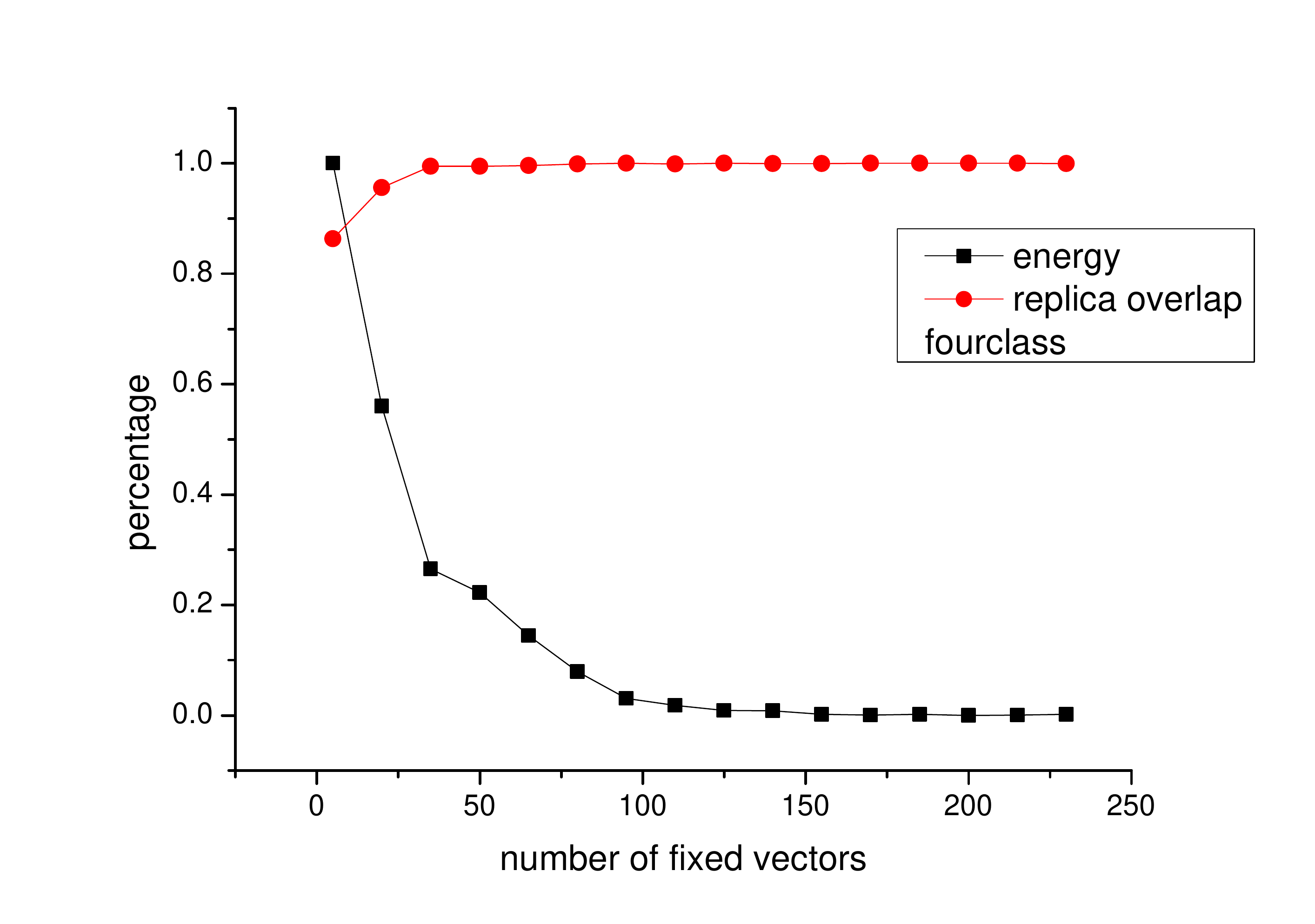}
\caption{A comparison of the average replica overlap and average energy (Eq. (\ref{Fmin}))) for the Four-class data set when $\mathcal{R} = 5$ replicas are used.}
\label{Bo_Sun4}
\end{figure}

\begin{figure}[h]
\centering
\includegraphics[width=0.95\columnwidth]{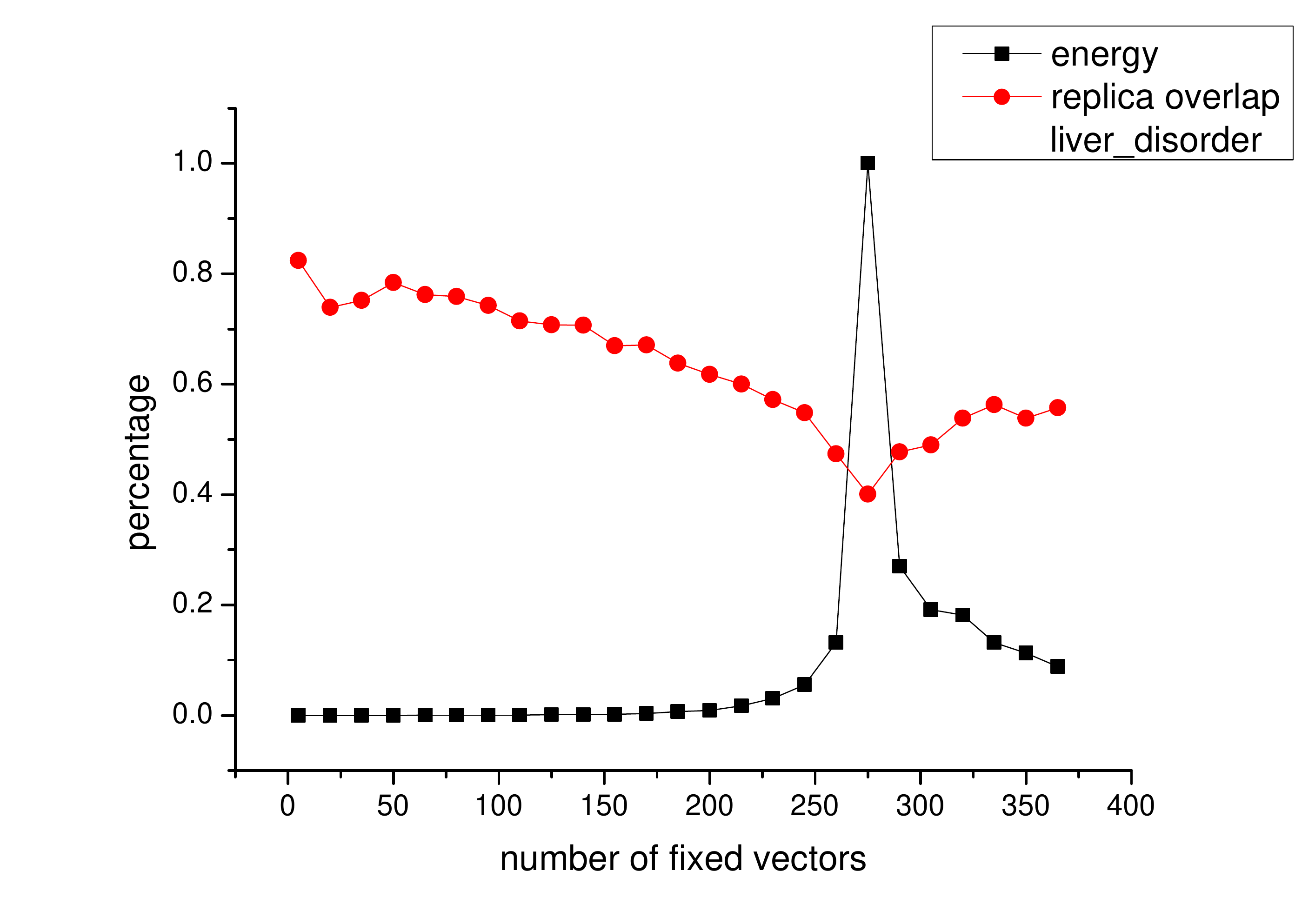}
\caption{A comparison between the average replica overlap and the average energy of a model with $\mathcal{R} = 5$ replicas for the liver disorders data set.}
\label{Bo_Sun5}
\end{figure}

\begin{figure}[h]
\centering
\includegraphics[width=0.95\columnwidth]{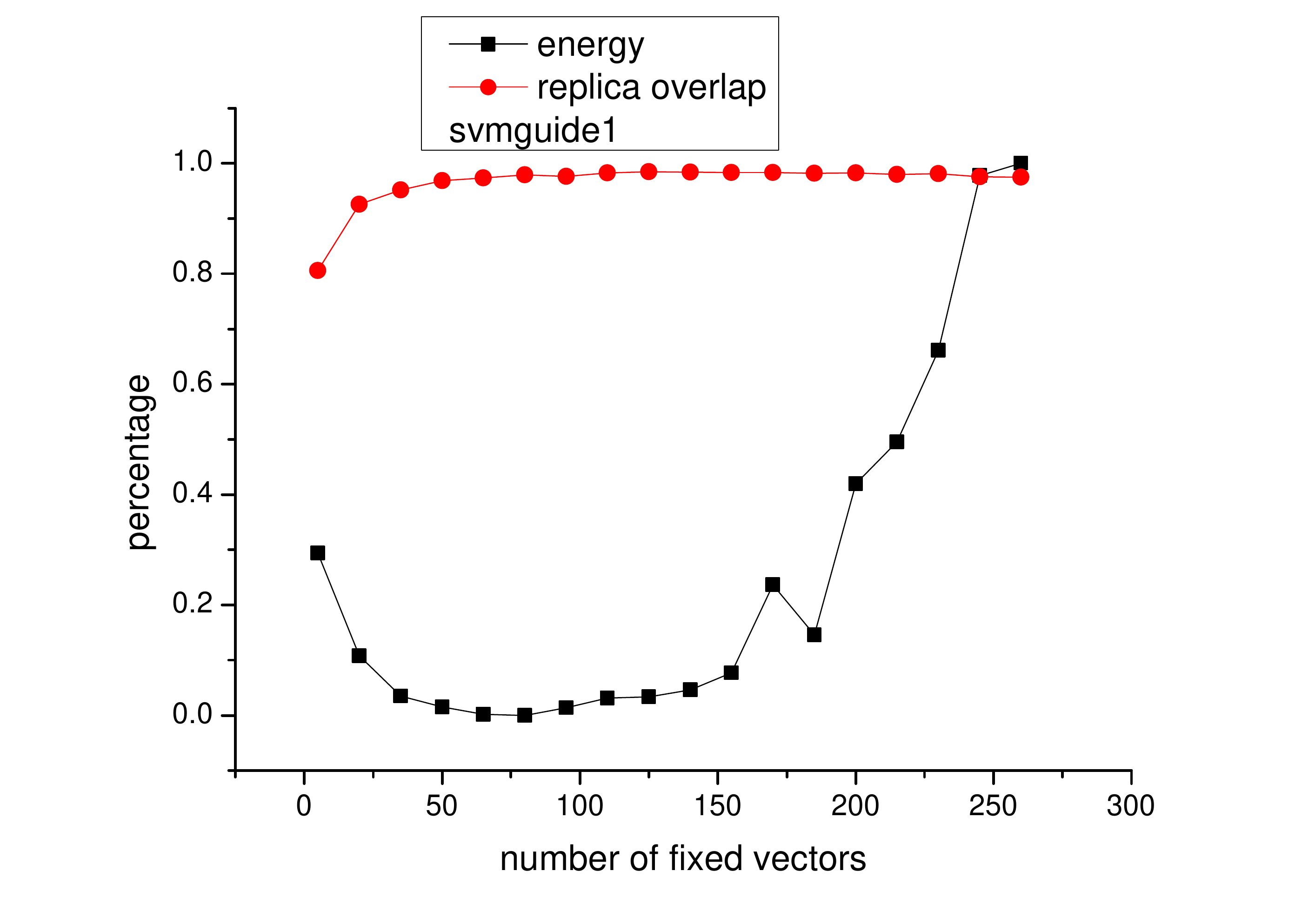}
\caption{The average replica overlap and the average energy when analyzing the svmguide1 benchmark data set with 5 replicas.}
\label{Bo_Sun6}
\end{figure}

In Figs. (\ref{Bo_Sun4}, \ref{Bo_Sun5}, \ref{Bo_Sun6}), we similarly demonstrate the correlation between the found accuracy and 
the lowest value of the penalty function (or energy) of Eq. (\ref{Fmin}) for different data sets. 

In summary, the single control parameter that governs our algorithm (the number of anchor points $v$) 
may be {\it automatically optimized} by an examination of the inter-replica overlaps as well as analysis of the energy costs of
Eq. (\ref{Fmin}). In Section \ref{layer:sec}, we will discuss more efficient determination of the optimal parameters by 
recursively applying machine learning onto itself. We first, however, explicitly turn to a notable attribute of the SRVM method
that has its origins in its stochastic nature.

\subsection{Class Imbalance and Alternative Performance Metrics}
\label{sec:imbalance}

\begin{figure}
	\centering
	\includegraphics[width=\linewidth]{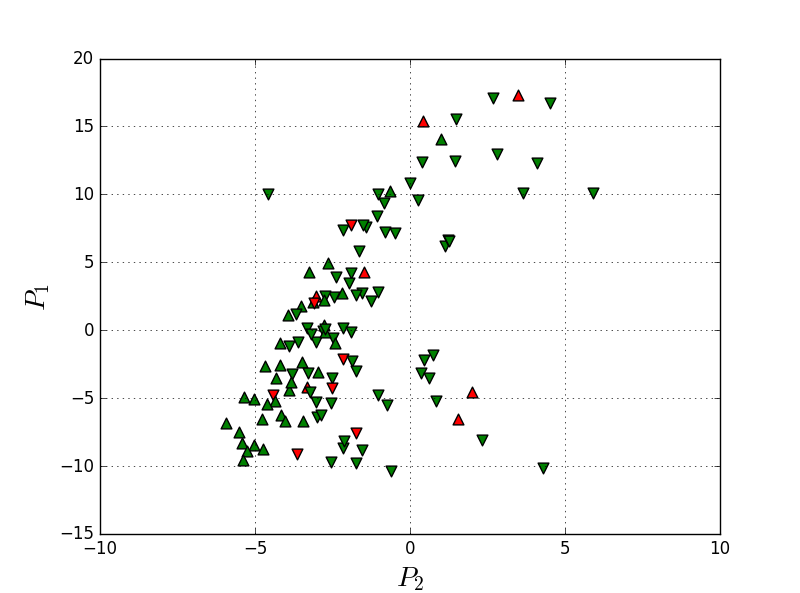}
	\caption{(Color Online). The LSVT data set projected into the plane of the first two principle components, so as to visualize model performance. Data points denoted with an upward pointing triangle are points with a known label `+1' and downward pointing triangle are those points having their known label be `-1'. Points which are colored green correspond to points correctly classified by the SRVM algorithm, whereas points colored red, were incorrectly classified.}
	\label{LSVTAll.}
\end{figure}

\begin{figure*}
	\centering
	\includegraphics[width=\linewidth]{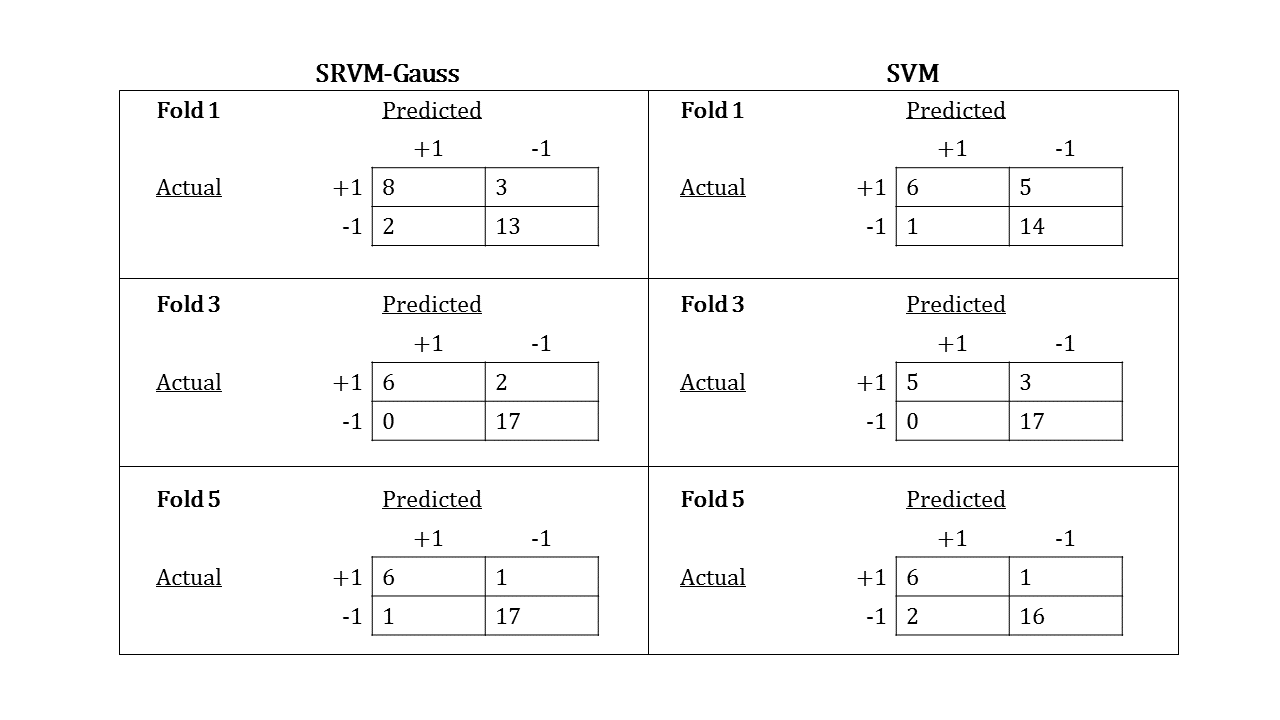}
	\caption{LSVT data set. Confusion Tables constructed from three folds of a five-fold cross validation for the SRVM and SVM algorithms. Overall, the confusion tables make clear that the SRVM algorithm is slightly less inclined toward false negatives (FN) than SVM, which is and important result, as the class imbalance is toward the negative side in the data set.}
	\label{CAll.}
\end{figure*}

In Fig. (\ref{LSVTAll.}), we depict the results of a principal component analysis of the LSVT data set (in the figure, we outline the two dominant principal components). This analysis enables us to visualize where our algorithm fails to find the correct answer for the LSVT data set.  As seen in this figure, while there is no apparent distribution in principal component space of the cases that we obtained incorrectly, the two classes are massively imbalanced (as is often the case in classification sets). Other metrics are necessary to compare our results to those of SVM (and other algorithms).
To that end, we briefly regress to the ``accuracy paradox'' \cite{AP}.  This colloquial ``paradox'' is simple to explain: if the data set given is heavily imbalanced so that most of the provided data belong to one type, one might as well just guess the dominant answer every time and miss subtle instances. This must be taken into consideration. 
To that end, in Table \ref{tab:problems9+} and Fig. (\ref{CAll.}), we provide a confusion table and look at how well we perform while qualifying true positives and negatives and false positive and negatives. The specificity and sensitivity are related to true positives and false negative rates. This information may be used to compute various metrics; these measures combine class imbalance, specificity, sensitivity, and performance. 
A particularly useful measure for assessing class imbalance (appearing in Table \ref{tab:problems+} )is the so-called ``Cohen's $\kappa$''. In terms of these metrics, the superiority of SRVM over SVM is made apparent (by a very large statistically significant difference). Strikingly, the statistical measures in Table \ref{tab:problems+} indicate that, colloquially speaking, SVM edoes more ``guessing'' as compared to the SRVM algorithm and tends to become ``lucky'' by predicting the dominant class more often. This is to be expected since SVM segments feature space with a unique kernel and only considers points on a specific boundary that need to be carefully classified. By contrast, the distinguishing attribute of SRVM is that numerous stochastic replicas are considered- a characteristic that tends to lead to less bias. Taken together, the Cohen's $\kappa$ values \cite{Cohen}, along with the $F_{1}$ \cite{F-score} (that does not take into account true negatives) and $C_{M}$ (Matthews Correlation coefficients) \cite{Matthews} metrics illustrate that SRVM exhibits a statistically significant advantage over the SVM algorithm insofar as the lack of inbuilt class imbalance bias is concerned.

\begin{table}[t]
	\begin{tabular}{| c | c | c | c | c |}
		\hline
		Label & Fold 1 & Fold 3 & Fold 5   \\ 
		\hline
		SRVM Right, SVM Right & 18 & 20 & 22   \\
		SRVM Wrong, SVM Wrong & 4 & 2 & 2  \\
		SRVM Right, SVM Wrong & 2 & 1 & 0  \\
		SRVM Wrong, SVM Right & 1 & 0 & 0  \\
		SRVM Mixed, SVM Right & 1 & 2 & 0  \\
		SRVM Mixed, SVM Wrong & 0 & 0 & 1  \\
		\hline
	\end{tabular}
	\caption{Number of test data points corresponding to a given pair of outcomes for the Gaussian kernel SRVM and SVM algorithms across three different folds of a cross-validation set for the LSVT benchmark. }
	\label{tab:problems9+}
\end{table}

\begin{table*}
	\begin{tabular}{*{8}{|@{\hskip 0.2in}c@{\hskip 0.2in}}|}
		\hline
		\emph{Fold} & \emph{Classifier} & \emph{Sensitivity} & \emph{Specificity} & \emph{Precision} & \emph{$\kappa$} & \emph{F$_1$} & \emph{$C_M$}\\
		\colrule
		\multirow{2}{1em}{\textbf{1}} & SRVM-Gauss & 0.727 & 0.866	& 0.8 & 0.56 & 0.761 & 0.603
		\\
		& SVM & 0.545 & 0.933 & 0.857 & 0.436 & 0.667 & 0.533
		\\
		\hline
		\multirow{2}{1em}{\textbf{3}} & SRVM-Gauss & 0.75 &	1 &	1 &	0.549 & 0.857 & 0.819
		\\
		& SVM & 0.625 &	1 &	1 &	0.447 & 0.769 & 0.728
		\\
		\hline
		\multirow{2}{1em}{\textbf{5}} & SRVM-Gauss & 0.857 & 0.944 & 0.857 & 0.541 & 0.857 & 0.801
		\\
		& SVM & 0.857 & 0.888 &	0.75 & 0.519 & 0.8 & 0.718		
		\\
		\hline
	\end{tabular}
	\caption{Alternative metrics for assessing the performance of the SRVM and SVM algorithms on the LSVT data set. These metrics take into account class imbalance in the training set, and are therefore a more robust and powerful measure of algorithm performance. The table makes clear that despite the statistically insignificant difference in accuracy between the algorithms, the SRVM method is consistently better across all metrics when class imbalance is accounted for.}
	\label{tab:problems+}
\end{table*}
\subsection{Layered Voting: Multiple Kernels and Recursive Learning}
\label{layer:sec}

As we remarked earlier, there are many possible ``interactions'' between individual replicas (see the schematic of Fig. (\ref{fig:MRAlandscape})). The equal weight average of Eq. (\ref{average}) is merely one of the simplest choices to deciding how fuse the results of different replicas into a collective prediction. 

\begin{figure}
	\centering
	\includegraphics[width=\linewidth]{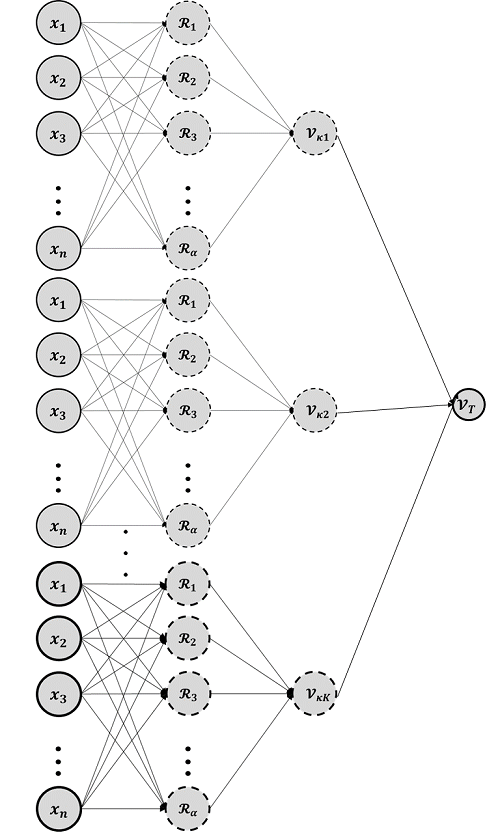}
	\caption{(Color Online.) Schematic representation of the layered kernel approach to the SRVM method. Allowing different multiple kernel functions to sequentially vote (rightmost layer) following a replica voting amongst functions of the same type (the intermediate layer in the above sketch) leads to an SRVM algorithm with an architecture that emulates a traditional neural net with a ``hidden'' intermediate layer. That is, the multiple replica results are voted on in the first layer (as described in 
	all earlier sections of this paper); the results of these votes using different kernels are now combined anew with a different voting functions. One may assign varying weights to the different kernels so as to optimize the accuracy; this will be taken up in a future paper.}
	\label{Net.}
\end{figure}

\begin{figure}[h]
\centering
\includegraphics[width=0.65\columnwidth]{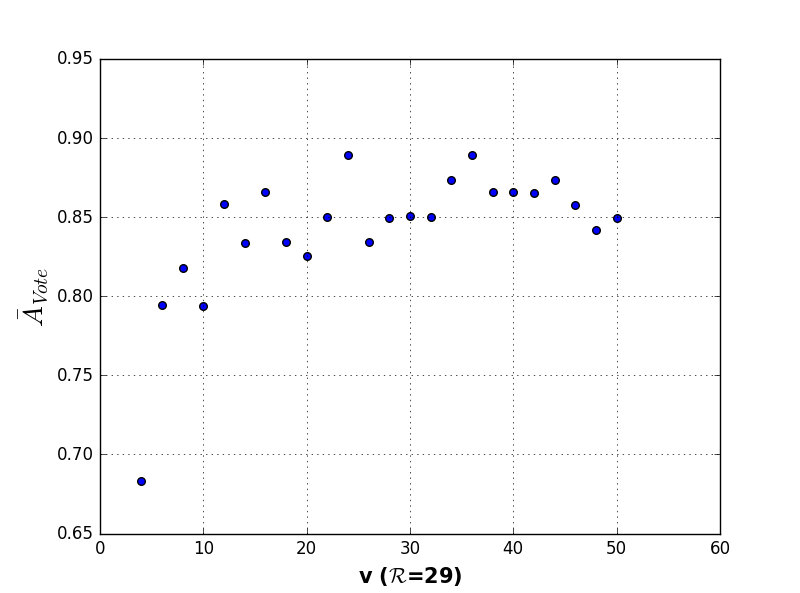}
\caption{LSVT data set analyzed with layered voting (see Section \ref{layer:sec}) with a uniform weight, see Eq. (\ref{average-general}). We plot the accuracy as a function of the number of anchor points 
$v$ for the Gaussian kernel. The number of replicas is held fixed at ${\mathcal{R}} = 29$. The accuracy is slightly higher than that of the single kernel (Fig. (\ref{Stab.})). The accuracy is expected to increase when the weights of the different voting kernels are optimized (and not arbitrarily set to a uniform equal value
as they are here)).}
\label{ZZoom:Reg_1}
\end{figure}

\begin{figure}[h]
\centering
\includegraphics[width=0.65\columnwidth]{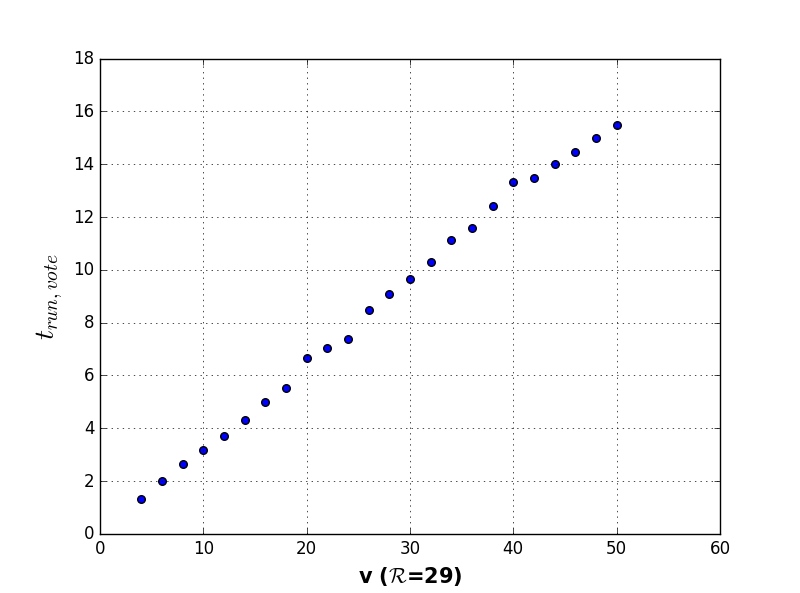}
\caption{LSVT data set analyzed with layered voting; see caption of Fig. (\ref{ZZoom:Reg_1}). Run time as a function of the number of anchor points for the Gaussian kernel.}
\label{ZZoom:Reg_2}
\end{figure}

\begin{figure}[h]
\centering
\includegraphics[width=0.65\columnwidth]{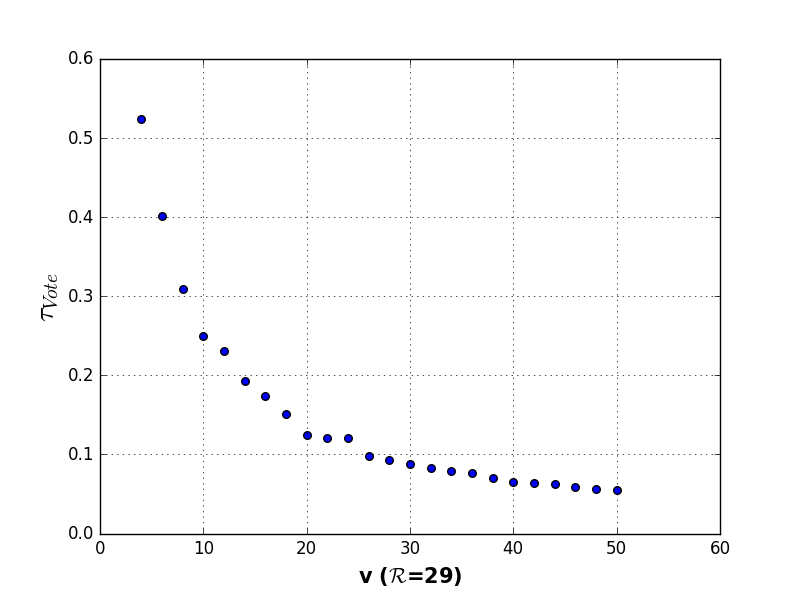}
\caption{LSVT data set analyzed with layered voting; see caption of Fig. (\ref{ZZoom:Reg_1}). Coefficient of performance (COP) as a function of the number of anchor points for the Gaussian kernel.}
\label{ZZoom:Reg_3}
\end{figure}

Following the conventional terminology of neural networks, we may add ``hidden layers'' to the SRVM by allowing different kernels to all vote. Each kernel predicts an outcome on its own for each instance. We can combine voting results from different kernels to come together by voting anew from the results from the first (single kernel) votes, see Fig. (\ref{Net.}). The advantage of such a modus operandi is that we can adjust weights for the different functions. Without adjusting for weights, instead of Eq. (\ref{average}), one may use the more general average of 
\begin{eqnarray}
\label{average-general}
\mathcal{V}_i=  sgn \Big( \frac{1}{\mathcal{R}}\sum_{\alpha=1}^{\mathcal{R}}\sum_{k=1}^{N_f} y_{i,k,p}^\alpha \Big), 
\end{eqnarray}
where the predicted value $y_{i,k,p}^\alpha$ for replica $\alpha$ is found using Eq. (\ref{Map}) with kernel $K$ belonging to the $k-$th entry of the list
of Eq. (\ref{Klist}) or other trivial extensions thereof (and $N_{f}$ the total number of functions in such lists). The weight of one function may be adjusted as the calculation proceeds to be higher or lower to increase the accuracy. Without adjusting for weights, we observe in Figs. (\ref{ZZoom:Reg_1},\ref{ZZoom:Reg_2},\ref{ZZoom:Reg_3}) that the accuracy, run time, and coefficient of performance are similar to those that we obtained earlier (within a single layer voting model- the usual SRVM). A trivial extension of Eq. (\ref{average-general}) is that of the weight adjusted voting, 
\begin{eqnarray}
\label{ag+}
\mathcal{V}_i= sgn \Big(\sum_{\alpha=1}^{\mathcal{R}}\sum_{k=1}^{N_k} w_{k,\alpha} y_{i,k,p}^\alpha \Big),
\end{eqnarray}
with the weights $w_{k, \alpha}$ satisfying, $\sum_{k=1,\alpha}^{N_{f}} w_{k,\alpha} =1$. 

As we explained in Section \ref{overlap+}, one may aim to find the optimal parameters by noting when these lead to a maximal overlap between the replicas.
Additionally, of course, one may see when these lead to accurate solutions- yet that either requires ``cheating''- i.e., (1) adjusting 
the parameters to obtain the known answer or to (2) the removal of some of the known input data to use it as a CV test (the latter
case is non-optimal since already known data are removed from the training set). At any rate, testing for overlaps and/or direct
accuracies by brute force change of parameters can be taxing. An alternate approach for {\it determining the optimal parameters and weights} such as those of $w_{k,\alpha}$ in Eq. (\ref{ag+}) (and simple multi-layer generalizations thereof), somewhat similar to reinforcement 
learning \cite{RL}, is to compute the overlap and/or accuracies for a set of parameters and then {\it recursively} use SRVM to extrapolate and 
decide on the optimal parameters. Since the accuracies/overlaps are continuous variables,
this task lies in the domain of ``regression'' (the prediction of an outcome that is a continuous
variable). That is, by successively applying SRVM to the accuracy or replica overlap results we may hone in 
on the region of the parameters where optimal performance may occur (similar to ternary search algorithms). 

\begin{figure}[h]
\centering
\includegraphics[width=0.8\columnwidth]{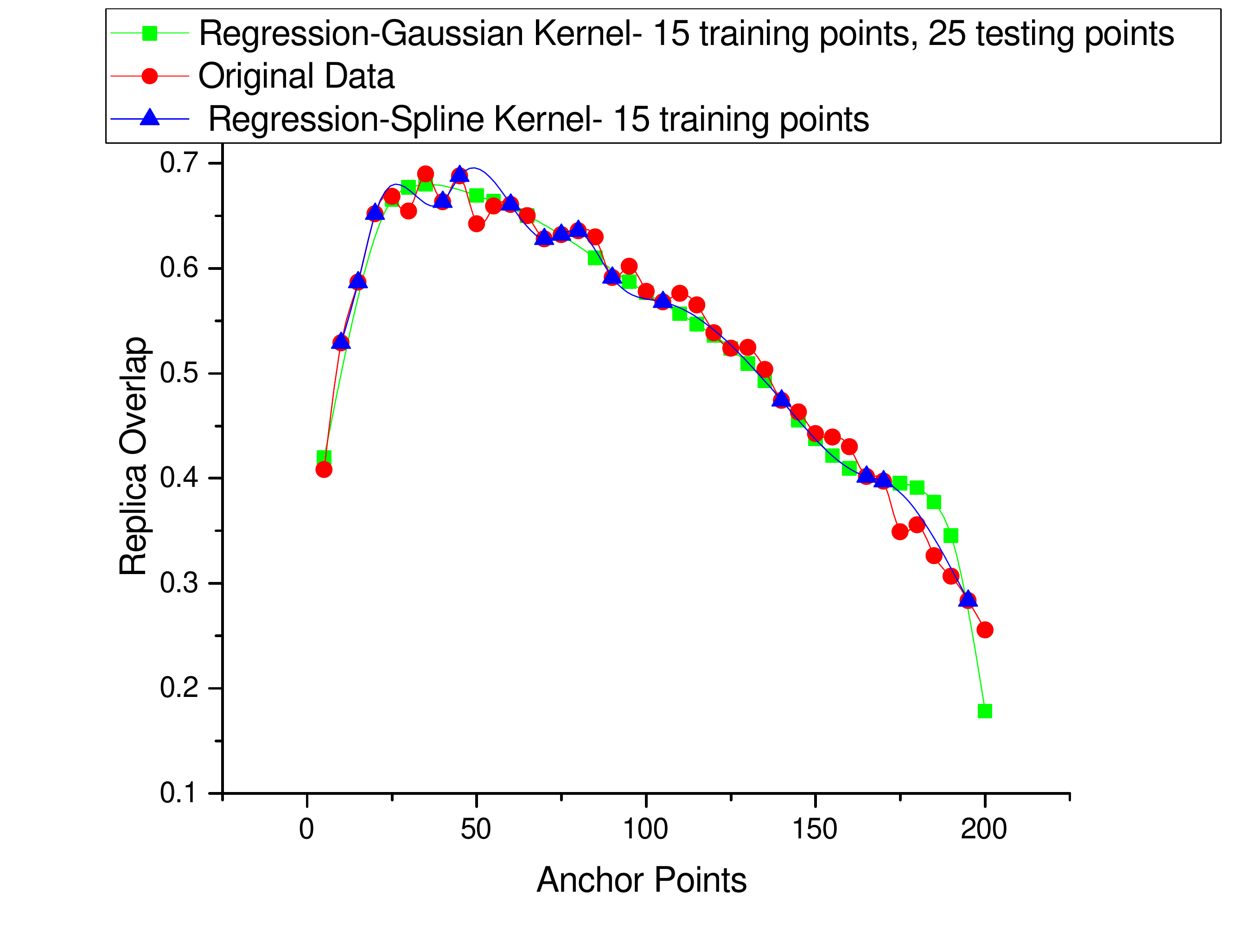}
\caption{(Color online.) Heart data set. 
We applied machine learning onto itself so as to determine the optimal parameters. Here, we provide the results of a regression analysis using
15 training points for  the inter-replica overlap as a function of the number of anchor points. 
Both the results of a Gaussian SRVM analysis with a simple cubic spline are contrasted with the empirical behavior. The points where optimal overlap
(and thus accuracy) appear in the regression curves do not greatly deviate from those in the empirical data.}
\label{Heart-selfo}
\end{figure}

\begin{figure}[h]
\centering
\includegraphics[width=0.8\columnwidth]{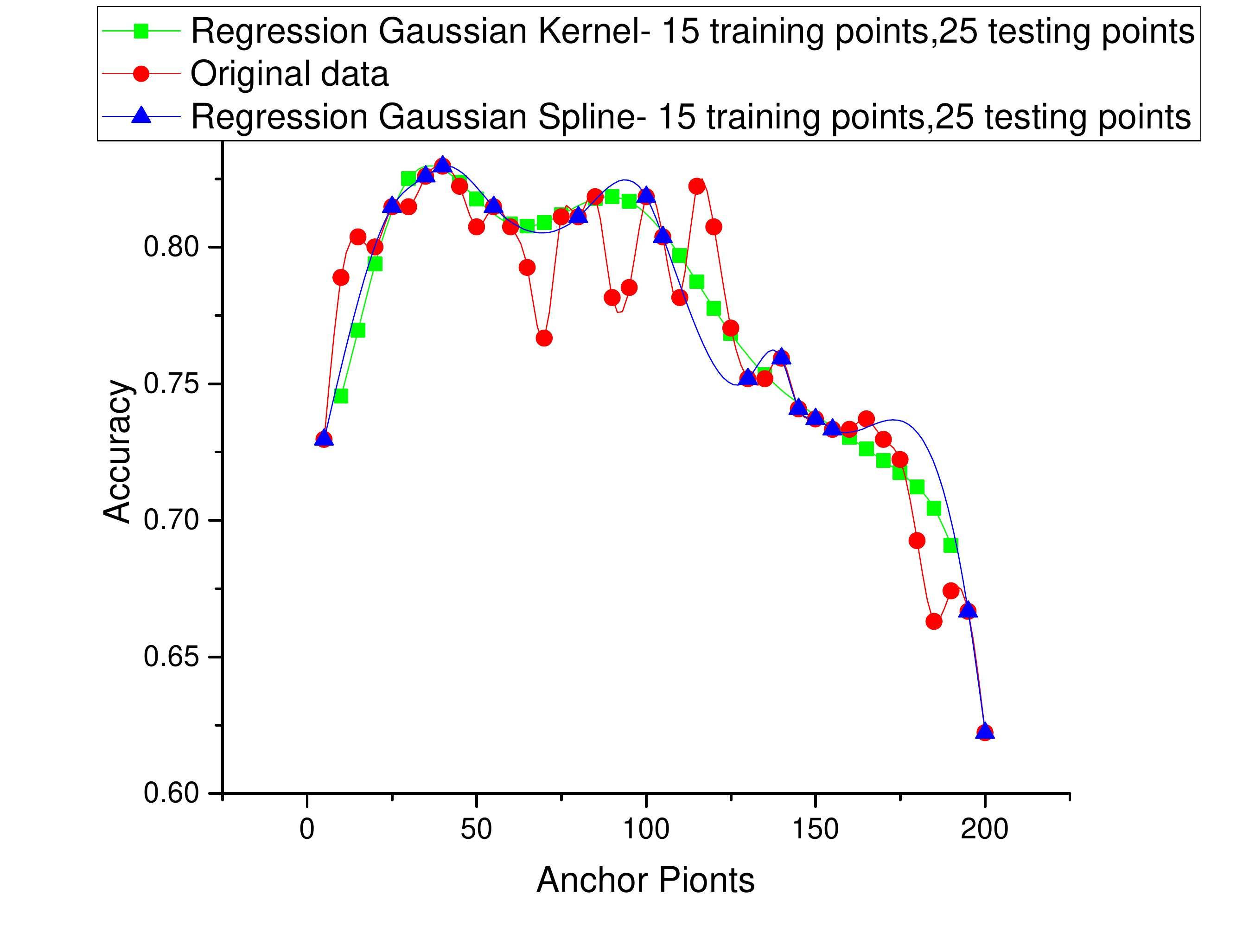}
\caption{(Color online.) Heart data set. 
Similar to Fig. (\ref{Heart-selfo}), here we apply machine learning onto itself so as to suggest the optimal number of anchor points for which the 
accuracy is maximal. We provide the results of a regression employing 
15 training points for the SRVM accuracy as a function of the number of anchor points. 
Both the results of a Gaussian SRVM analysis with a simple cubic spline are contrasted with the empirical behavior. The number of anchor points at which maximal accuracy is predicted to occur (from the regression analysis) indeed tracks the empirical data.}
\label{Heart-selfa}
\end{figure}

To illustrate the basic premise, we provide the results of such an analysis for suggesting the optimal 
number of anchor points $v$ for which the highest overlap and accuracy for the Heart data set appear. In Figs. (\ref{Heart-selfo}, \ref{Heart-selfa})
we show the results of two regression analysis using the {\bf (i)} Gaussian kernels of Eq. (\ref{Klist}) (with 10 anchor points $\{\vec{\chi}_j^\alpha\}_{j=1}^{10}$ in each of the replicas $\alpha$ used) and {\bf (ii)} cubic splines. Similar to the well known Runge phenomenon \cite{Runge} for high order polynomial fits for equally spaced points, we observed that when no clamping was done at the endpoints at the $v$ domain, these one-dimensional regression 
curves performed well throughout apart from the regimes near the endpoints. When we
fixed the values of the accuracies at the two endpoints and performed regression analysis
with either {\bf{(i)}} or {\bf{(ii)}} given 15 training points for different values of $v$, the resulting
curves were relatively close to the actual data for all $v=40$ points that we tested earlier.
Most importantly (see Figs. (\ref{Heart-selfo}, \ref{Heart-selfa})), the maxima of the regression curves were close to those found 
in the complete data set. This simple example illustrates the viability of using 
{\it{machine learning recursively onto itself}} in order to find the optimal parameters
that might maximize its accuracy. The parameters governing our algorithm
may be automatically optimized by such a recursive scheme.

We must remark that in this one-dimensional problem of determining the optimal number of anchor points $v$ for which maximal accuracy may appear, one may readily forgo the use recursive SRVM (or, similarly, any machine learning algorithm) and instead employ a simple ternary search to guide the search parameters for which maximal viable accuracy may be achieved. We next briefly discuss regression more generally.  

Thus, in conclusion, as we motivated above and illustrated for a simple example, we may bypass the need for an exhaustive parameter search and instead
{\it employ recursive machine learning to find the optimal parameters.}

\begin{figure}
	\centering
	\includegraphics[width=\linewidth]{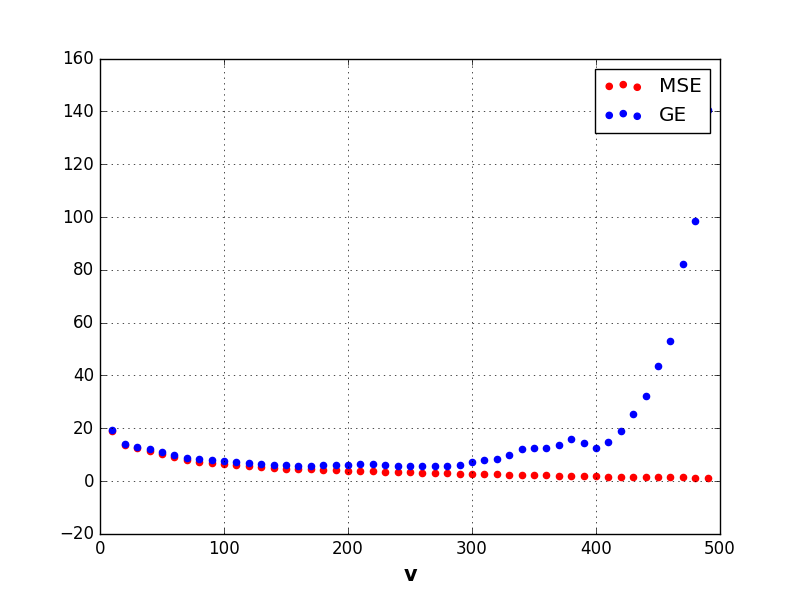}
	\caption{LVST data set. Plot of the mean standard training error (MSE) and the testing error (GE) for increasing number of parameters in the standard SRVM algorithm. The optimal number of parameters corresponds to the value of $v$ where GE and MSE have the lowest values for a single $v$. In this case that occurred at $v$=250.}
	\label{FVMAll.}
\end{figure}

\begin{figure*}
	\centering
	\begin{subfigure}[t]{.33\textwidth}
		\centering
		\includegraphics[width=\linewidth]{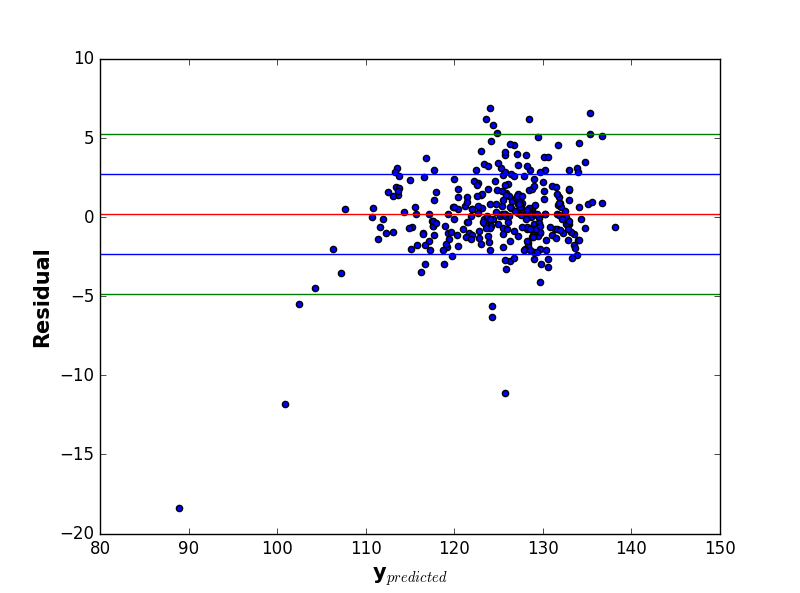}
		\caption{}\label{1Zoom:Reg_1}
	\end{subfigure}
	\begin{subfigure}[t]{.33\textwidth}
		\centering
		\includegraphics[width=\linewidth]{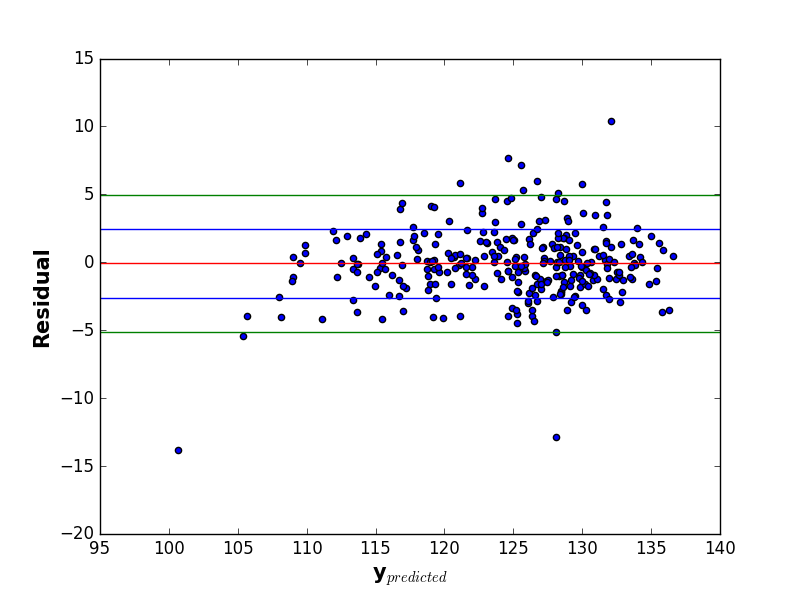}
		\caption{}\label{1Zoom:Reg_2}
	\end{subfigure}
	\begin{subfigure}[t]{.33\textwidth}
		\centering
		\includegraphics[width=\linewidth]{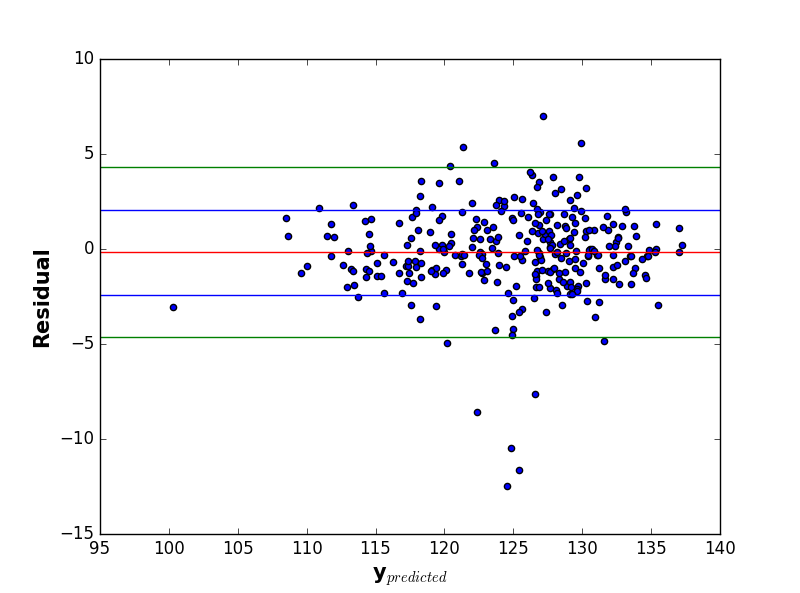}
		\caption{}\label{1Zoom:Reg_3}
	\end{subfigure}
	
	\medskip
	
	\begin{subfigure}[t]{.33\textwidth}
		\centering
		\vspace{0pt}
		\includegraphics[width=\linewidth]{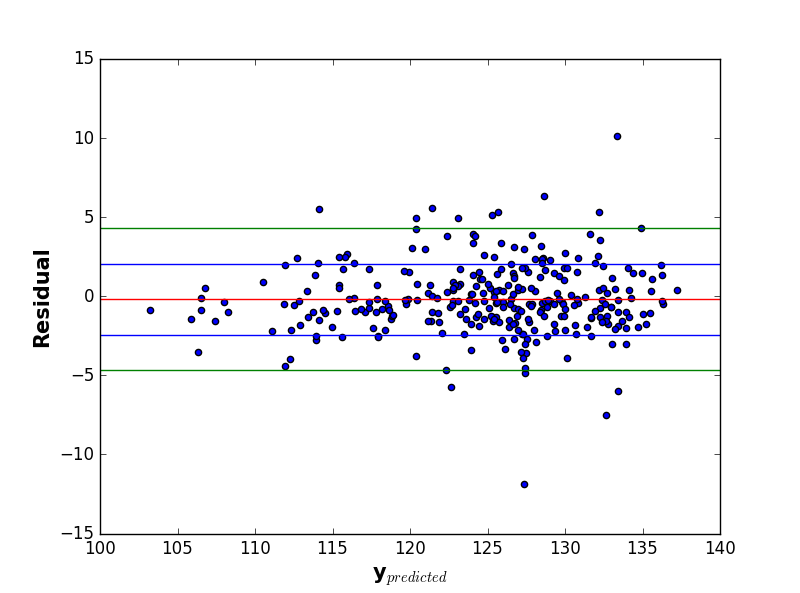}
		\caption{}\label{2Zoom:Reg_3}
	\end{subfigure}
	\begin{subfigure}[t]{.33\textwidth}
		\centering
		\vspace{0pt}
		\includegraphics[width=\linewidth]{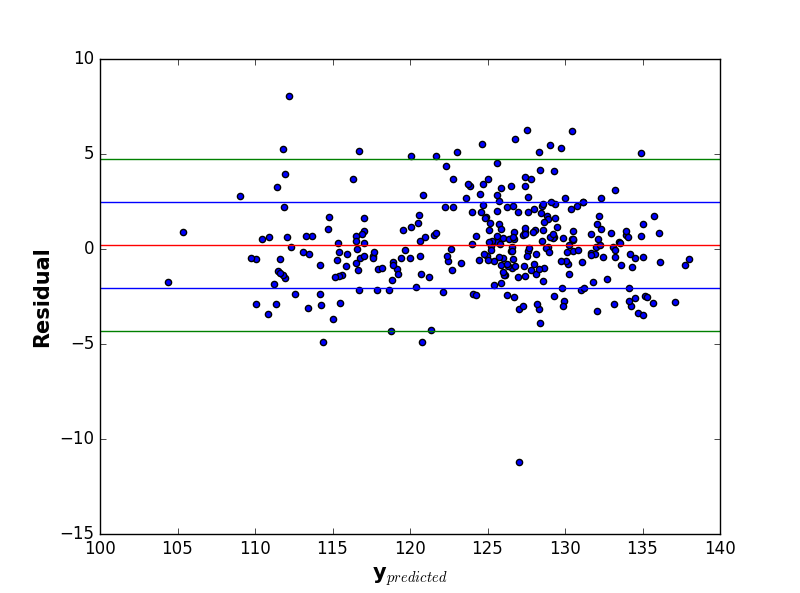}
		\caption{}\label{3Zoom:Reg_3}
	\end{subfigure}
	\begin{minipage}[t]{.33\textwidth}
		\caption{(Color Online.) Scatter plots of the residuals vs predicted value for cross validation of the SRVM regression algorithm applied to the LSVT data set with $v$=250 anchor points. The lines mark the locations of the mean (red), $\sigma$ (blue) and 2$\sigma$ (green). It is clear from the scatter plots, that the residuals of the SRVM regression model are randomly distributed and fall within the appropriate values (colored lines) for the normal distribution, suggesting accurate model performance.  \label{1Zoom}}
	\end{minipage}
\end{figure*}

\begin{figure*}
	\centering
	\begin{subfigure}[t]{.33\textwidth}
		\centering
		\includegraphics[width=\linewidth]{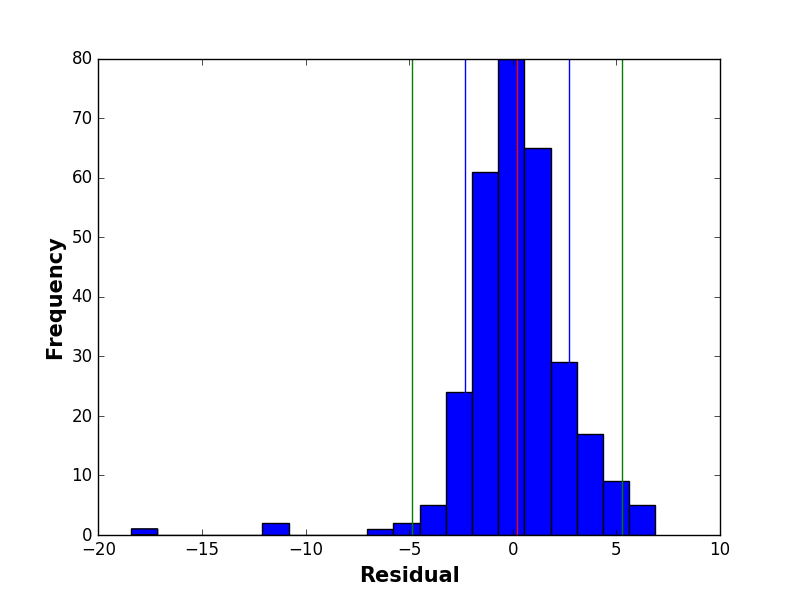}
		\caption{}\label{4Zoom:Reg_1}
	\end{subfigure}
	\begin{subfigure}[t]{.33\textwidth}
		\centering
		\includegraphics[width=\linewidth]{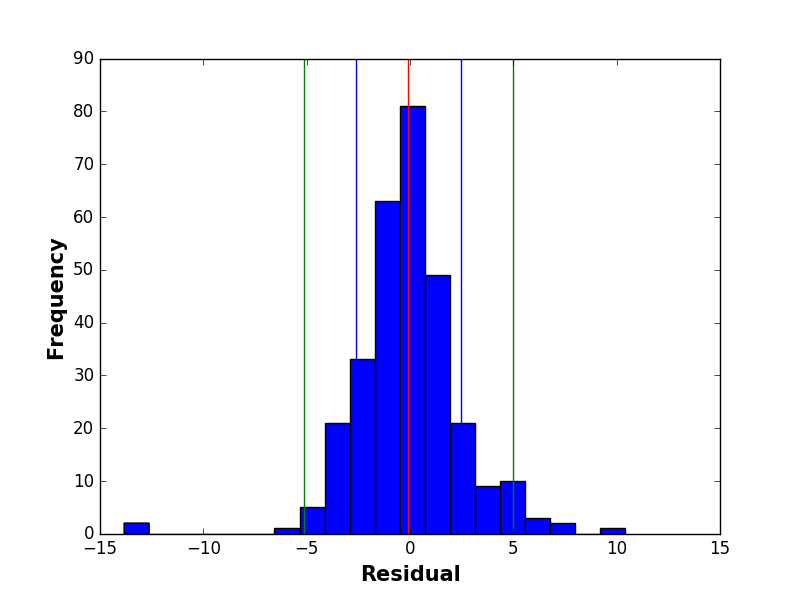}
		\caption{}\label{4Zoom:Reg_2}
	\end{subfigure}
	\begin{subfigure}[t]{.33\textwidth}
		\centering
		\includegraphics[width=\linewidth]{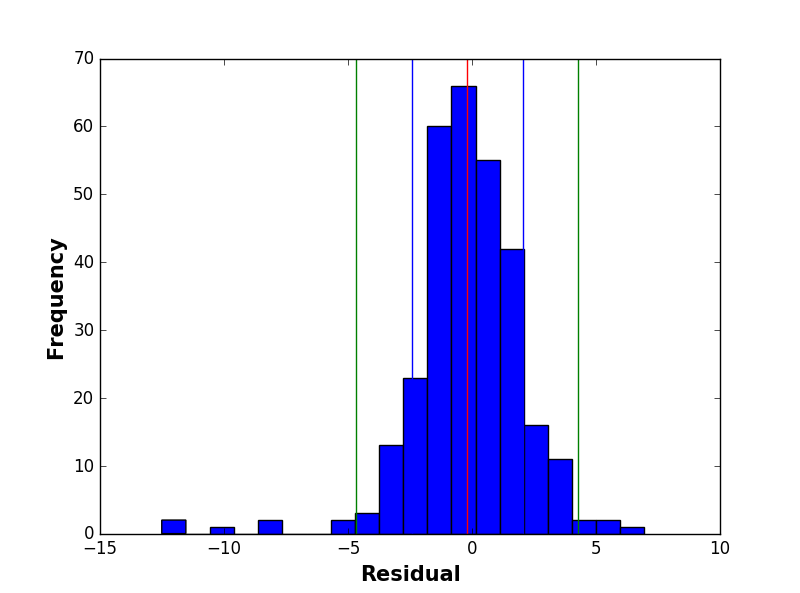}
		\caption{}\label{4Zoom:Reg_3}
	\end{subfigure}
	
	\medskip
	
	\begin{subfigure}[t]{.33\textwidth}
		\centering
		\vspace{0pt}
		\includegraphics[width=\linewidth]{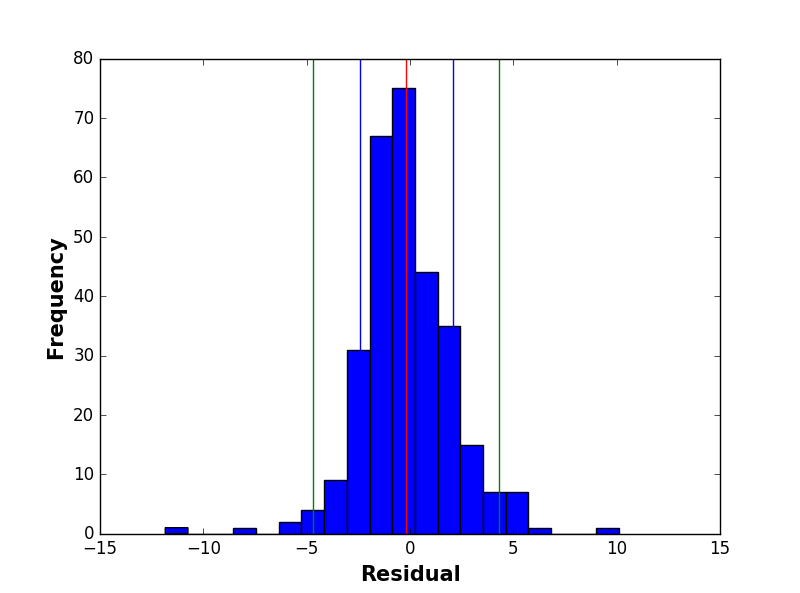}
		\caption{}\label{4Zoom:Reg_4}
	\end{subfigure}
	\begin{subfigure}[t]{.33\textwidth}
		\centering
		\vspace{0pt}
		\includegraphics[width=\linewidth]{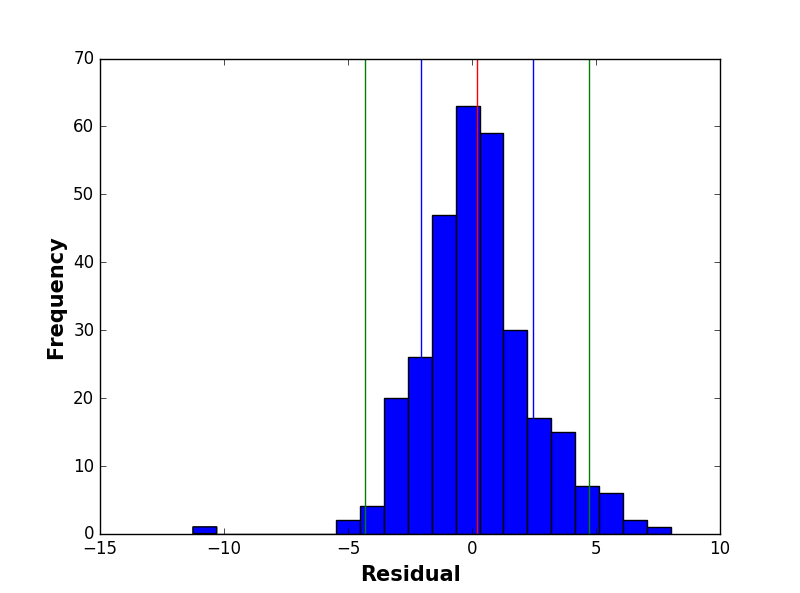}
		\caption{}\label{4Zoom:Reg_5}
	\end{subfigure}
	\begin{minipage}[t]{.33\textwidth}
		\caption{(Color Online.) \label{2Zoom} Histograms of the residuals for five-fold cross validation testing of the SRVM regression algorithm for the LSVT data set with $v$=250 anchor points. The vertical lines mark the locations of the mean (red), $\sigma$ (blue) and 2$\sigma$ (green).  Based on the histograms, the residuals appear to be roughly normally distributed with the exception of some slight skewing in the tails. This result indicates a strong performance of the model.}
	\end{minipage}
\end{figure*}

\begin{figure*}
	\centering
	\begin{subfigure}[t]{.33\textwidth}
		\centering
		\includegraphics[width=\linewidth]{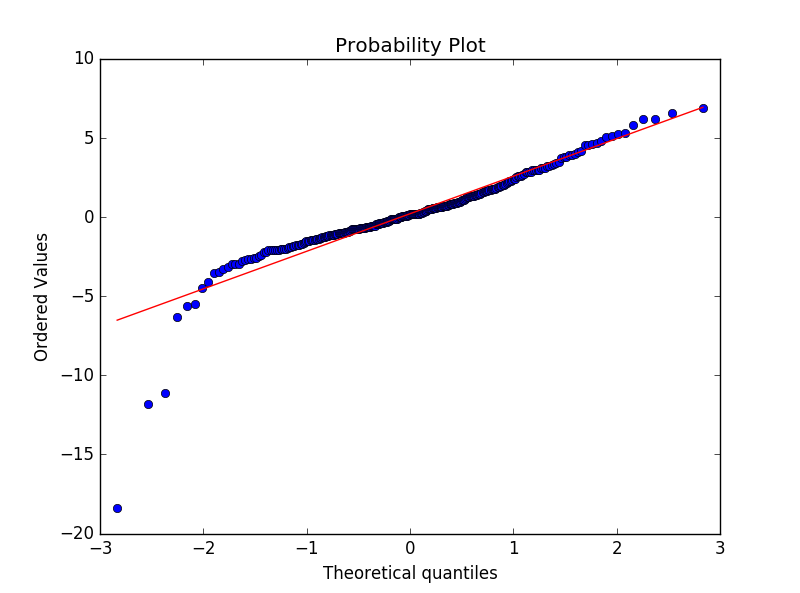}
		\caption{}\label{6Zoom:Reg_1}
	\end{subfigure}
	\begin{subfigure}[t]{.33\textwidth}
		\centering
		\includegraphics[width=\linewidth]{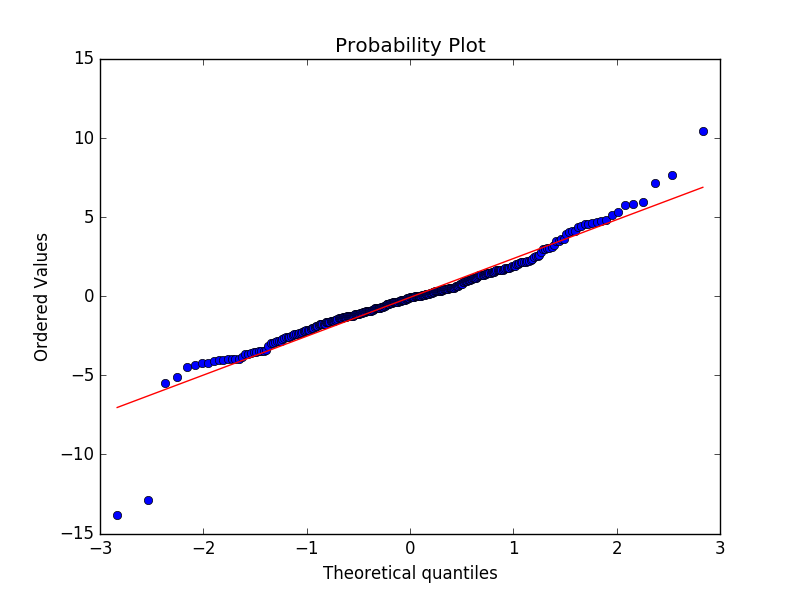}
		\caption{}\label{6Zoom:Reg_2}
	\end{subfigure}
	\begin{subfigure}[t]{.33\textwidth}
		\centering
		\includegraphics[width=\linewidth]{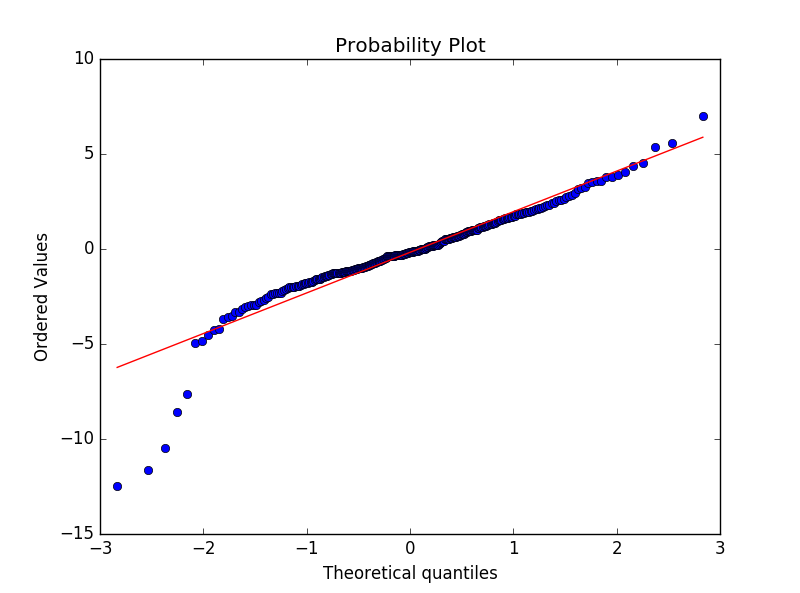}
		\caption{}\label{6Zoom:Reg_3}
	\end{subfigure}
	
	\medskip
	
	\begin{subfigure}[t]{.33\textwidth}
		\centering
		\vspace{0pt}
		\includegraphics[width=\linewidth]{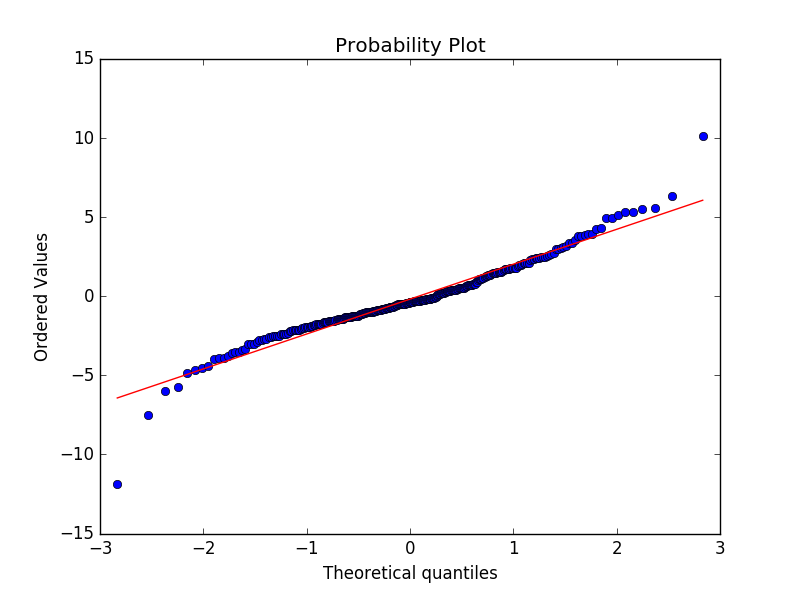}
		\caption{}\label{6Zoom:Reg_4}
	\end{subfigure}
	\begin{subfigure}[t]{.33\textwidth}
		\centering
		\vspace{0pt}
		\includegraphics[width=\linewidth]{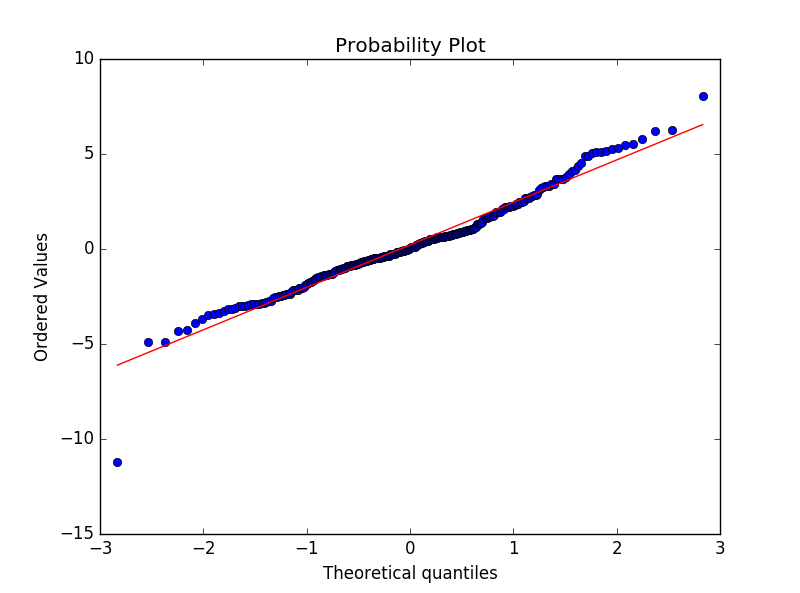}
		\caption{}\label{6Zoom:Reg_5}
	\end{subfigure}
	\begin{minipage}[t]{.33\textwidth}
		\caption{(Color Online.) Probability plots of the residuals for cross validation testing of the SRVM regression algorithm applied to the LSVT data set with $v$=250 anchor points. The quantiles of the residuals are plotted against the expected quantiles of the normal distribution. With the exception of minor skewing in the tails, the quantiles appear normal, suggesting strong model performance. \label{3Zoom}}
	\end{minipage}
\end{figure*}

\section{Regression via replicas}
\label{sec:srvr}

The functions $f$ and the associated kernels $K$ that we presented in Section \ref{nut} are continuous. Thus, on an intuitive level, they are more naturally related to continuous value prediction 
(a regression) than to discrete quantifier (such as those associated with the classification problems that we considered in earlier Sections). 

In this Section, we will explicitly examine whether SRVM is indeed a viable regression machine learning algorithm. As we will explain
below, what is important for a regression solver is to have normal statistical properties. Indeed, for regression problems, there are no ``benchmarks'' that are as clearly defined for continuous regression data as they are for discrete classifiers (where an answer is clearly wrong or right). With this in mind, 
we examined the features of the LSVT data set (for which we developed a binary classifier) and tested to see whether
our predictors $f^{\alpha}(\vec{x})$ (without the thresholding of Eq. (\ref{Thresh})) comprise good (continuous) regression
predictors to the binary data. 

A natural route for SRVM is to expand in kernels of the full vectors $\vec{x}$ (as in the kernels of Eq. (\ref{Klist}) employed in the expansion of Eq. (\ref{Map}) that we have largely used in the current work with the exception of the multinomial kernels of Eqs. (\ref{multi}, \ref{multig}). However, the manner in which a regression is usually performed for other existing machine learning approaches is different. Instead of expanding in functions of all $d$ components (features of a vector) $\vec{x}$, most researchers typically examine functions of individual features, e.g., for $d$ features. That is, one typically posits that a function $\sum_{l=1}^{d} c_{l} f_{l}(x_{l})$ (instead of considering functions of whole instance feature vectors $\vec{x}$) may be the optimal kernel to use. Underlying this common practice of single feature expansion is that multicollinearity is assumed and then this assumption may be consistently tested for; this also enables a study of the individual significance of a given feature.
Translated into our framework, such a regression is tantamount to an expansion of the form
\begin{eqnarray}
{y_{i,p}^\alpha \equiv f^{\alpha}(\vec{x}_i) \equiv \sum_{l=1}^{d} \sum_{j=1}^{v} c_{lj}^\alpha ~K_{l}(x_{li},\{\vec{\chi}_j^\alpha\})},
\label{MapRAlt}
\end{eqnarray}
where $x_{li}$ denotes the $l$-th feature (component) of the vector $\vec{x}_{i}$.

In our regression studies, we performed regression in both ways. That is,  \newline

{\bf(A)} We expanded in uniform kernels of all $d$ components vectors $\vec{x}$, and  \newline

{\bf(B)} Similar to Eq. (\ref{multi}), we also examined the system when expanding in kernels of individual functions of the single components $x_{l}$ (single features of the $d-$ component vectors $\vec{x}$). \newline

In our regression analysis, we mainly focused on method {\bf(A)} (that of expanding in functions of the full feature vector $\vec{x}$ as in Eq. (\ref{Klist})). When searching for optimal parameters, one has to focus on the mean squared error versus the generalization error since the mean squared error (as is visible in Fig. (\ref{FVMAll.})) will always decrease with more anchor points $v$. However, the generalization error will increase dramatically when overfitting occurs beyond a threshold number of anchor points. Inspecting Fig. (\ref{FVMAll.}), we may ascertain the optimal number $v$ of anchor points. 

Assessing the quality of a regression is notably more challenging than determining the accuracy in a classification problem. The predicted outcome is clear cut for the discrete variable in a classification problem; this is obviously not the case for regression outcomes which are continuous functions. Instead of seeing whether the ``exact'' outcome is achieved (an impossible feat for continuous real numbers), additional, more detailed checks, are necessary. The commonplace minimization of the sum of square errors is indeed how we found the optimal parameters (that are used in the plots). As is well appreciated, the raw sum of square errors is not a sufficiently illuminating metric for judging the quality of regression solvers. 

For a regression to perform optimally, aside from predicting results that are close to the correct answers, its residuals (errors) should be random and normally distributed about the true population; the residuals should have no autocorrelation (no bias of one data point influencing another). In Figs. (\ref{1Zoom},\ref{2Zoom}), we provide scatter plots and histograms of the residuals with the mean in red, standard deviations in blue and green.
it is seen that the histograms are very normal. One may also examine the probability plots of Fig. \ref{3Zoom}; 
apart from skewing at the tails, normal residuals indicate high quality regression. Autocorrelation statistical tests further suggest that no significant autocorrelations are present. All of our tests indicate that no bias is present in our regression.

\section{Possible algorithm independent bounds on the accuracy}
\label{al_in}

\begin{figure*}
	\centering
	\begin{subfigure}[t]{.33\textwidth}
		\centering
		\includegraphics[width=\linewidth]{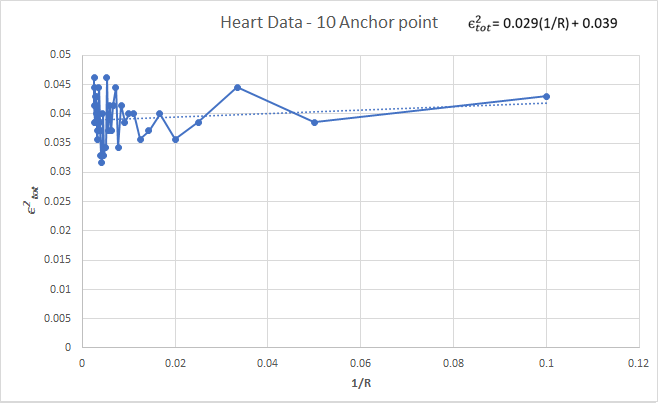}
		\caption{}\label{Heart-error_1}
	\end{subfigure}
	\begin{subfigure}[t]{.33\textwidth}
		\centering
		\includegraphics[width=\linewidth]{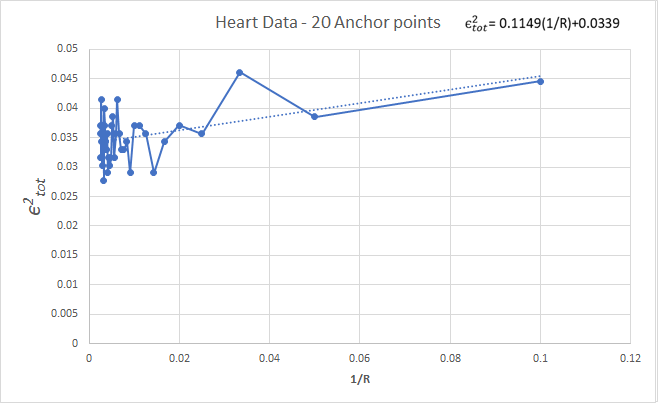}
		\caption{}\label{Heart-error_2}
	\end{subfigure}
	\begin{subfigure}[t]{.33\textwidth}
		\centering
		\includegraphics[width=\linewidth]{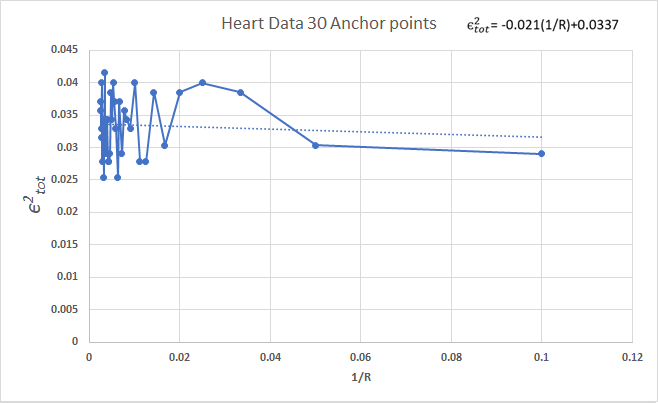}
		\caption{}\label{Heart-error_3}
	\end{subfigure}
	
	\medskip
	
	\begin{subfigure}[t]{.33\textwidth}
		\centering
		\vspace{0pt}
		\includegraphics[width=\linewidth]{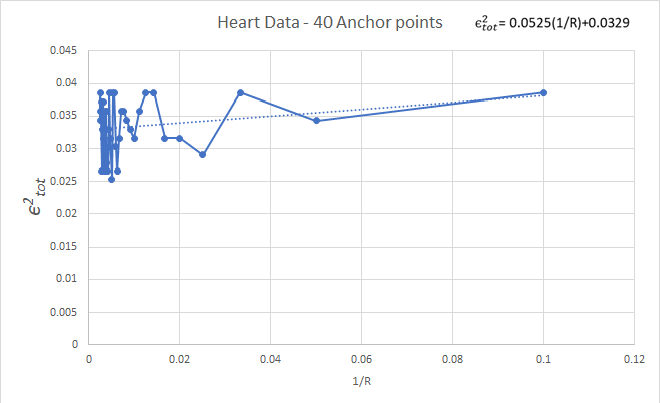}
		\caption{}\label{Heart-error_4}
	\end{subfigure}
	\begin{subfigure}[t]{.33\textwidth}
		\centering
		\vspace{0pt}
		\includegraphics[width=\linewidth]{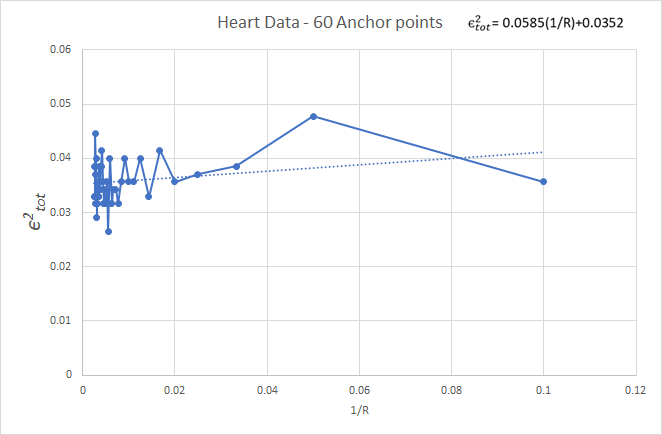}
		\caption{}\label{Heart-error_5}
	\end{subfigure}
	\begin{minipage}[t]{.33\textwidth}
		\caption{ \label{Heart-error} The dependence of the error rate $\epsilon_{tot}$ on the number of replicas for Heart data. The dependence has been shown for different numbers of anchor points. The linear curves are the prediction of Eq. (\ref{tsmt'}).}
	\end{minipage}
\end{figure*}

\begin{figure*}
	\centering
	\begin{subfigure}[t]{.33\textwidth}
		\centering
		\includegraphics[width=\linewidth]{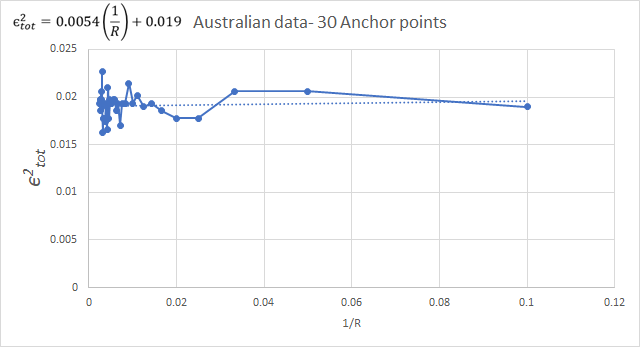}
		\caption{}\label{Aus-error_1}
	\end{subfigure}
	\begin{subfigure}[t]{.33\textwidth}
		\centering
		\includegraphics[width=\linewidth]{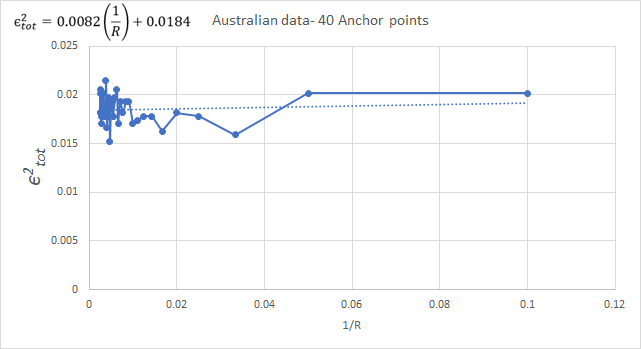}
		\caption{}\label{Aus-error_2}
	\end{subfigure}

	\begin{subfigure}[t]{.33\textwidth}
		\centering
		\vspace{0pt}
		\includegraphics[width=\linewidth]{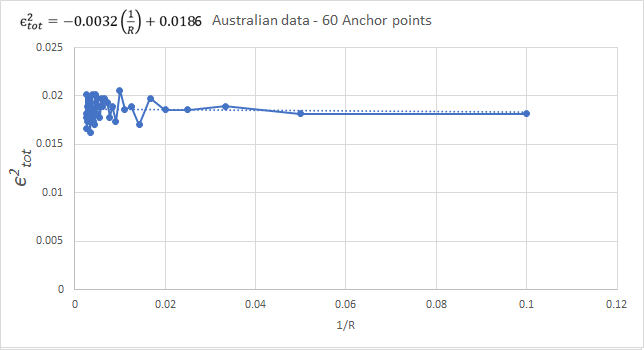}
		\caption{}\label{Aus-error_4}
	\end{subfigure}
\medskip
	\begin{subfigure}[t]{.33\textwidth}
		\centering
		\vspace{0pt}
		\includegraphics[width=\linewidth]{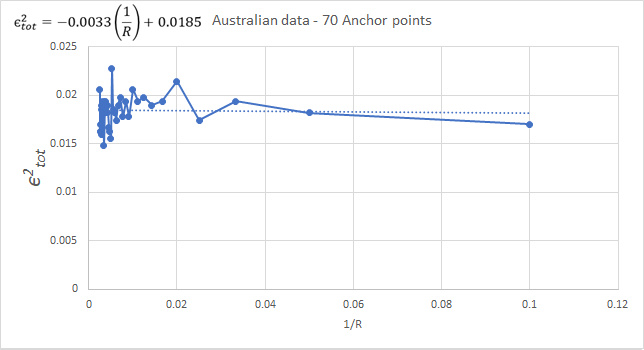}
		\caption{}\label{Aus-error_5}
	\end{subfigure}

	\begin{subfigure}[t]{.33\textwidth}
		\centering
		\vspace{0pt}
		\includegraphics[width=\linewidth]{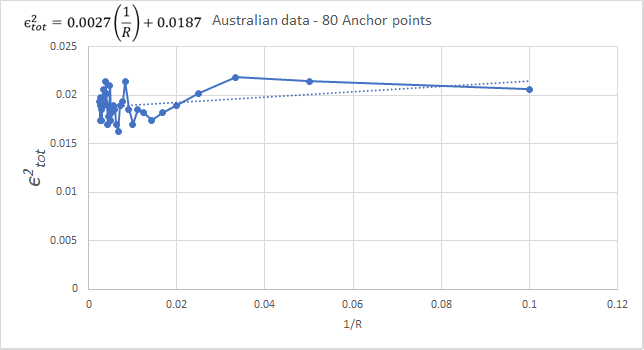}
		\caption{}\label{Aus-error}
	\end{subfigure}

	\begin{minipage}[t]{.33\textwidth}
		\caption{ \label{2Zoom} The dependence of the error rate on the number of replicas for Australian data. The dependence has been shown for different numbers of anchor points. }
	\end{minipage}
	\caption{ \label{error-Australian} The dependence of the error rate $\epsilon_{tot}$ on the number of replicas $R$ for the Australian data set. Figures are displayed for a varying numbers of anchor points. }\label{error-Austrlian}
\end{figure*}

As is well appreciated, different machine learning methods have their distinct virtues. Some methods (such as artificial neural networks) seem to work ``magically'' well on a large number of data sets for reasons that, to date, largely remain shrouded in mystery. In this Section, we wish to suggest that SRVM may lead to universal asymptotic limits on the accuracy of these and all algorithms. The logic underlying this speculation is as follows. In reality, any physical process exhibits an inherent error. That is, there is an underlying ``theory'' or ``model independent'' error to any physical process. To give a colloquial example: suppose that we knew all of the scores in various matches. Even with much knowledge on the scores of the final games in all prior matches and information about individual players, it still is, of course, impossible to predict with certainty what the result of a new soccer match will be. There is, even in ``classical'' systems such as soccer games an important element of stochasticity including pure ``luck''. That is, any given finite set of features will be insufficient to provide error free predictions no matter how complex our algorithm may be. As in putative physical theories, the total error associated with a given theory vis a vis the measured data will be generally the simple sum,
\begin{eqnarray}
\label{tsmt}
\epsilon_{tot}^{2} = \epsilon_{measurement}^{2} + \epsilon_{theory}^{2}.
\end{eqnarray}
Here, $\epsilon_{tot}$ denotes the total error of the prediction vis a vis the measured data, $\epsilon_{measurement}$ is the inherent underlying measurement error (including any external noise that is out of our control and, employing features that are incomplete and cannot allow an accurate prediction even if they were all known with absolute precision), and $\epsilon_{theory}$ is the error in the theory or machine learning predictor that we use. 
The two diametrically opposite limits of Eq. (\ref{tsmt}) are intuitively clear. If, one is given the accurate correctly measured features with which predictions may, in principle, be made with absolute certainty (i.e., 
$\epsilon_{measurement} =0$) then all error in the prediction is due the use of an inaccurate theory ($\epsilon_{tot} = \epsilon_{theory}$). At the other opposite extreme, if one has the correct theory (e.g., equations that correctly describe physical processes) then any error in the predictions will be due to either incomplete or inaccurate input data ($\epsilon_{tot} = \epsilon_{measurement}$). The latter errors are not limited to literal physical measurements alone. For instance, if there are numerous spin glass \cite{SG1} ground states that are consistent with any given the assignments of a small finite number of spins then one may not accurately predict the spin at each site due to the underlying degeneracy \cite{SG2,BinomialSG}- the given features do not suffice for such a complete prediction of the ground state that one is asked to predict. Now, if the errors in $R \gg 1$ different theoretical models (``replicas'' in the parlance of SRVM) are independent of each other then error in the average prediction of the collection of $R$ theories (each with individual theoretical error $\epsilon_{a}$), i.e., the SRVM prediction, will, by the central limit theorem, scale
as 
\begin{eqnarray}
\label{tsmt'}
\epsilon_{tot}^{2} = \epsilon_{measurement}^{2} + \frac{\sum_{a=1}^R \epsilon_{a}^{2}}{R} .
\end{eqnarray}
A consequence if Eq. (\ref{tsmt'}) is that by increasing the number of replicas $R$, the error rate decreases and converges to the model independent value $\epsilon_{measurement}$ (the inherent stochastic error underlying the systems and the features employed). Of course, some theories may be more powerful than others and lead to better predictions. However, if Eq. (\ref{tsmt'}) is correct, then asymptotically, irrespective of which algorithm is employed for the individual replicas, a universal limiting error $\epsilon_{measurement}$ will be reached. In the context of the examples studied in this work, if we test theories with a different number of anchor points then some may outperform others for given number of replicas yet in the large replica limit, all SRVM averages should coincide on the same answer and associated errors. We qualitatively tested to see if this prediction might be consistent with our data. Specifically,
in Fig. \ref{Heart-error} and Fig. \ref{error-Australian}, we respectively fit the SRVM cross-validation errors (regarded here as $\epsilon_{tot}$) of the Heart and Australian data sets to Eq. (\ref{tsmt'}). A general monotonically decreasing trend of the average accuracy is seen with increasing replica number (apart from situations in which the accuracy is already near optimal when $R$ is small).
More notably, the extrapolated $1/R \to 0$ intercept in these figures seems to be uniform across all different anchor point solvers. If Eq. (\ref{tsmt'}) is correct then these intercepts correspond to the limiting stochastic error $\epsilon_{measurement}$ of the system that no algorithm can surpass.  One cannot, of course, assert that these results demonstrate the validity of our conjecture. However, the observed behaviors are consistent with Eq. (\ref{tsmt'}). 

\section{Conclusions} \label{sec:conclusion}

In summary, we presented a new machine learning algorithm- the ``Stochastic Replica Voting Machine'' (SRVM)- that largely drew its inspiration from 
Landau theories and the existence of various continuous real function fits to the same existing data sets along with the ``wisdom of the crowds'' phenomena. This method uses expansions of data via kernels as do many other algorithms, e.g., \cite{kernel-review}, including SVM.  However unlike SVM, our method does not follow the error optimization and regularization scheme.  The guiding principle underlying our approach is that of invariance and stability of the predicted results when different {\it random functions} are used. In the context of the results presented in the current work, known data are fitted to fix the kernel coefficients in multiple stochastic functions. Once these coefficients are fixed, predictions are made by the ensemble of these random functions (``replicas'') as to the correct classification/regression of new data. Each of the functions ``votes'' for the predicted outcome. The system then averages over all predictions by weighing these in a chosen manner. We tested the algorithm's performance on multiple known benchmark problems. Overall, we found the accuracy of our algorithm to be comparable to that of standard well used techniques such as those of Support Vector Machines (SVM). By contrast to SVM, however, the optimal parameters in our model are set by using all of the given data (not tossing away a subset of these when using cross-validation). In our framework, the optimal parameters are ascertained by observing when the different stochastic functions (the ``replicas'') have a high degree of overlap. That is, we see for which parameters there exists a consensus in the predictions of the replicas regarding the outcome for particular data. No less notably, due to the intrinsic stochastic character of multiple functions used, the system is far superior to SVM in avoiding class imbalance bias. Similar to ``Random Forest Decision Tree'' \cite{decisions} algorithms, SRVM invokes voting between different predictions. However, contrary to numerous random forest and neural network methods, SRVM {\it does not} introduce ``decision trees''. Rather, in SRVM, actual real functions (such as those of Eq. (\ref{Klist})) are employed. If, from physics based or other considerations, information is known about the expected functional dependence of the results on the input features then one may expand in a basis of stochastic functions of that expected form instead of employing the generic functions that we used in the current work. We remark that the use of multiple classifiers (different from our stochastic functions) to enhance accuracy further appears in other machine learning approaches such as those of unweighted ``bagging'' \cite{bag} or more sophisticated ``boosting''' \cite{boosting} methods that have been prevalent in, e.g., neural networks; it is conceivable that our accuracy might be further improved by incorporating aspects of these schemes when combining the bare SRVM algorithm that we described in the current work with other known classifiers. Indeed, as we detail elsewhere \cite{us-materials}, a function describing any particular neural network can be regarded as yet another member of the ensemble of functions used in an SRVM implementation.  We stress that unlike Markov Chain Monte Carlo (MCMC) \cite{mcmc} methods, the crux of our general approach hinges, in the absence of given special details, on the use random stochastic functions of different types (not that of sampling from a single distribution function). Contrary to MCMC, neural networks (including deep learning) \cite{ANN,DL}, and many other methods, the number of parameters in our simple approach is rather limited. Optimizing for these few parameters (in our case the number of anchor points) can be done in an automated way (see Section \ref{layer:sec}). That is, the correct parameters to be used do not need to be introduced by human training but are rather {\it self-generated}. Not much training is required in order to achieve high accuracy. For all of the examples that we studied, we did not find a significant difference in accuracy between instances in which (1) direct arithmetic averages of the individual replica predictions (Eq. (\ref{average})) were used and (2) when individual replica predictions are weighted differently depending on, e.g., how close these are to the anticipated correct classification values (e.g., the individual replica predictions that are closer to $\pm 1$ values for the binary classification problem may, in the final vote, be given higher weight over other predictions when those predictions have a modulus that is very different from unity). That is, we found that the results for the examples that we studied were largely insensitive to the particular choice of voting function of the individual replica predictions $\{y^{\alpha}(\vec{x})\}_{\alpha =1}^{\mathcal{R}}$. The accuracies that we obtained when using disparate voting functions for the Heart benchmark are provided in Tables \ref{tab:problemss1} and \ref{tab:problemss2}.

\begin{table*}[tt]
	\begin{tabular}{| c | c | c | c | c | c | c |}
		\hline
		Voting Function $F(f^{\alpha})=$ & $f^{\alpha}$ & $sgn(f^{\alpha})$ & $Logistic(f^{\alpha})$ & $\tanh(10f^{\alpha})$ & $\tanh(100f^{\alpha})$& $\tanh(1000f^{\alpha})$\\ 
		\hline
		Accuracy & 0.8059 & 0.8111 & 0.8107 & 0.8081 & 0.8051 & 0.8059\\
		\hline

	\end{tabular}
	\caption{Heart data set. The individual replica outcomes $f^{\alpha}$ are given by Eq. (\ref{Map}) with the Gaussian kernel of Eq. (\ref{Klist}) that was largely used throughout this work. In this table, we provide a comparison of the resulting accuracies when different voting functions $F(f^{\alpha})$
	are used to provide a final predicted answer that is set to be
$sgn(\sum_{\alpha=1}^{\mathcal{R}} F(f^\alpha(\vec{x})))$. The above functions $F$ replace, in all but one case (namely that of $F(f^{\alpha}) = sgn(f^{\alpha})$, the voting of Eqs. (\ref{Thresh},\ref{average}) that we employed in the current work. The differences between the accuracies are statistically insignificant.}
	\label{tab:problemss1}
\end{table*}

\begin{table*}[ttt]
	\begin{tabular}{| c | c | c | c | c | }
		\hline
		Gaussian Weighting  & $\sigma$ = 1 & $\sigma$ = 10 & $\sigma$ = 100 & $\sigma$ = 1000 \\
		\hline
		Gaussian Weighted Accuracy & 0.7606 & 0.7988 & 0.8051 & 0.8109 \\
		\hline
		
	\end{tabular}

	\caption{Heart data set. A comparison of the accuracies when different Gaussians of varying standard deviations are employed as the voting functions $F$ (see the caption of  Table \ref{tab:problemss1}). In this particular case, the voting function $F(f^{\alpha}) = e^{-(f^{\alpha}-1)^{2}/(2\sigma^{2})}- e^{(f^{\alpha}+1)^{2}/(2 \sigma^{2})}$. Apart from the $\sigma=1$ instance, the resulting accuracies are statistically the same.}
	\label{tab:problemss2}
\end{table*}

As we alluded to in Section \ref{layer:sec}, for more complex problems, one may envision {\it applying machine learning onto itself} functions; such a recursive modus operandi may enable one to potentially determine the optimal manner in which voting is to be taken from the individual replica results. Given the simplicity of our algorithm and its numerous natural extensions, much more work can be done to further streamline the algorithm and apply it to many different data sets. Aside from the numerous data set benchmarks tested in the current work, two additional materials oriented classification problems (both binary and ternary) were studied in \cite{us-materials}. The current results of our supervised machine learning study augment those of an earlier replica type approach for unsupervised learning and the solution of combinatorial problems in which the notions of stability and (potentially recursive) voting or information theory correlations/inference were employed \cite{CD3,CD4,vision,vision1,vision2,phase1,phase2,TSM,mychapter}. We may, indeed, very readily combine our supervised machine learning approach with clustering ideas for unsupervised machine learning. In this latter clustering approach, instead of minimizing errors between the predictions of random functions on known training data vis-\`a-vis the known outcomes (as is done for supervised machine learning), 
we may minimize an energy function that favors clustering of feature space points (or pixels in image segmentation applications) that share similar features (and maximize information theory correlations of candidate replica solutions). Thus, instead of minimizing the cost function measuring the quality of fits relative to known training points, data may be classified via the added contributions of the trained random SRVM kernels augmented by additional weights that measure the correlation between different points ${\vec{x}}$ in feature space that are classified as belonging to the same set (as in the unsupervised clustering approach of \cite{CD3,CD4,vision,vision1,vision2,phase1,phase2,TSM,mychapter}). There are many other natural extensions of the method presented here. For instance, instead of the linear expansions in the kernel functions (Eq. (\ref{Map})), one may, of course, consider higher order expansions in the kernels. A notable advantage of the SRVM method is that it approximates the data by mathematical functions of the input features that may, hopefully, be rationalized for (instead of more abstract constructs). Another possible advantage of SRVM is that it may enable the generation of algorithm independent bounds on the accuracy (as suggested in Section \ref{al_in}).  

We conclude with a brief speculation. As we repeatedly underscored (and do so once again now), the key notion behind our approach was that of stability. Stochastic functions were employed as individual predictors. If these stochastic predictors all consistently agreed on the same classification (or regression) of a given data point $\vec{x}$  then, as our inter-replica overlap analysis demonstrated, these predictions were all likely to be correct. Given this correlation, one may turn this result on its head and ponder whether fundamental physical theories are, in effect, Landau type theories in disguise and if spatio-temporal coordinates are not absolute (up to standard covariant transformations) but are rather assigned emergent features enabling the most consistent predictions concerning the behaviors of these systems. 
If a particular physical Lagrangian or effective Landau theory with generic low order polynomial coupling terms and gradients is assumed to describe a particular system then one may assign coordinates to various data points such that the predictions using this action are the most stable (such a possibility is similar to the reassignment of
fitness variables in chemical analysis \cite{permutation}). That is, we ask if the abstract features $\vec{x}$ in unsupervised learning (in which the pertinent features are inferred) might also correspond to true physical coordinates such that the ensuing representation of the data concerning particle coordinates is smooth thus enabling a description of its behavior by low cost of simple (Lagrangian, energy, or other) and generically low order smooth 
functions ${\cal{F}}(\phi (\vec{x}), \partial^{k} \phi(\vec{x}))$ of the collective descriptors $\phi$ 
and their gradients relative to the individual coordinates (features) $x_{k}$. \cite{qi}

\section*{ACKNOWLEDGMENTS}

We are grateful to the support of the National Science Foundation under grant number
NSF DMR-1411229. ZN is also grateful to the Aspen Center for Physics, which is 
supported by National Science Foundation grant PHY-1607611, where
this work was completed.


\begin{thebibliography}{99}
\bibitem{artifact} In the ``Heart'' data set, as the number of anchor points increases the accuracy decreases due to overfitting.
When the number of anchor points becomes equal to the number of training points in the data  (i.e., when $v=v_{0}=270-54=216$). When the number of anchor points (i.e., the number of unknown coefficients in Eq. (\ref{Map})) exceeds the number of training points in the dataset the system undergoes a transition to an under-determined. This change to an under-determined ``phase'' is evinced by numerous metrics (including Fig. (\ref{fig:Heart-RMS})).
\end{thebibliography}

%

\end{document}